\documentclass[sigconf]{acmart}

\usepackage{balance}
\usepackage{amsmath}
\usepackage{enumerate}
\usepackage{bbding}
\usepackage{longtable}
\usepackage{graphicx}
\usepackage{subfigure}
\usepackage{multirow}
\usepackage{bm}
\usepackage{ifpdf}
\AtBeginDocument{%
  \providecommand\BibTeX{{%
    \normalfont B\kern-0.5em{\scshape i\kern-0.25em b}\kern-0.8em\TeX}}}

\copyrightyear{2020}
\acmYear{2020}
\setcopyright{acmcopyright}
\acmConference[MM '20]{Proceedings of the 28th ACM International Conference on Multimedia}{October 12--16,2020}{Seattle, WA, USA}
\acmPrice{15.00}
\acmDOI{10.1145/3394171.3413709}
\acmISBN{978-1-4503-7988-5/20/10}

\begin{document}

\fancyhead{} 
\title{Deep Multi-modality Soft-decoding of Very \\Low Bit-rate Face Videos}

\author{Yanhui Guo}
\email{guoy143@mcmaster.ca}
\affiliation{%
  \institution{McMaster University}
  \city{Halmilton}
  \state{Canada}
}

\author{Xi Zhang}
\email{zhangxi_19930818@sjtu.edu.cn}
\affiliation{%
  \institution{Shanghai Jiao Tong University}
  \city{Shanghai}
  \state{China}
}

\author{Xiaolin Wu}
\email{xwu@ece.mcmaster.ca}
\affiliation{%
  \institution{McMaster University}
  \city{Halmilton}
  \state{Canada}
}
\renewcommand{\shortauthors}{Yanhui Guo et al.}
\newcommand{\tabincell}[2]{\begin{tabular}{@{}#1@{}}#2\end{tabular}}

\begin{abstract}
  We propose a novel deep multi-modality neural network for restoring very low
  bit rate videos of talking heads.  Such video contents are very common in social media, teleconferencing, distance education,
  tele-medicine, etc., and often need to be transmitted with limited bandwidth.
  The proposed CNN method exploits the correlations among three modalities, video, audio and emotion state of the speaker,
  to remove the video compression artifacts caused by spatial down sampling and quantization. The deep learning approach turns out
  to be ideally suited for the video restoration task,
  as the complex non-linear cross-modality correlations are very difficult to model analytically and explicitly.
  The new method is a video post processor that can significantly boost the perceptual quality of aggressively
  compressed talking head videos, while being fully compatible with all existing video compression standards.
\end{abstract}

\begin{teaserfigure}
   \subfigure[input]{
    \includegraphics[width=0.3\linewidth]{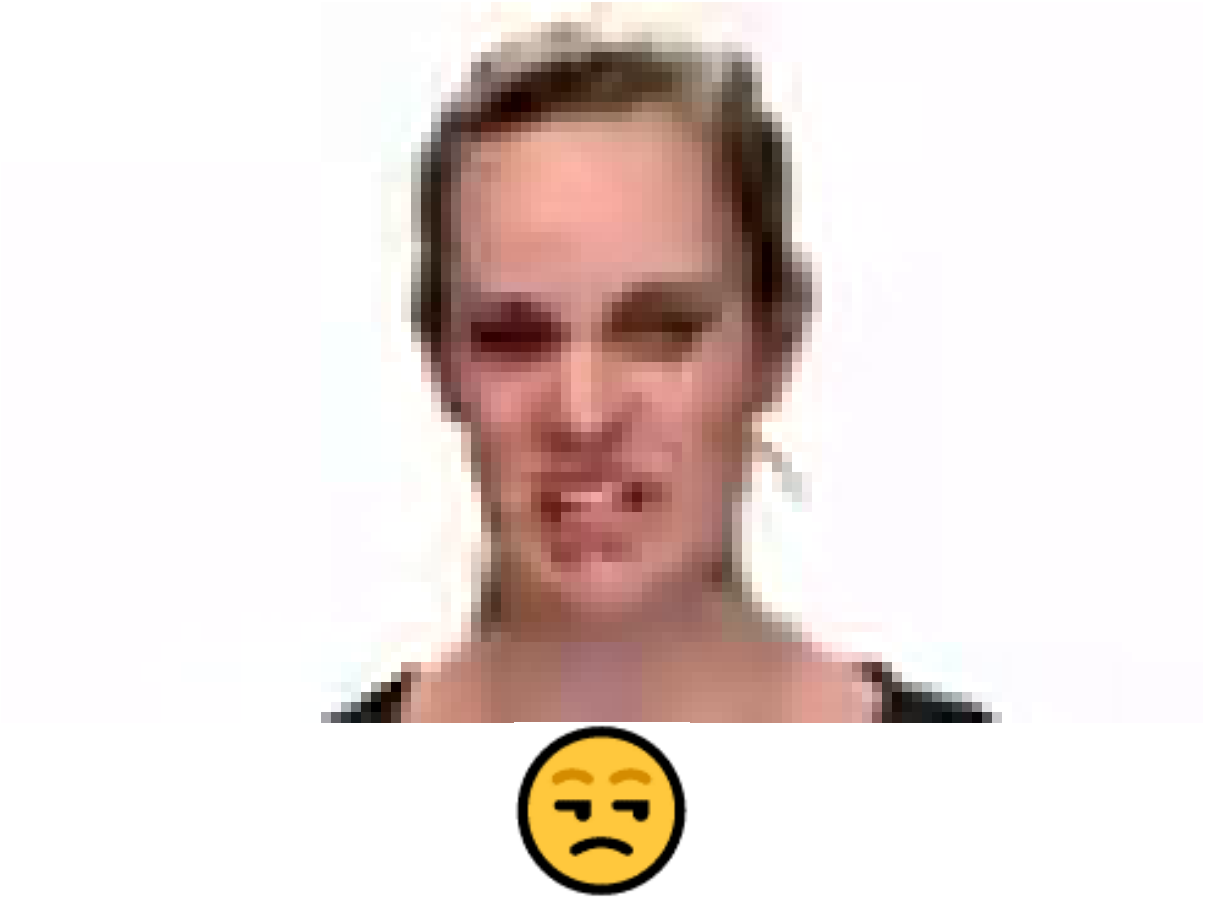}
  }
  \subfigure[image only]{
    \includegraphics[width=0.3\linewidth]{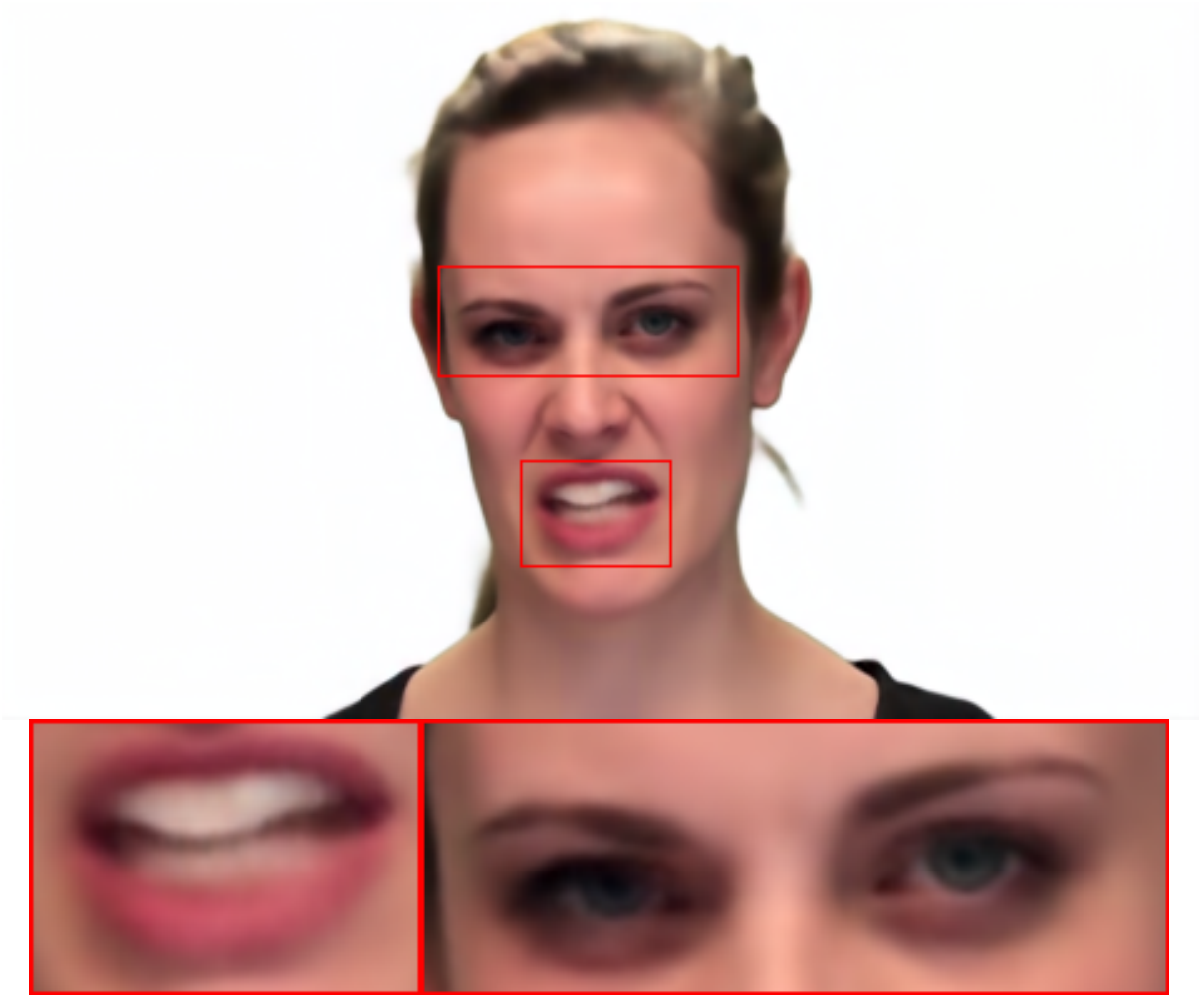}
  }
  \subfigure[+ audio]{
    \includegraphics[width=0.3\linewidth]{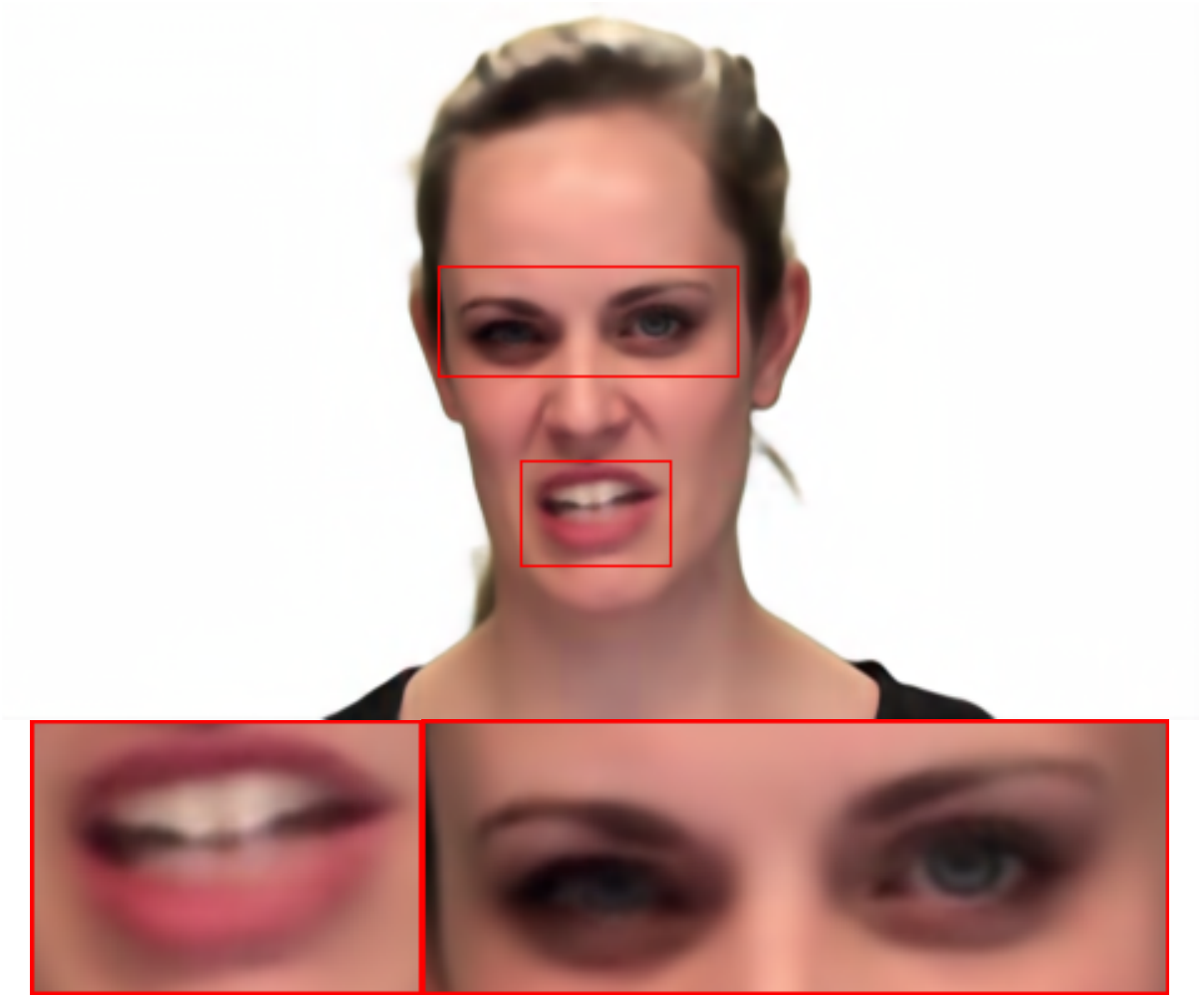}
  }
  \\
  \subfigure[+ audio + emotion]{
    \includegraphics[width=0.3\linewidth]{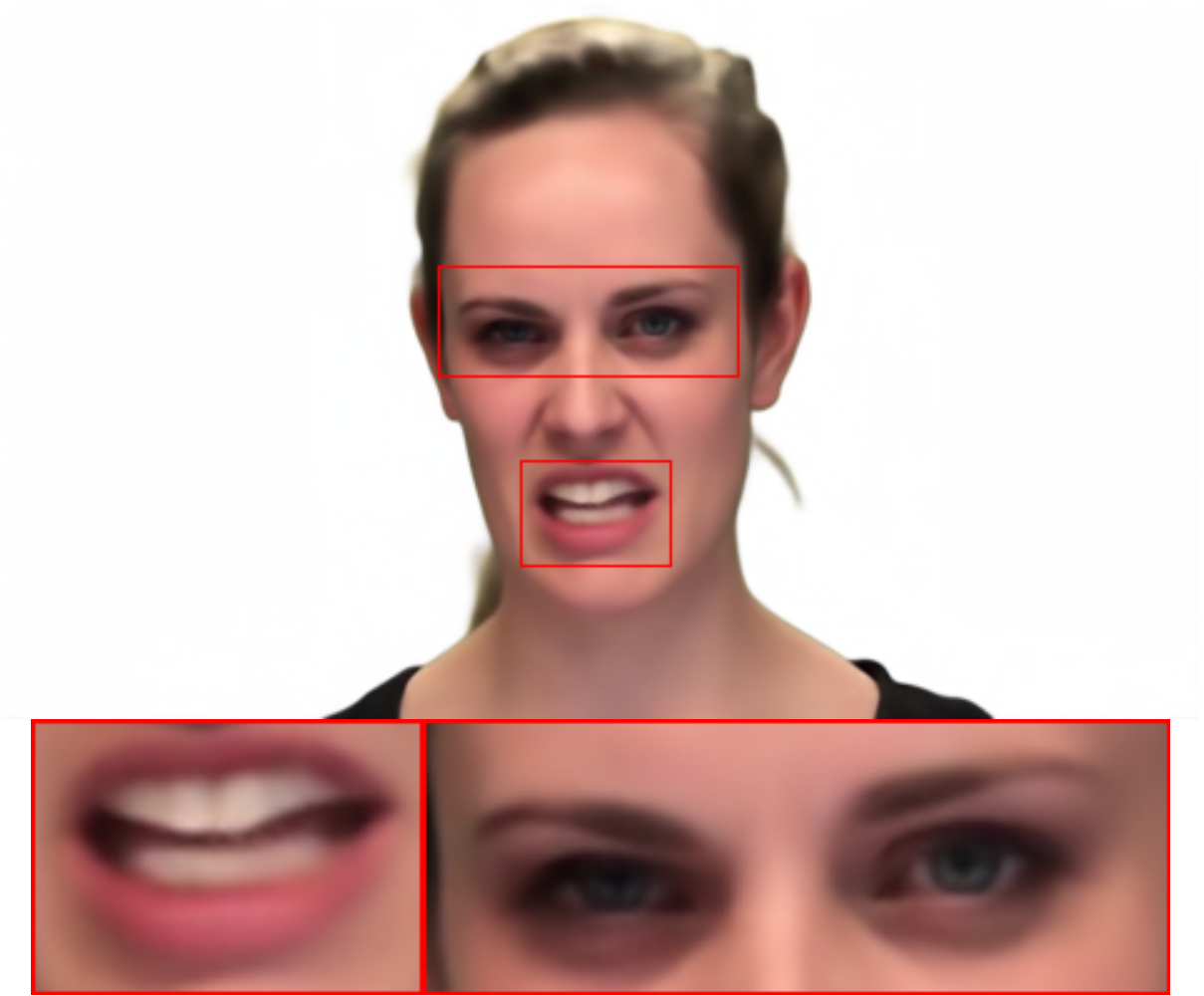}
    \label{MMVR_L1_ablation_teaser}
  }
  \subfigure[+ audio + emotion + cGAN ]{
    \includegraphics[width=0.3\linewidth]{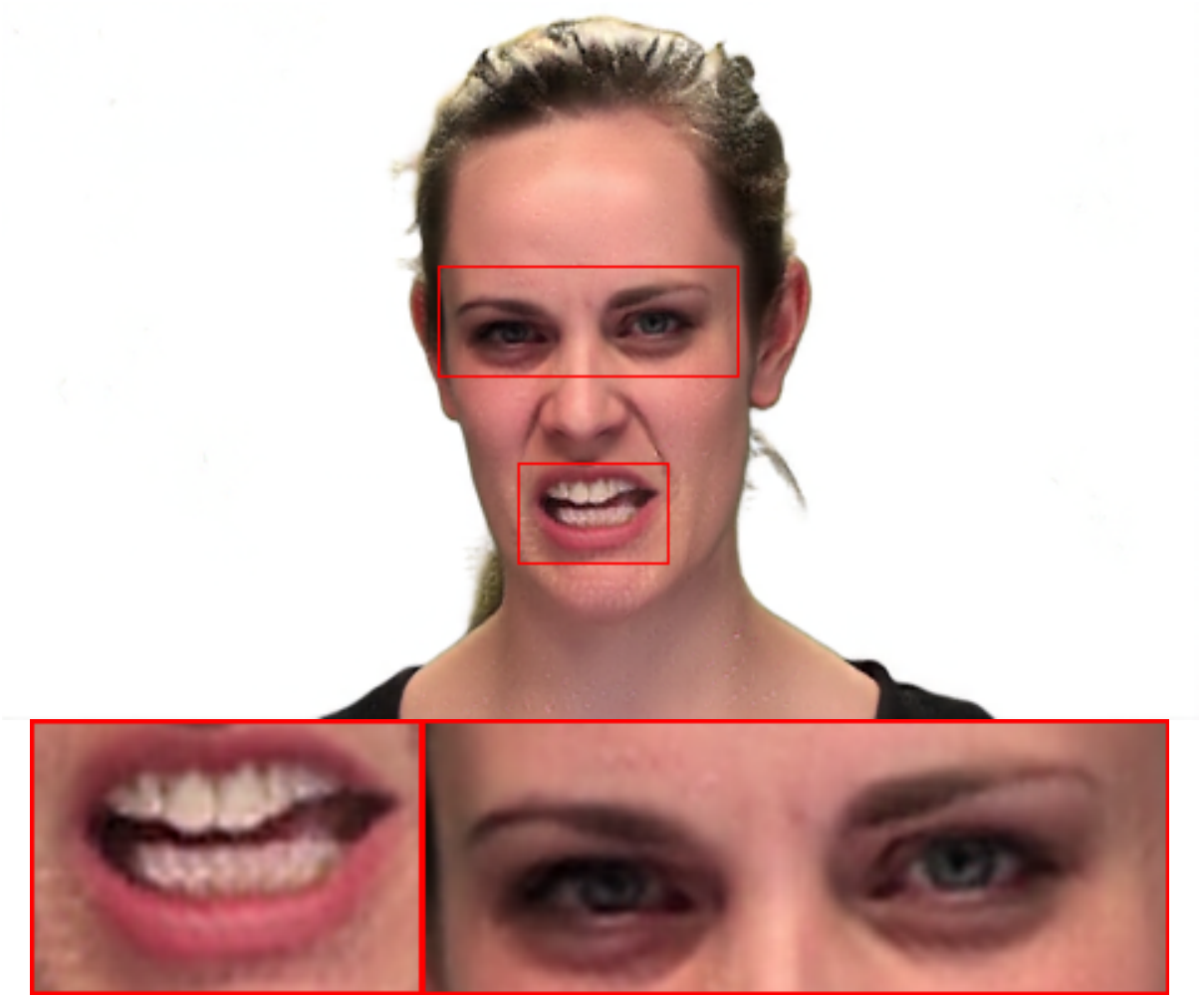}
  }
  \subfigure[ground truth]{
    \includegraphics[width=0.3\linewidth]{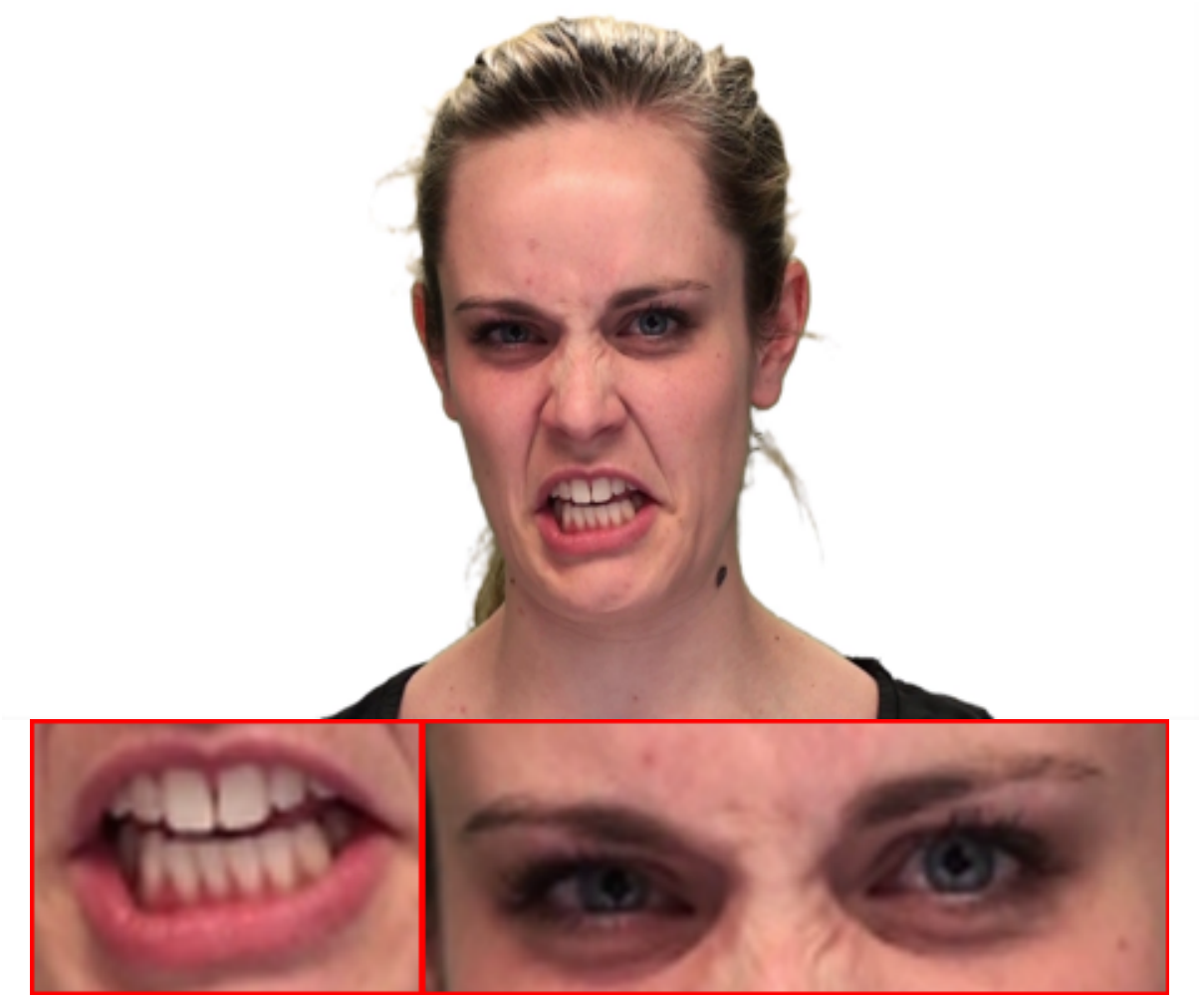}
  }
  \caption{Restoration results with increasing number of modalities used in the proposed video soft decoding CNN.}
  \label{teaser}
\end{teaserfigure}



\maketitle



\section{Introduction}
Stream videos constitute by far the largest volume of internet traffic, and
the video data volume is still rapidly growing with a wide range of multimedia applications, including video-on-demand services, social media, teleconferencing, tele-education, tele-medicine, tele-cooperations, etc.
Although the bandwidth and storage cost per megabyte is steadily decreasing, the rapid growth of the video data more than offset the cost reduction of mass data storage and communication bandwidth.
Cost aside, the bandwidth in wireless video communications among a large number of users can be easily overwhelmed.  Video compression is an indispensable internet and media technology to break the above bottlenecks.
In the history of internet and digital media industry, a number of video compression standards (e.g., MPEG-4 \cite{MPEG4}, H.264 \cite{H264}, and HEVC \cite{HEVC}) were developed to achieve progressively improved bandwidth economy of video transmissions.
Under severe bandwidth constraints, such as in the events of network congestions and only partially available network services, video streaming has to be carried out at very low bit rate. Video compression standards achieve very low bit rates by more
aggressive quantization and spatial down sampling of the frames.  These rate reduction operations inevitably cause compression artifacts, such as ringing, blurring, jaggies and blocking. In this paper, we propose a novel deep learning multi-modality
method to remove the compression artifacts and thus improve the visual quality of low bit rate videos. The proposed video restoration CNN is designed to post-processes the decoded videos of a given video compression standard, and for this reason it is called
CNN soft decoding. In the system aspect, being a postprocess, the proposed soft decoding method is innately compatible to all existing video compression standards.


The main innovation of this research is a multi-modality approach to the problem of removing video compression artifacts, as illustrated in Fig.~\ref{MMVR_L1_ablation_teaser}.  As a showcase for the new approach, we focus on soft decoding of very low bit rate face videos in
lively speech or conversation. Arguably, the most common and interesting object in social media and internet streaming videos is the face of a person speaking.  Talking heads are the focal point in video calls, teleconferences, Internet talk shows (TED and the alike), movies,
television, self media, etc.  In these applications, viewer satisfactions are largely dependent on the clarity and expressiveness of faces in decompressed videos. In our multi-modality video restoration approach, we exploit the correlations of the face video signal with the facial expressions and
with the audio signal. The facial expression or the emotion of a speaker is a very strong prior for the restoration of low bit rate face videos, because it puts structural constraints on the shapes and relative positioning of eyes, eye brows, lips, nose and chin.  In contrast, existing CNN methods
for video restoration are all of the single modality, such as those for compression artifacts removal  \cite{yang2018multi,lu2018deep,xue2019video} and those for video super-resolution \cite{liao2015videosp,Jo_2018_CVPR,EDVR,DBPN}.


In practice, the video encoder can use existing emotion detection algorithms \cite{Face_AUs,RAVDESS} to generate and transmit the side information on facial expressions. This side information is very compact and is almost free in bandwidth, merely being a flag of 15 different expressions (in our experiment setting),
but it can yield great gain in perceptual quality at the decoder side. To the success of the CNN soft decoding, knowing the facial expressions of the faces is critical and particularly so when network congestions and sporadicity drop the bit rate to so low a level that the decompressed face
features become hardly legible. In this case, the facial expression prior can be incorporated into and used by a generative adversary network (GAN) to hallucinate the face with the targeted expression.

In support of the expression generation GAN, the CNN soft video decoding is tuned for
faithful semantic reconstruction instead of pure signal level fidelity.
The audio signal is the second modality that can provide information to refine the reconstruction of low bit rate face videos.  The strong correlation between a speaker's audio and face video is self-evident.  Physiologically, facial muscles, particularly those in the lips, shape the sound and air stream into speech.
It is therefore natural to consider the joint audio-visual statistics when solving the inverse problem of restoring compressed face videos. As the underlying inverse problem is so ill posed due to the very low bit rate, one needs to use all available priors to restrict the solution space as much as possible.
Deep convolutional neural networks offer an ideal machinery to exploit both facial expression and audio priors in the soft decoding task.  In what follows, we devise such a CNN called multi-modality soft decoding network (MMSD-Net) for joint compression artifacts removal and super-resolution of downsampled face videos.\par

This paper is organized as follows. In section \ref{Related}, we make a brief review of recent related works. The proposed MMSD-Net video restoration method is presented in section \ref{Methodology}. In section \ref{Experiments}, we report the setup and results of our experiments, together with a comparison study with competing methods. Section \ref{Conclusion} concludes.

\section{Related Work} \label{Related}
\paragraph{\textbf{Deep video restoration}}
The simplest video restoration approach is to restore degraded frames one by one.  In disregard of the effectiveness, all existing image restoration methods can be used as single-frame solutions to the problem.
Liao {\it et al.} seemed to be among the first to work on multi-frame video super-resolution by factoring in motions with a conventional optical
flow analysis \cite{liaovideosp2015}.
Tao {\it et al.} improved the performance of video superresolution by using  sub-pixel motion estimation and compensation \cite{tao2017detail}.
Jo {\it et al.} proposed an end-to-end deep neural network that predicts dynamic upsampling filters and a residual image, exploiting joint spatio-temporal correlations \cite{Jo_2018_CVPR}.
Wang {\it et al.} proposed a generic video restoration framework using a pyramid, cascading and deformable alignment module and a temporal-spatial attention fusion module, and they achieved the state-of-art performance in video restoration\cite{EDVR}.
He {\it et al.} developed a video decompression technique making use of the coding block information of the encoder structures \cite{heHevc2018}.
Lu {\it et al.} built a recursive filtering scheme based on the Kalman model and restored decoded frames through a deep Kalman filtering network \cite{lu2018deep}.
Lately, Yang {\it et al.} published a SVM based detector to locate peak quality frames in compressed video that are fed into a multi-frame convolutional neural network to restore compressed videos \cite{yang2018mfqe}.
Xu {\it et al.} proposed a non-local ConvLSTM method to exploit consecutive frames in video compression artifact reduction and claimed the state-of-the-art performance \cite{xu2019non}.

\paragraph{\textbf{Multi-modality fusion with deep neural networks}}
Cross modality fusion is a difficult and interesting problem.
Ngiam {\it et al.} presented a simple bimodal autoencoder to learn a shared representation between modalities \cite{ngiam2011multimodal}.
Andrew {\it et al.} adapted canonical correlation analysis to deep neural networks, maximizing the correlation between representations \cite{andrew2013pmlr}.
Kim {\it et al.} brought up a hierarchical 3-level CNN architecture to combine multi-modality sources \cite{kimICMI2015}.
Zhang {\it et al.} proposed a multimodal deep convolution neural network for emotion recognition via fusing audio, visual and text modalities\cite{zhangmdcnn2016}.
Joze {\it et al.} proposed a multi-modality transfer module using squeeze and excitation operations and applied it to hand gesture recognition, speech enhancement, and action recognition \cite{joze2019mmtm}.
P{\'e}rez-R{\'u}a {\it et al.} tackled multi-modality fusion problem of searching good architectures of neural network by an efficient sequential model-based exploration approach \cite{perez2019mfas}.

\paragraph{\textbf{Deep joint audio-aided video synthesis}}

Some interesting papers were published on the video synthesis of talking heads to match an audio track.  Wiles {\it et al.} introduced a face generation CNN that can puppeteer a source face given a driving vector
such as audio data or pose code vector \cite{Wiles_2018_ECCV}.
Jamaludin {\it et al.} generated synthetic talking face video frames using a joint embedding of the face and audio \cite{Jamaludin_2019}.
Suwajanakorn {\it et al.} proposed a recurrent neural network to automatically map raw audio features to mouth shapes of a given speaker video
and produce photorealistic lip synchronization \cite{suwajanakorn2017synthesizing}.
All these works are about audio-assisted face video synthesis, while in this work we focus on the removal of compression artifacts in talking head videos using not only audio but also prior information on the speaker's emotion.


\section{Methodology} \label{Methodology}

\subsection{Framework}

\begin{figure*}[ht]
  \centering
  \includegraphics[width=1\textwidth]{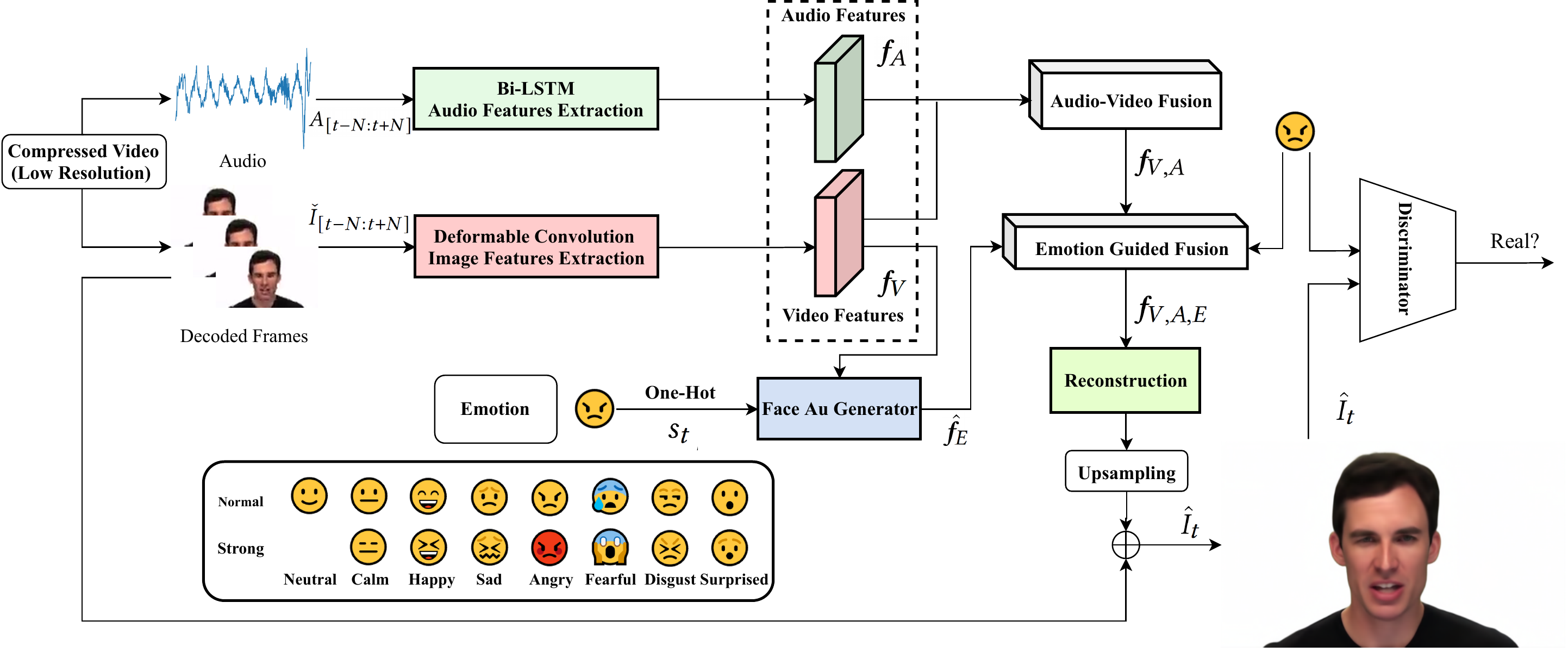}
  \caption{The framework of the proposed Multi-modality Soft Decoding Neural Network (MMSD-Net).}
  \label{figure_overall}
\end{figure*}

Formally, let $\{\mathnormal{\check{I}}_{\mathnormal{t}} | t= 0,1,2 \dots\}$ represent an aggressively compressed video sequence which is first spatially down sampled and then coded by a video compression standard (e.g. H.264/H.265).
When playing the video, the compressed video will be decompressed
and resized to a desired resolution $\mathnormal{\hat{I}_{\mathnormal{t}}}$.
The goal of video soft decoding is to restore the hard decoded video
$\mathnormal{\hat{I}_{\mathnormal{t}}}$ to the state of the
original high quality video $\mathnormal{I_{\mathnormal{t}}}$ the best way possible in a chosen quality metric.

In our proposed MMSD CNN, given $2\mathnormal{N}+1$ consecutive low-quality frames $\mathnormal{\check{I}}_{\left [\mathnormal{t}-\mathnormal{N}:\mathnormal{t}+\mathnormal{N}\right ]}$, the network is to produce the refined
high-quality central frame $\mathnormal{\hat{I}_{\mathnormal{t}}}$ by maximally removing artifacts.  For the video contents of talking heads, the audio signal $A_{\left [\mathnormal{t}-\mathnormal{N}:\mathnormal{t}+\mathnormal{N}\right ]}$
and the emotion state $s_t$ are used to assist the video restoration.
Specifically, the video reconstruction problem can be stated as following:
\begin{equation}
  \mathnormal{\hat{I}_{\mathnormal{t}}} = \mathnormal{G}\left(\mathnormal{\check{I}}_{\left [\mathnormal{t}-\mathnormal{N}:\mathnormal{t}+\mathnormal{N}\right ]},A_{\left [\mathnormal{t}-\mathnormal{N}:\mathnormal{t}+\mathnormal{N}\right ]},s_t \right)
\end{equation}
where $\mathnormal{G}(\cdot)$ represents the proposed MMSD neural network to be optimized.

The architecture of the proposed MMSD-Net for low bit rate video restoration is illustrated in Fig.\ref{figure_overall}.  It consists of three modality branches for video, audio and emotion, respectively.
In the visual modality branch, we use a CNN subnet to extract inter-frame features. This subnet can be conceptually understood as a module to align features of neighboring frames.

In the audio modality branch, we apply a bi-directional LSTM to analyze the temporal audio signal and extract features. In order to fuse the audio and video features, we convert temporal audio features into 2D maps using upsampling subnet.

In the emotion modality branch, we adopt the well-known facial action coding system (FACS) that anatomically characterizes facial expressions \cite{ekman1978}. FACS decomposes facial expressions into so-called action units (AU).
We insert it into MMSD-Net a face AU generation subset to predict the AUs of face video frames from combined audio-video features and the side information on the known facial expression of the current frame provided by the encoder.
The extracted features of all three modalities are fused and fed into a reconstruction subnet consisting of a cascade of 10 residual convolution blocks and upsampling layers to estimate the latent high quality video.
In the multi-modality fusion, we apply the attention technique to judiciously associate the audio and emotion features with the relevant parts of the face.

Next, we introduce the details of individual elaborate modules of our proposed MMSD-Net.

\begin{figure}[ht]
  \centering
  \includegraphics[width=0.45\textwidth]{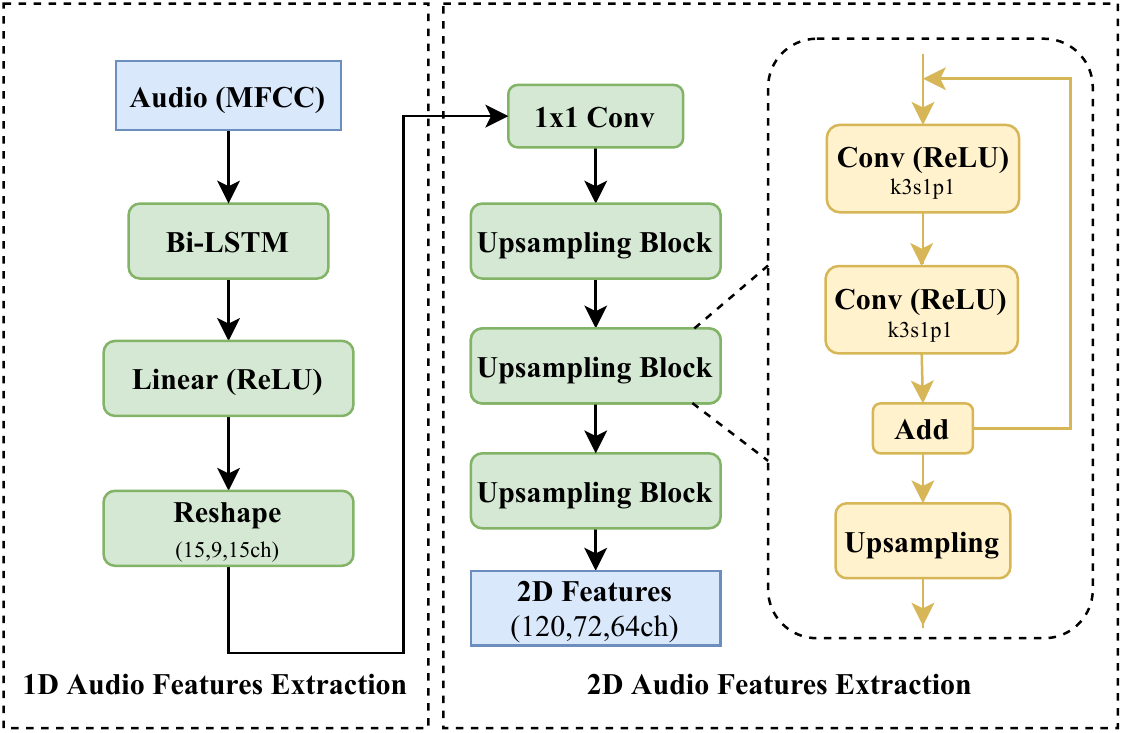}
  \caption{Audio features extraction branch. Left: 1D audio features extraction. Right: Convert temporal audio features into 2D maps.}
  \label{figure_afg}
\end{figure}

\subsection{Multi-modality features extraction}

As video soft decoding relies on correlations of consecutive frames, MMSD-Net needs to extract and process inter-frame features for the restoration task.
It takes $2N+1$ consecutive frames as the input and restores the middle frame $\mathnormal{\check{I}_{\mathnormal{t}}}$.
In order to account for motions between adjacent frames in feature extraction, MMSD-Net convolutes feature maps of consecutive time instances using deformable kernels
\cite{dai2017deformable} to generate inter-frame features. This deformable convolution approach proved to be effective in video restoration tasks \cite{EDVR, TDAN}.
The authors argued that it has the function of frame alignment. The resulted aligned frame features are passed through four residual blocks to generate
$64$ video feature maps $\bm{f}_{V}$, each of which is of dimension $120 \times 72$, the same as the input frames.  These video feature maps will be fused with the features of other modalities.

Before being processed by MMSD-Net the audio signal is converted to standard representation of mel-frequency cepstral coefficients (MFCC) \cite{MFCC_1, MFCC_2}.  To restore frame $t$ the network takes the $2N+1$ consecutive
audio frames $A_{\left [\mathnormal{t}-\mathnormal{N}:\mathnormal{t}+\mathnormal{N}\right ]}$ as input. Considering the temporal nature of audio, a 3 layers bi-directional LSTM subset
processes $A_{\left [\mathnormal{t}-\mathnormal{N}:\mathnormal{t}+\mathnormal{N}\right ]}$ to extract a 1-D audio feature vector from the MFCC coefficients.
For easy operations in the subsequent audio-video fusion stage, we need to transfer the 1D audio feature sequence to a stack of 2D $120 \times 72$ feature maps $\bm{f}_{A}$.  This is achieved by one fully connected layer
of dimension $15 \times 9 \times 15$ and three upsampling blocks, as shown in Fig.~\ref{figure_afg}.

\begin{figure}[ht]
  \centering
  \includegraphics[width=0.45\textwidth]{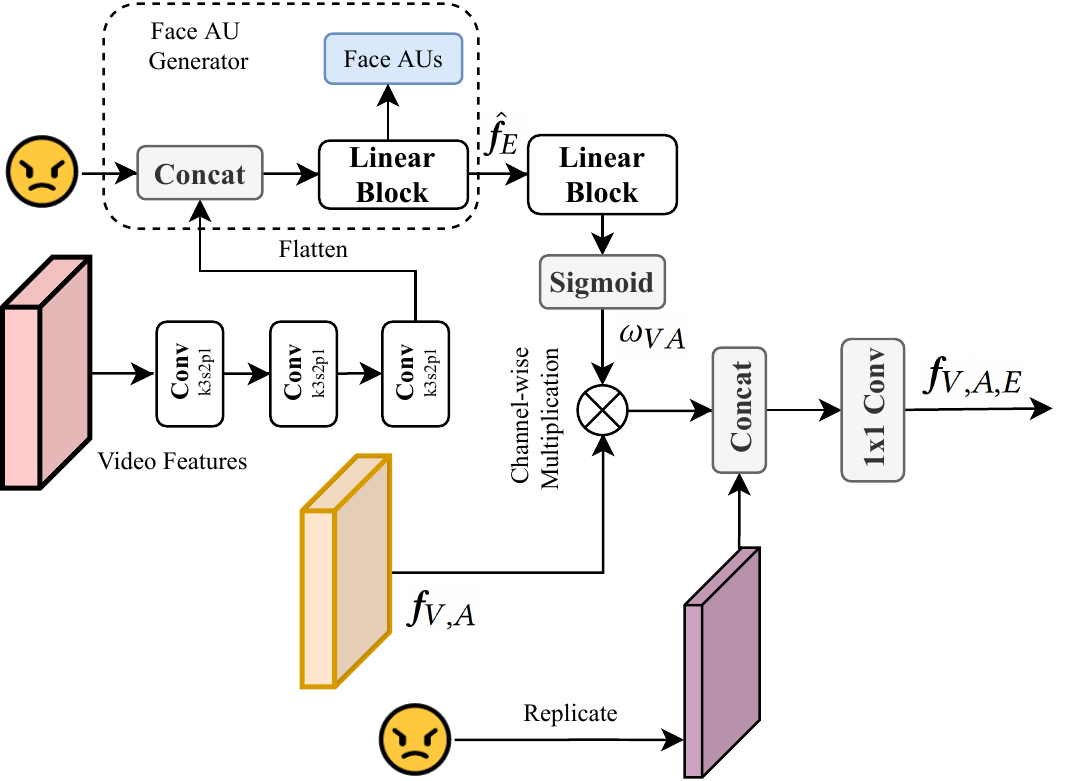}
  \caption{The subnetwork for fusing the three modalities: video, audio, emotion.}
  \label{figure_eaf}
\end{figure}

To extract emotion features for restoring frame $t$, we include in MMSD-Net a subnet that predicts the AUs of the face in this frame from the video features and the known emotion one-hot vector provided by the encoder,
as shown in Fig.\ref{figure_eaf}. The prediction is made by three fully connection layers using ReLU activation function and dropout layers, labeled as Linear Block in the figure.

\begin{figure}[ht]
  \centering
  \includegraphics[width=0.45\textwidth]{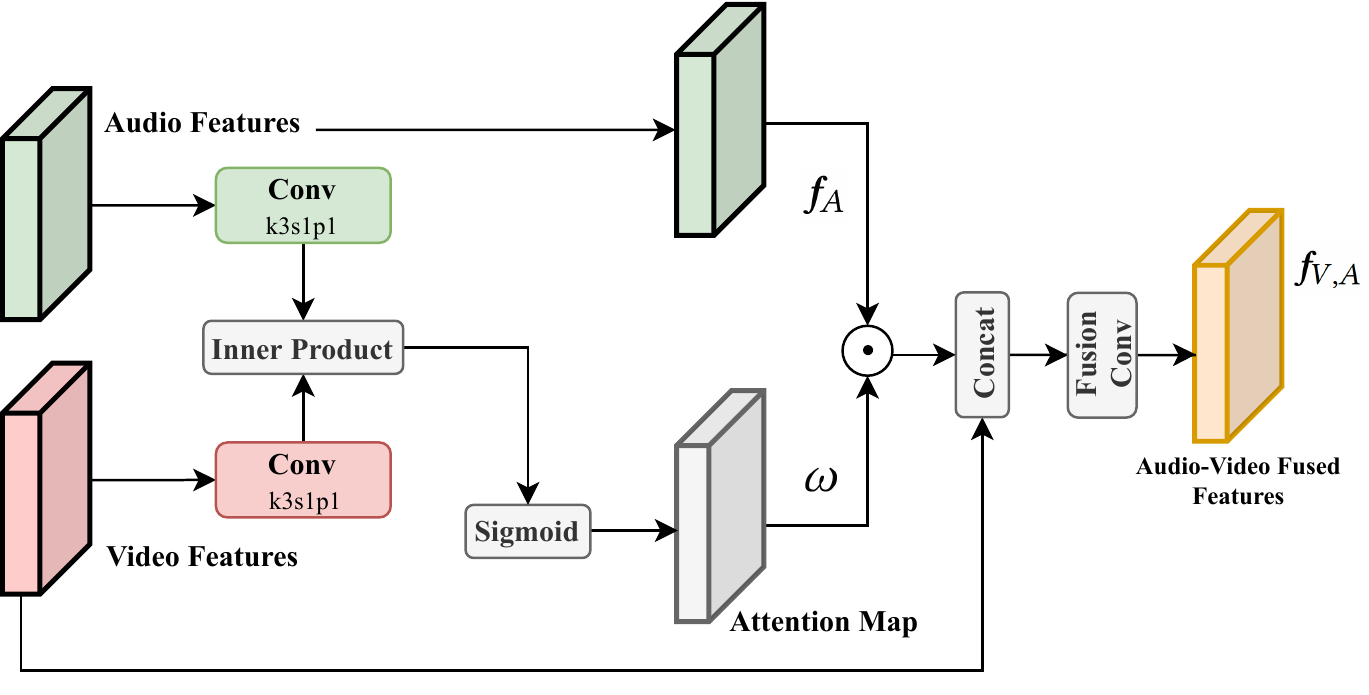}
  \caption{The details of audio-video fusion layer.}
  \label{figure_atf}
\end{figure}

\subsection{Video-audio fusion}
In order to control the complexity of MMSD-Net, we fuse the three modalities, video, audio and emotion, in two cascaded stages.
First, the audio and video features are fused, and the resulting joint audio-video features is further fused with emotion features.

The simplest way of fusing audio and video features is to concatenate them directly. However, as the audio signal mostly correlates to
the mouth region of the image (lips, chin, teeth), the multi-modality restoration should exploit the audio-video correlations in the
relevant spatial parts rather than in the entire image. In very low bit rate video of talking heads, it is difficult to reliably identify the mouth region.
Therefore, we adopt the CNN attention technique \cite{wang2018non,Attention_NIPS} and introduce in MMSD-Net an attention fusion subnet, as illustrated in Fig.\ref{figure_atf}.
After $120 \times 72$ video feature maps $\bm{f}_{V}$ and $120 \times 72$ audio feature maps $\bm{f}_{A}$ are extracted as explained in the preceding subsection,
the attention fusion subnet generates a $120 \times 72$ attention weighting map $\omega$ such that for spatial location $(x,y)$
\begin{equation}
  \omega(x,y) = \mathit{Sigmoid} (\theta(\bm{f}_{V}(x,y)) \cdot \phi (\bm{f}_A(x,y)))
\label{eq_weight}
\end{equation}
where $\cdot$ is the inner product operator for the video and audio feature vectors, $\theta(\cdot)$ and $\phi(\cdot)$ are embedding operators that are implemented by convolutions.  The weight $\omega(x,y)$ is defined by the inner product of the video and audio feature vectors
$\theta(\bm{f}_{{V}}(x,y))$ and $\phi(\bm{f}_{A}(x,y))$,
because it is a measure of the video-audio correlation at location $(x,y)$.  Note that the inner product is computed in the transform spaces of $\theta(\cdot)$ and $\phi(\cdot)$ that are optimized in the training stage of MMSD-Net.
In Eq(\ref{eq_weight}), the sigmoid function is used to map the resulting weight value into interval $(0,1)$.

Next, the audio feature map $\bm{f}_{A}$ are spatially weighted by $\omega(x,y)$ and then fused with the video feature map $\bm{f}_{V}$ via a convolution block.  The resulting fused audio-video feature maps are
\begin{equation}
  \bm{f}_{V, A} =  \mathit{Conv}\left(\left[\bm{f}_{V},\omega \odot \bm{f}_{A} \right] \right)
\end{equation}
where $\left [\cdot ,\cdot \right ]$ is the concatenation operator, $\odot$ stands for the pixel-wise multiplication, and $\mathit{Conv}$ is the fusion convolution with $1\times 1$ kernels. Then, as illustrated in Fig.~\ref{figure_overall},
the joint audio-video feature maps $\bm{f}_{V,A}$ is to be further fused with the emotion features.

\begin{figure*}[ht]
  \centering
  \centering
  \subfigure[Input]{
  \begin{minipage}[t]{0.05\linewidth}
      \includegraphics[width=1\linewidth]{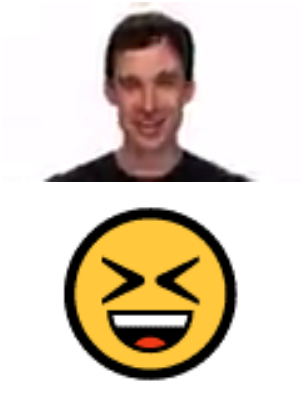}\vspace{25pt}
      \includegraphics[width=1\linewidth]{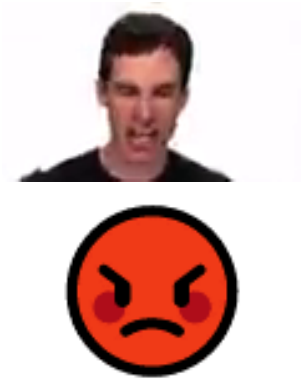}\vspace{25pt}
      \includegraphics[width=1\linewidth]{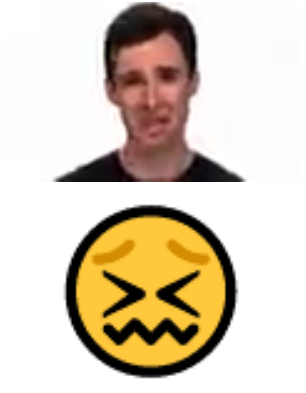}\vspace{25pt}
      \includegraphics[width=1\linewidth]{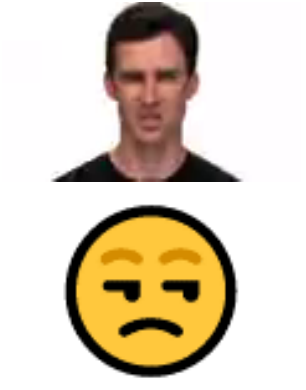}\vspace{27.5pt}
      \includegraphics[width=1\linewidth]{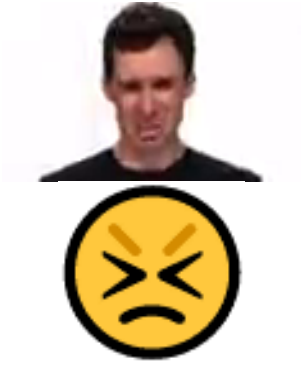}
  \end{minipage}
  }
  \subfigure[Bicubic]{
    \begin{minipage}[t]{0.144\linewidth}
      \includegraphics[width=1\linewidth]{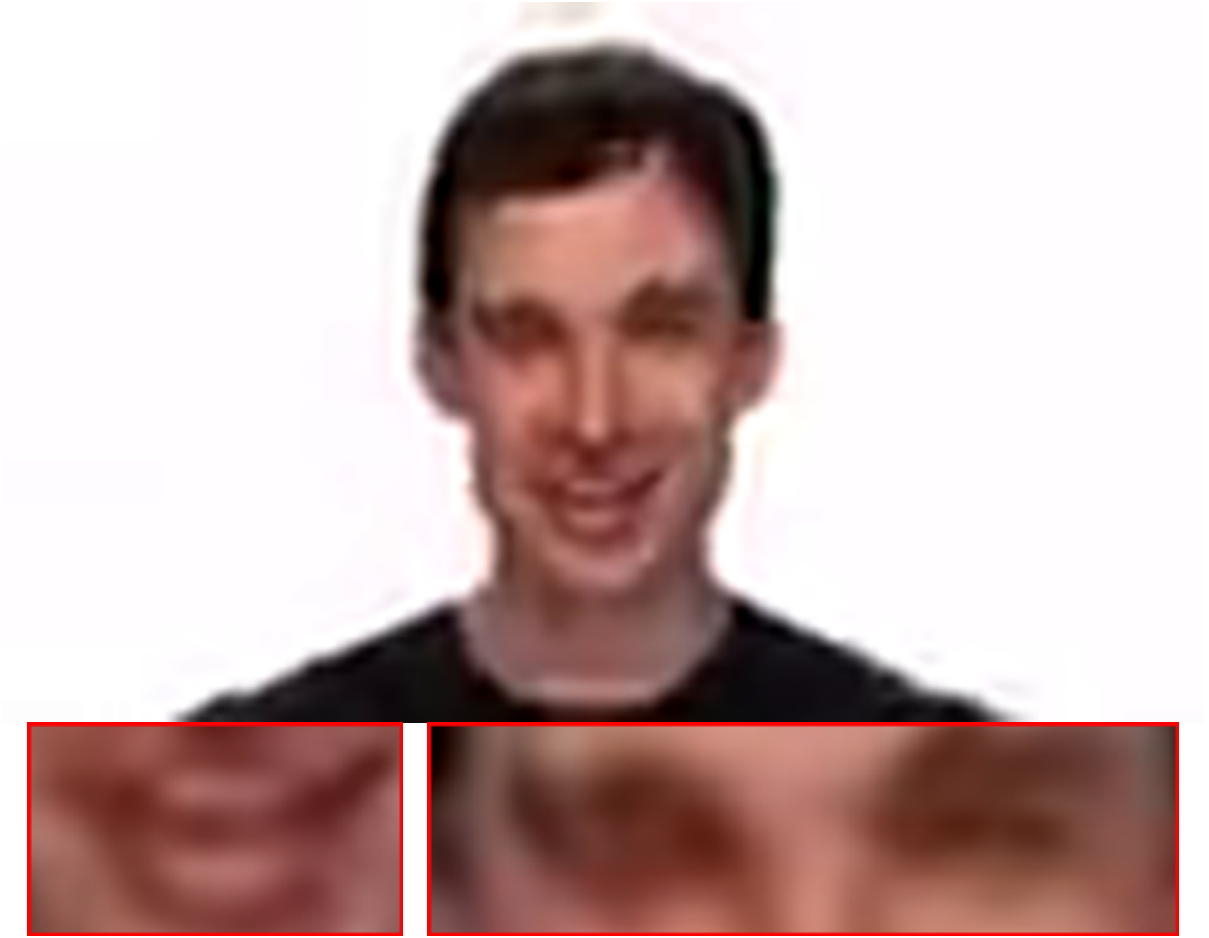}\vspace{1pt}
      \includegraphics[width=1\linewidth]{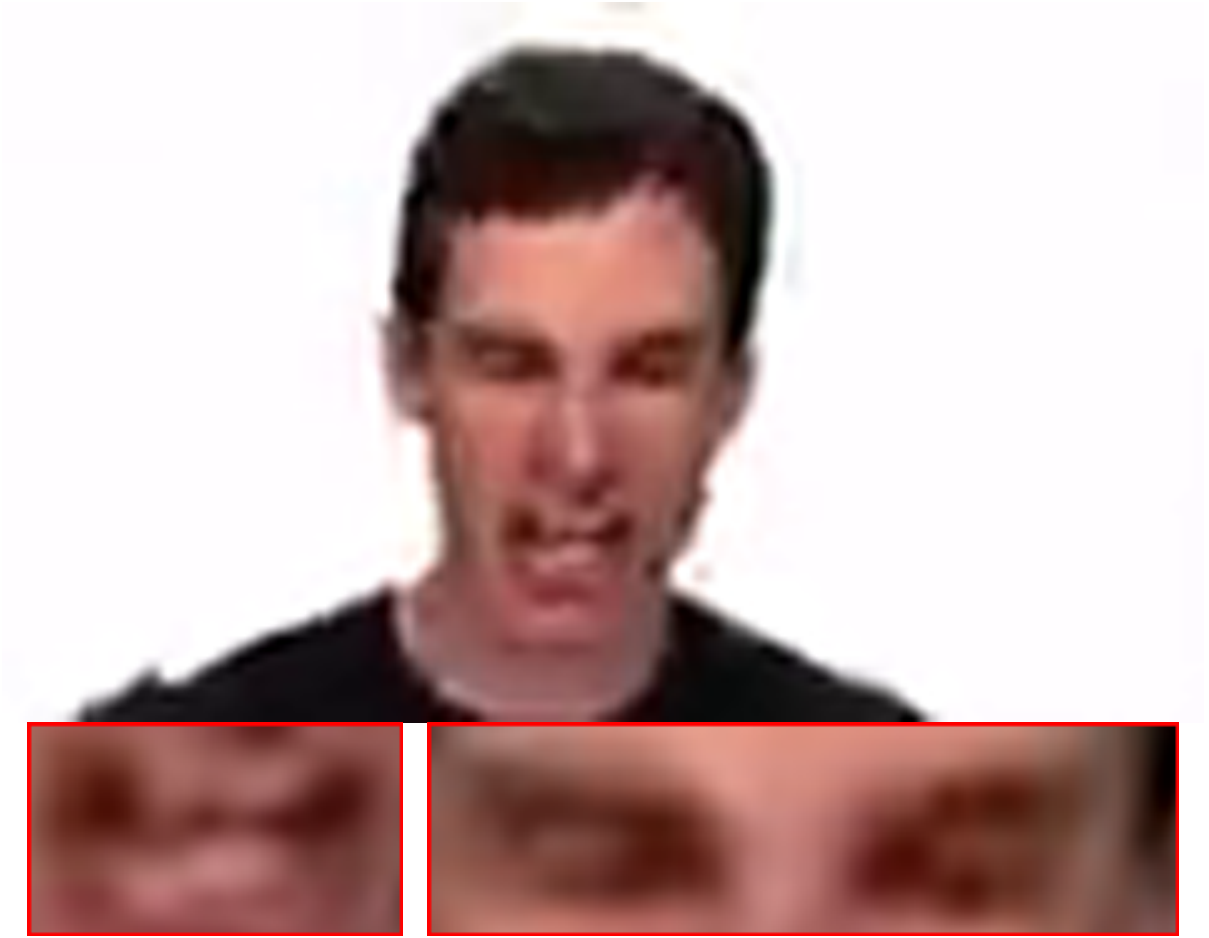}\vspace{1pt}
      \includegraphics[width=1\linewidth]{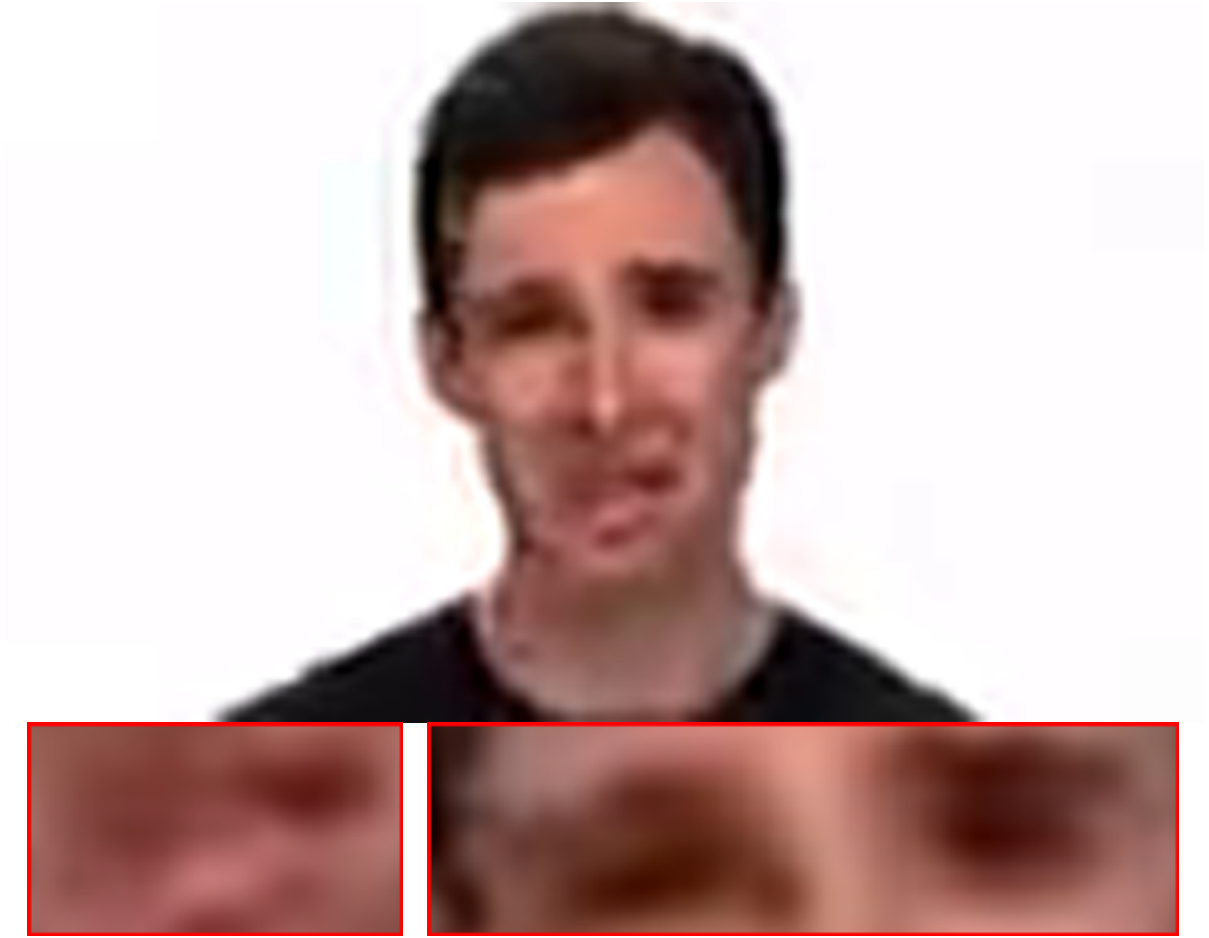}\vspace{1pt}
      \includegraphics[width=1\linewidth]{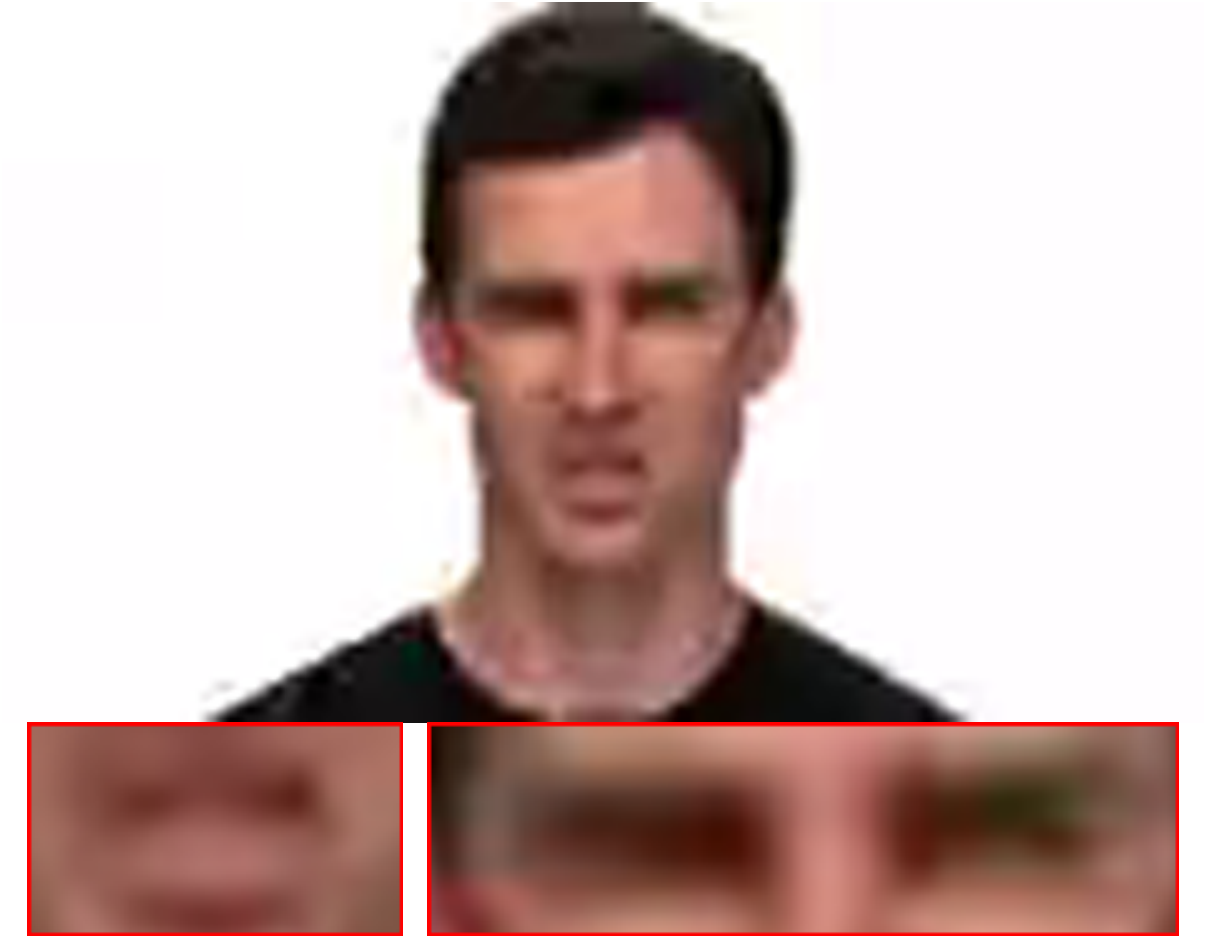}\vspace{1pt}
      \includegraphics[width=1\linewidth]{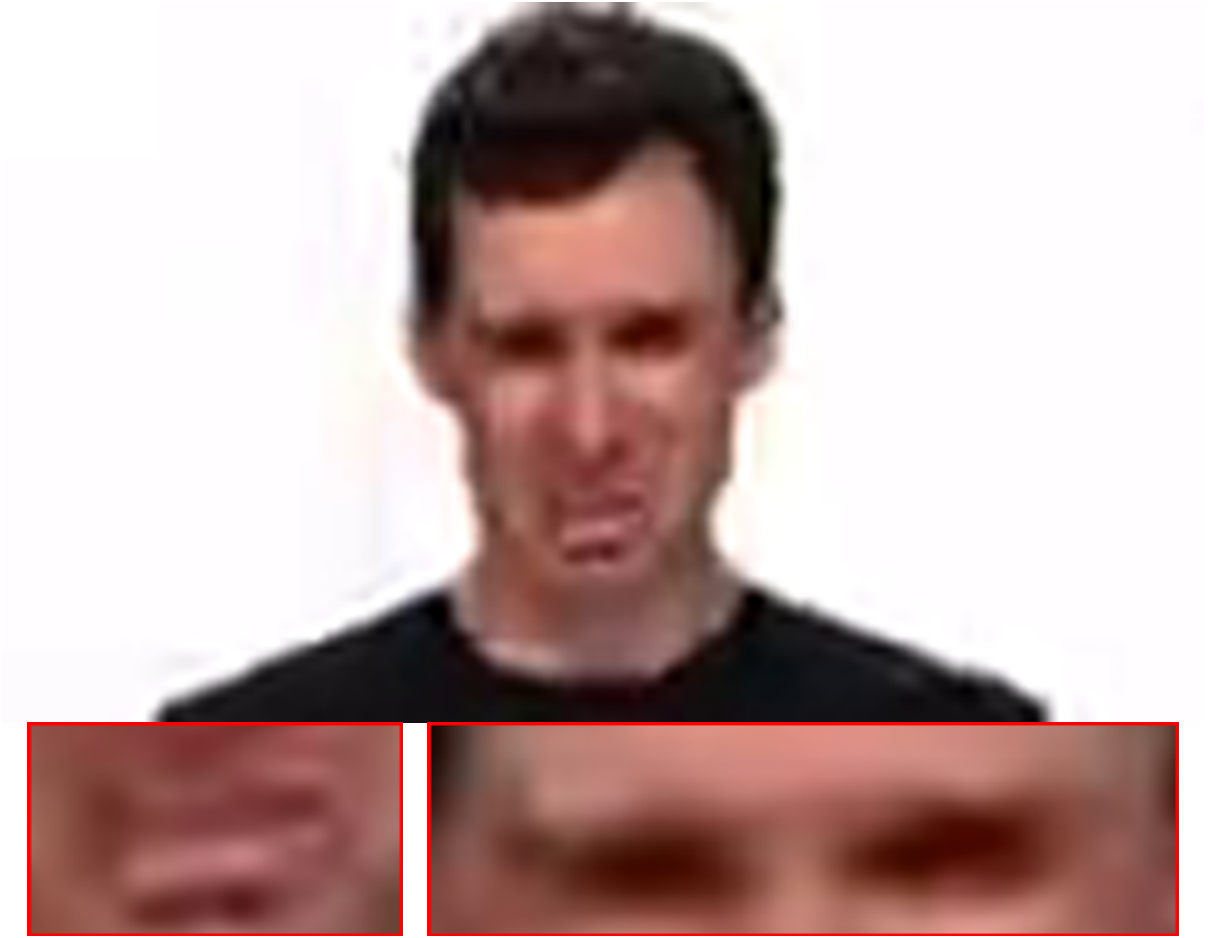}
    \end{minipage}
  }
  \subfigure[DBPN]{
    \begin{minipage}[t]{0.144\linewidth}
      \includegraphics[width=1\linewidth]{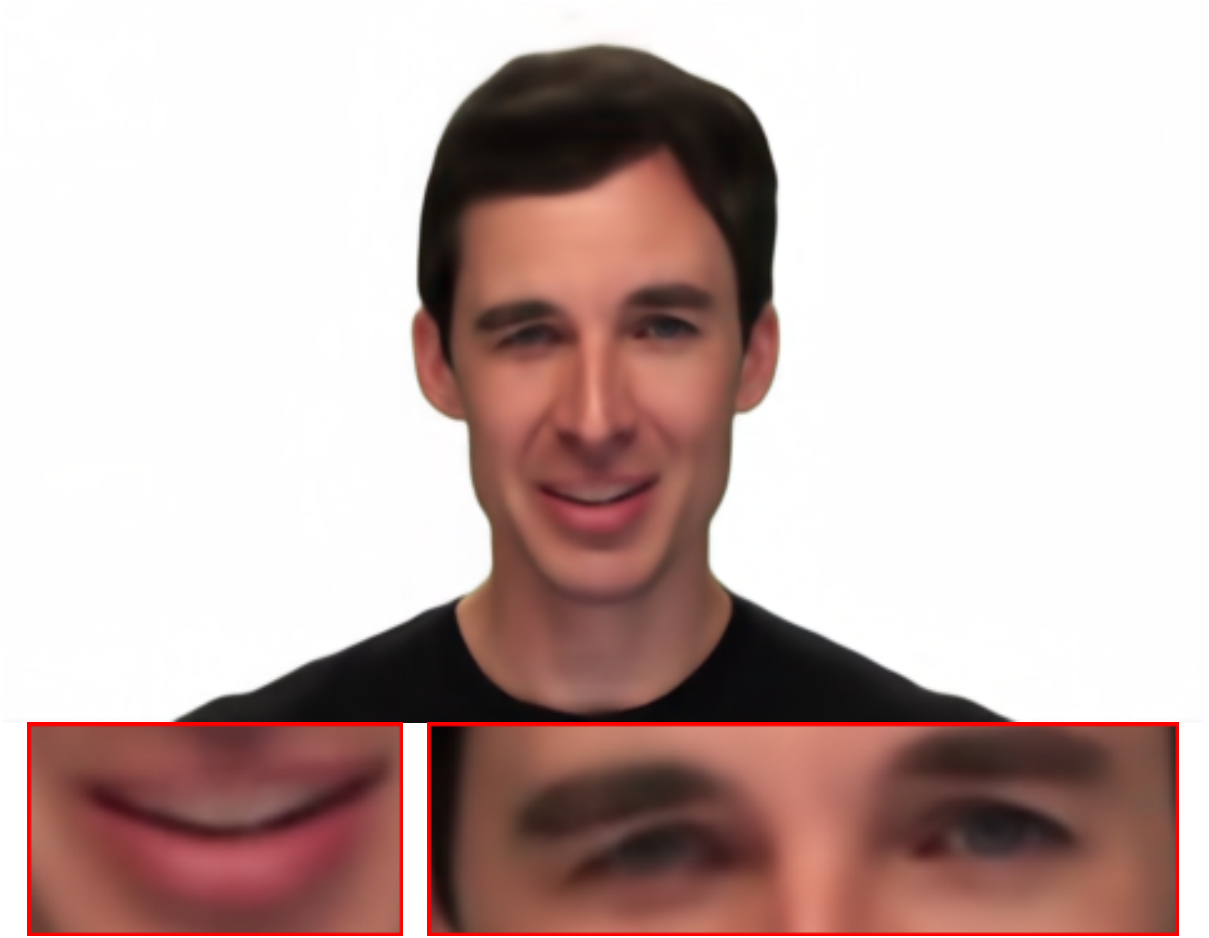}\vspace{1pt}
      \includegraphics[width=1\linewidth]{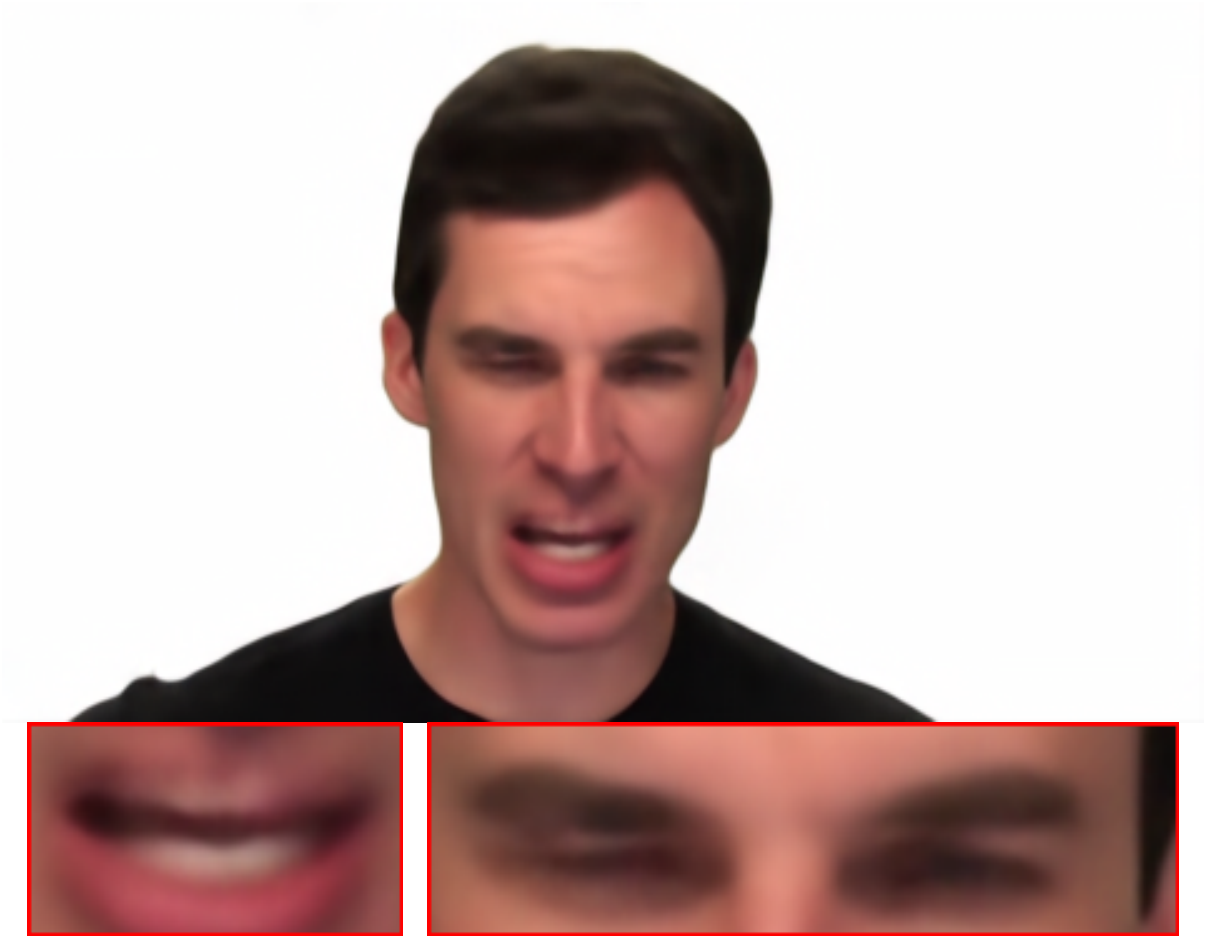}\vspace{1pt}
      \includegraphics[width=1\linewidth]{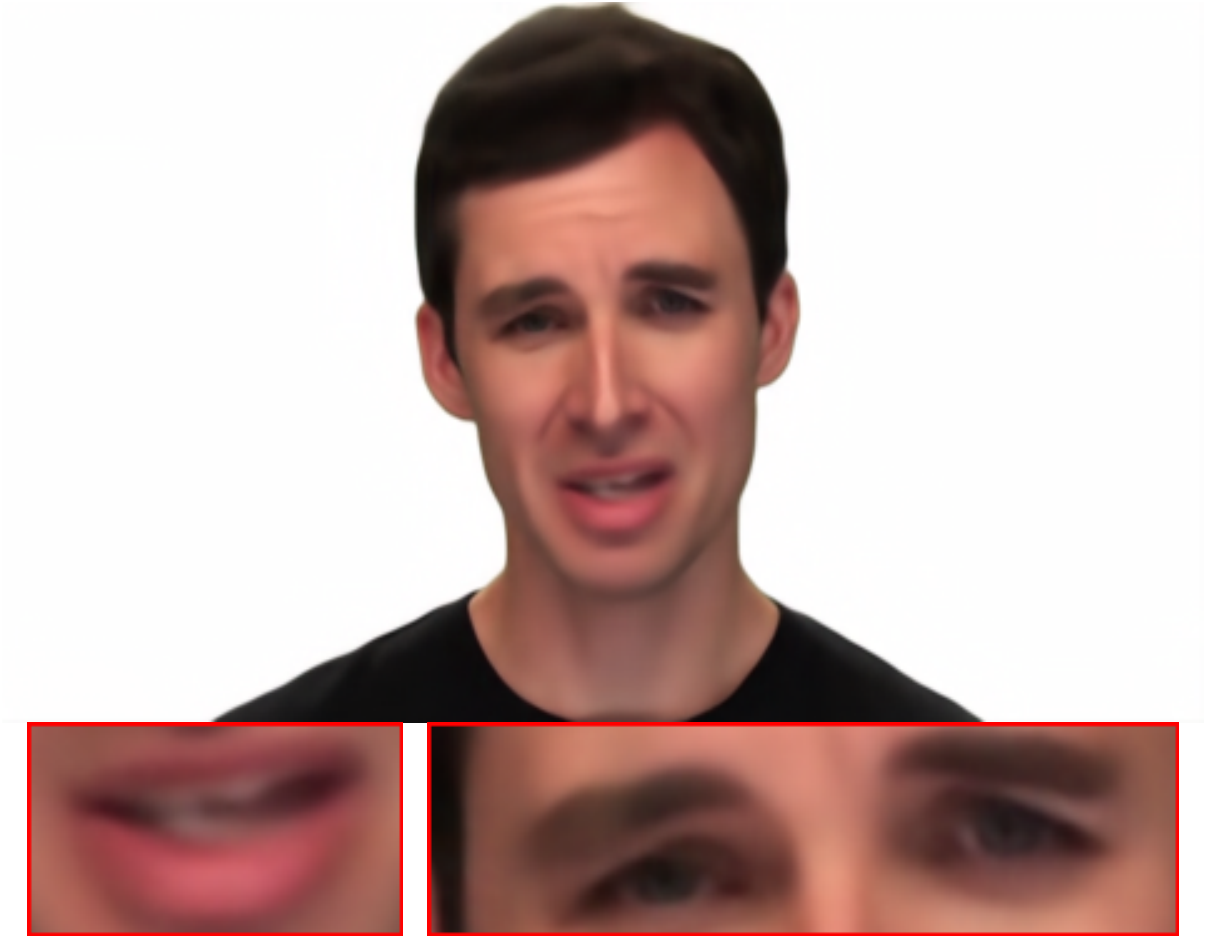}\vspace{1pt}
      \includegraphics[width=1\linewidth]{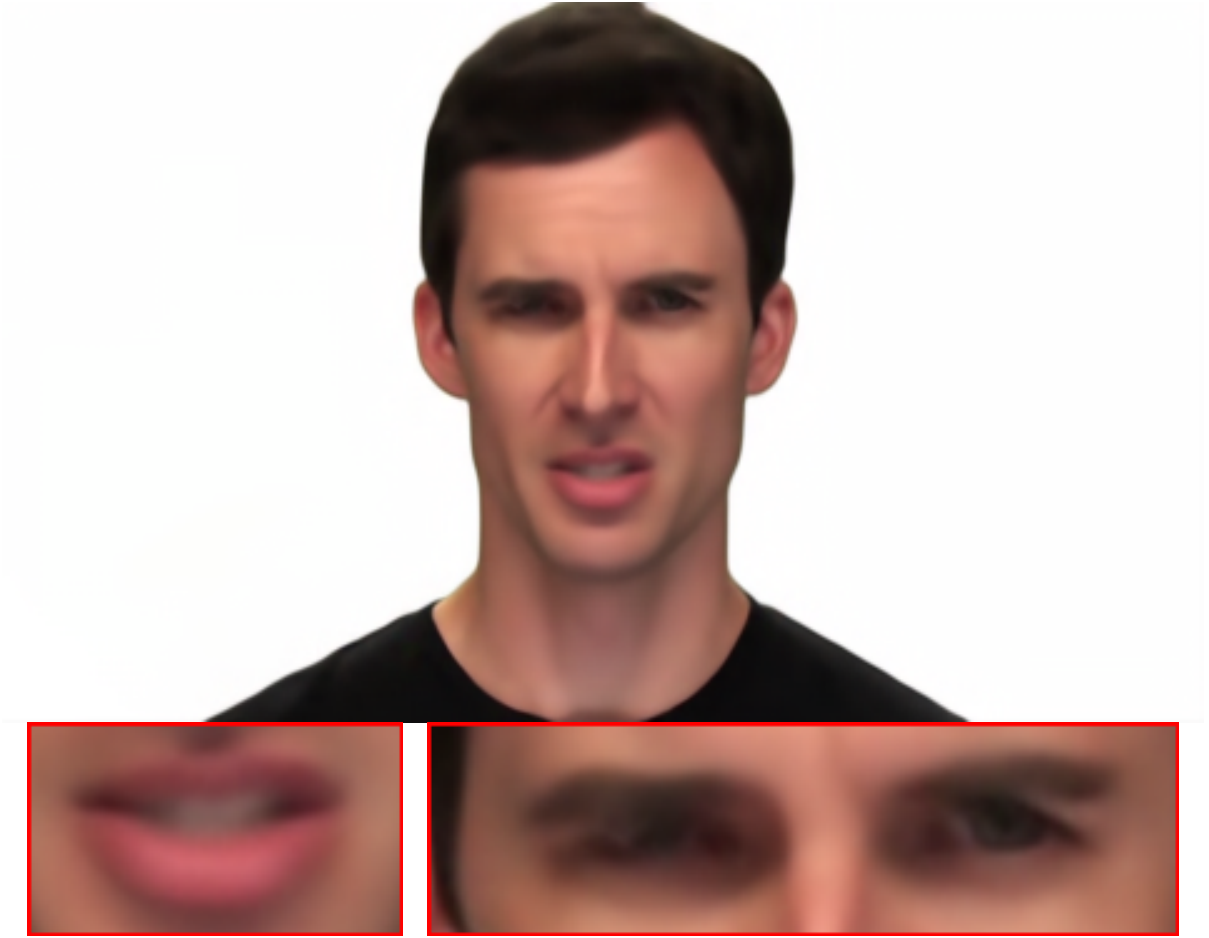}\vspace{1pt}
      \includegraphics[width=1\linewidth]{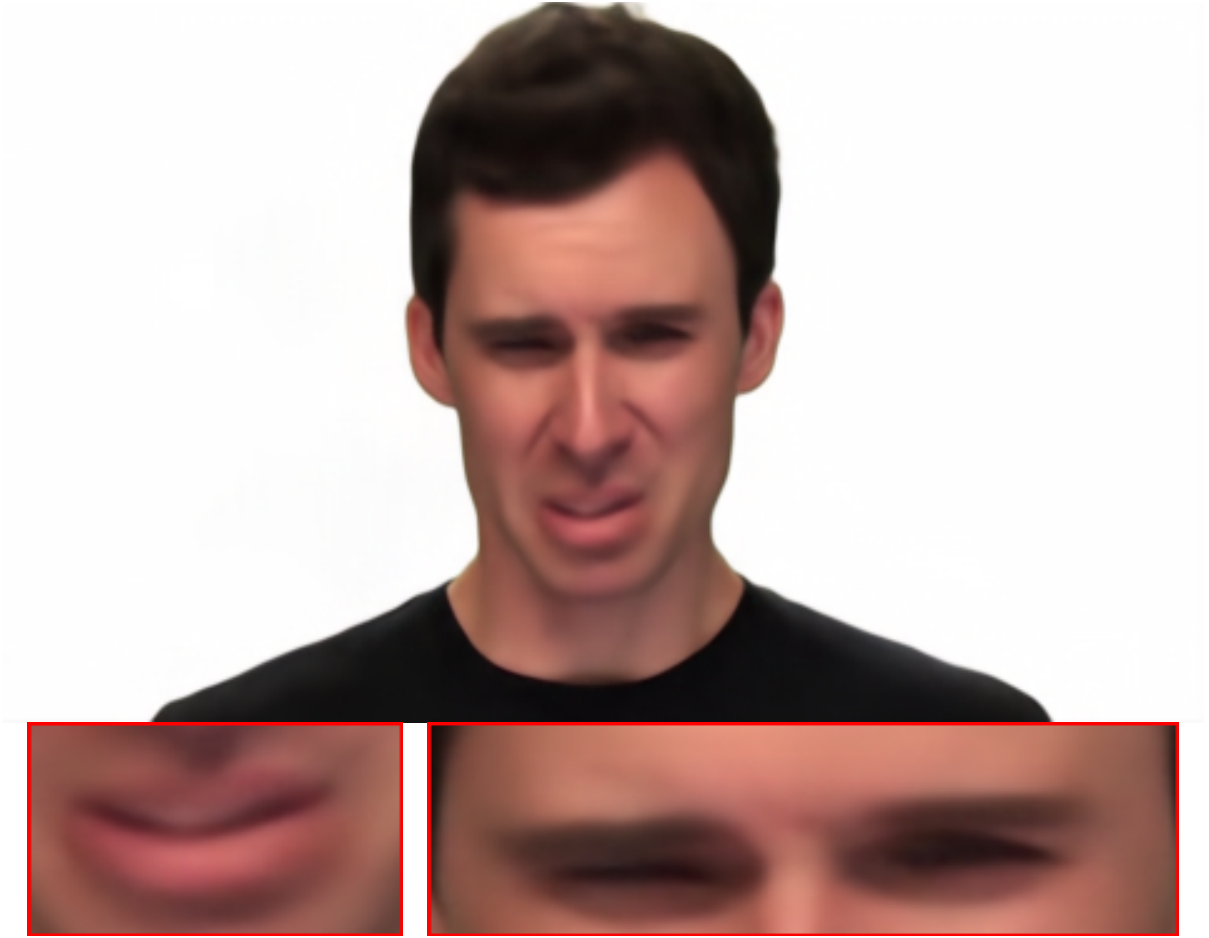}
    \end{minipage}
    }
  \subfigure[EDVR]{
    \begin{minipage}[t]{0.144\linewidth}
      \includegraphics[width=1\linewidth]{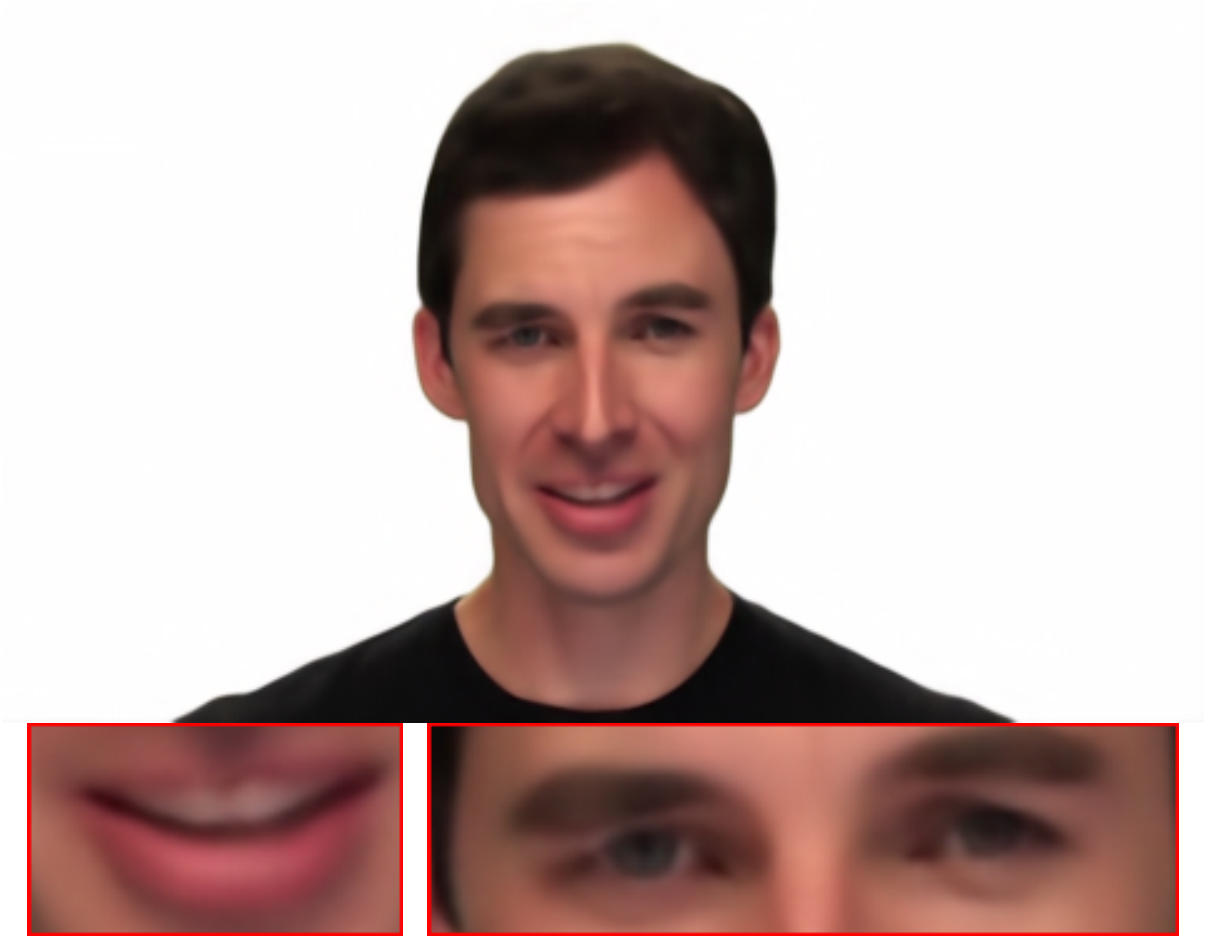}\vspace{1pt}
      \includegraphics[width=1\linewidth]{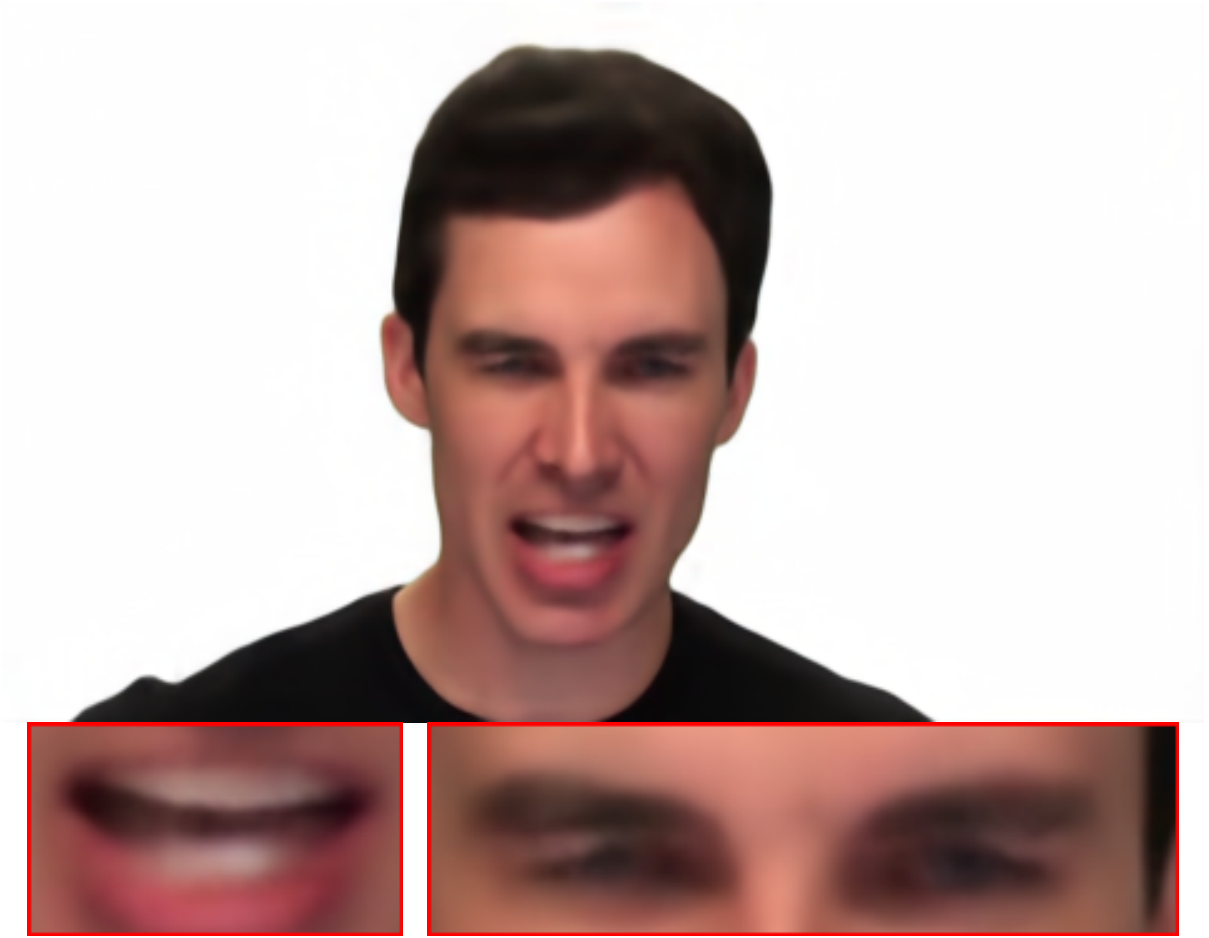}\vspace{1pt}
      \includegraphics[width=1\linewidth]{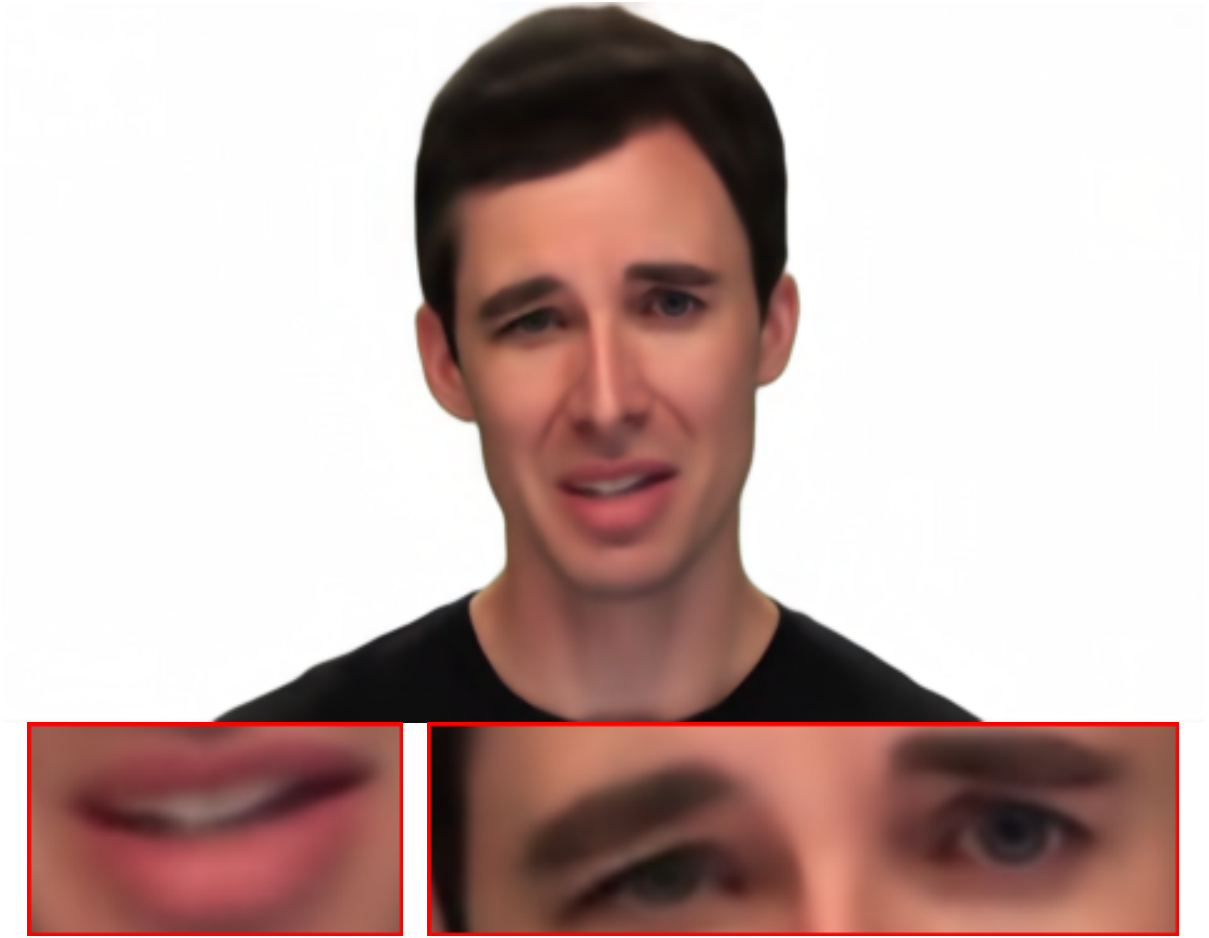}\vspace{1pt}
      \includegraphics[width=1\linewidth]{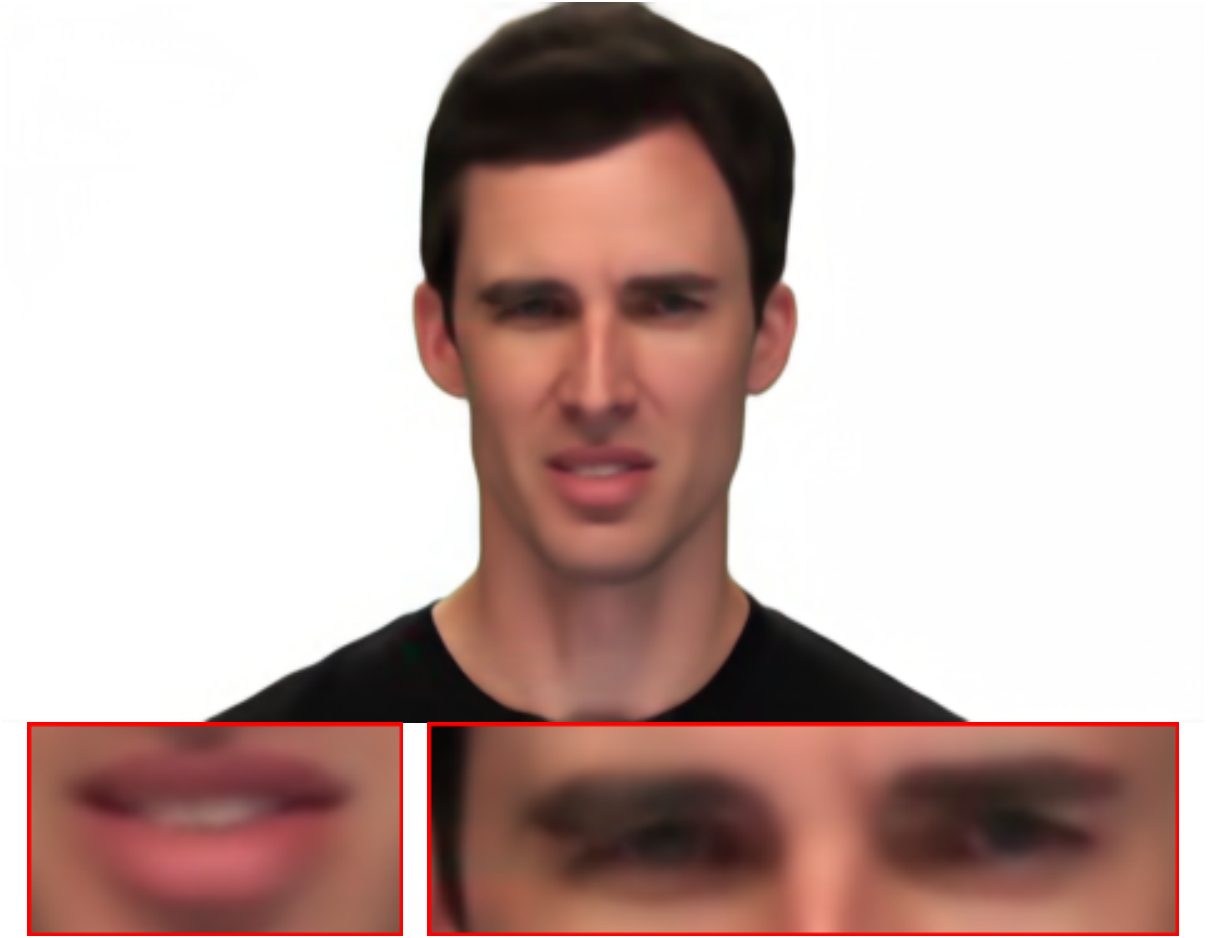}\vspace{1pt}
      \includegraphics[width=1\linewidth]{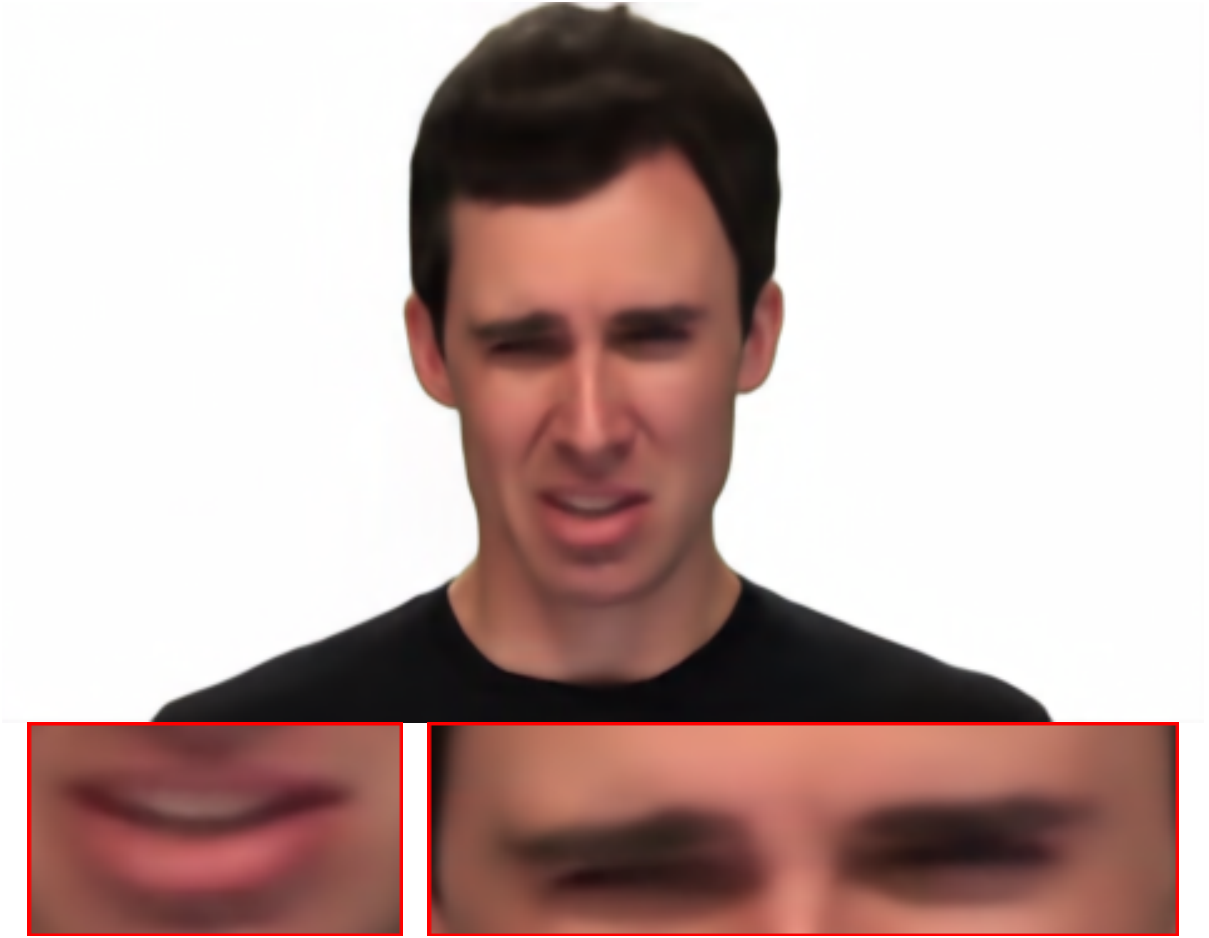}
    \end{minipage}
    }
  \subfigure[MMSD]{
    \begin{minipage}[t]{0.144\linewidth}
      \includegraphics[width=1\linewidth]{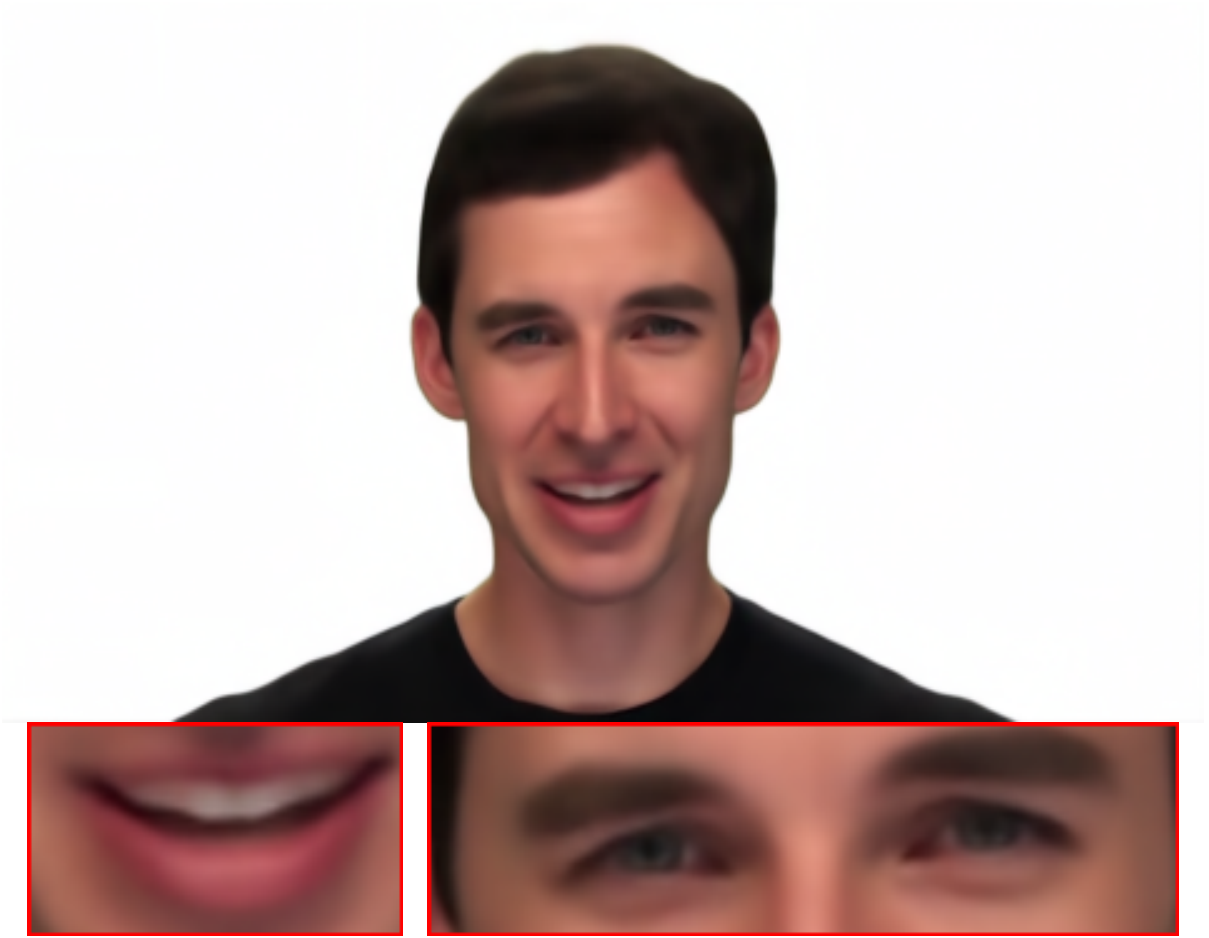}\vspace{1pt}
      \includegraphics[width=1\linewidth]{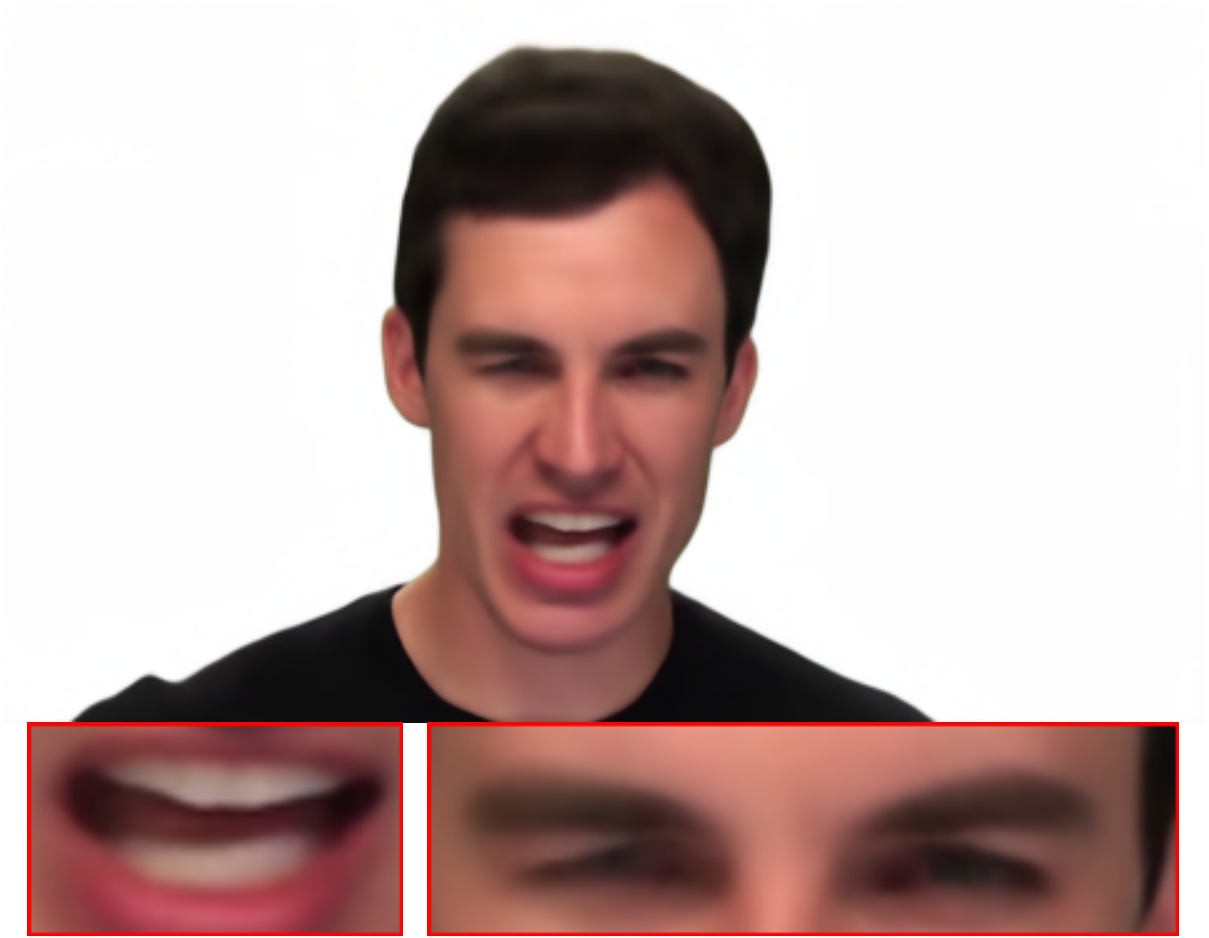}\vspace{1pt}
      \includegraphics[width=1\linewidth]{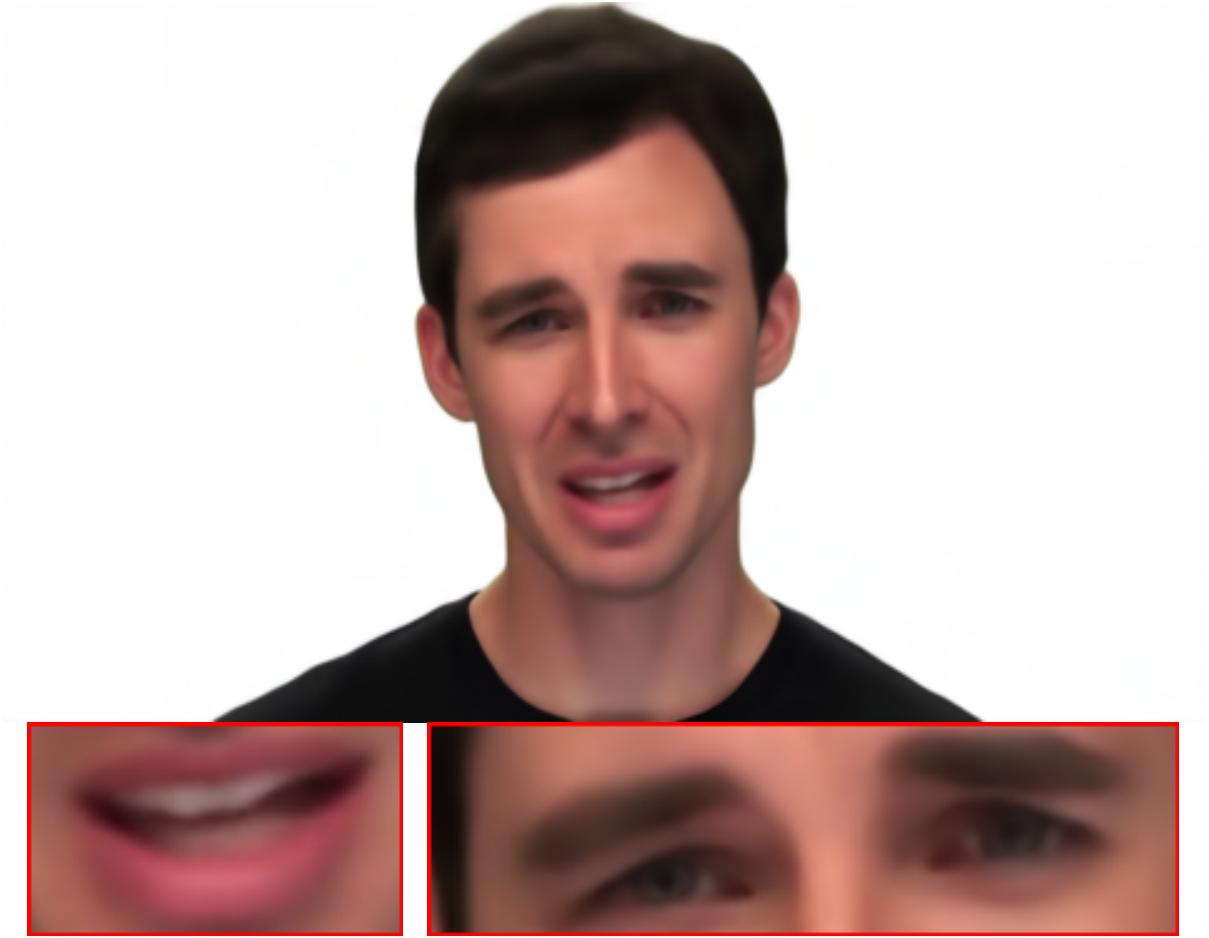}\vspace{1pt}
      \includegraphics[width=1\linewidth]{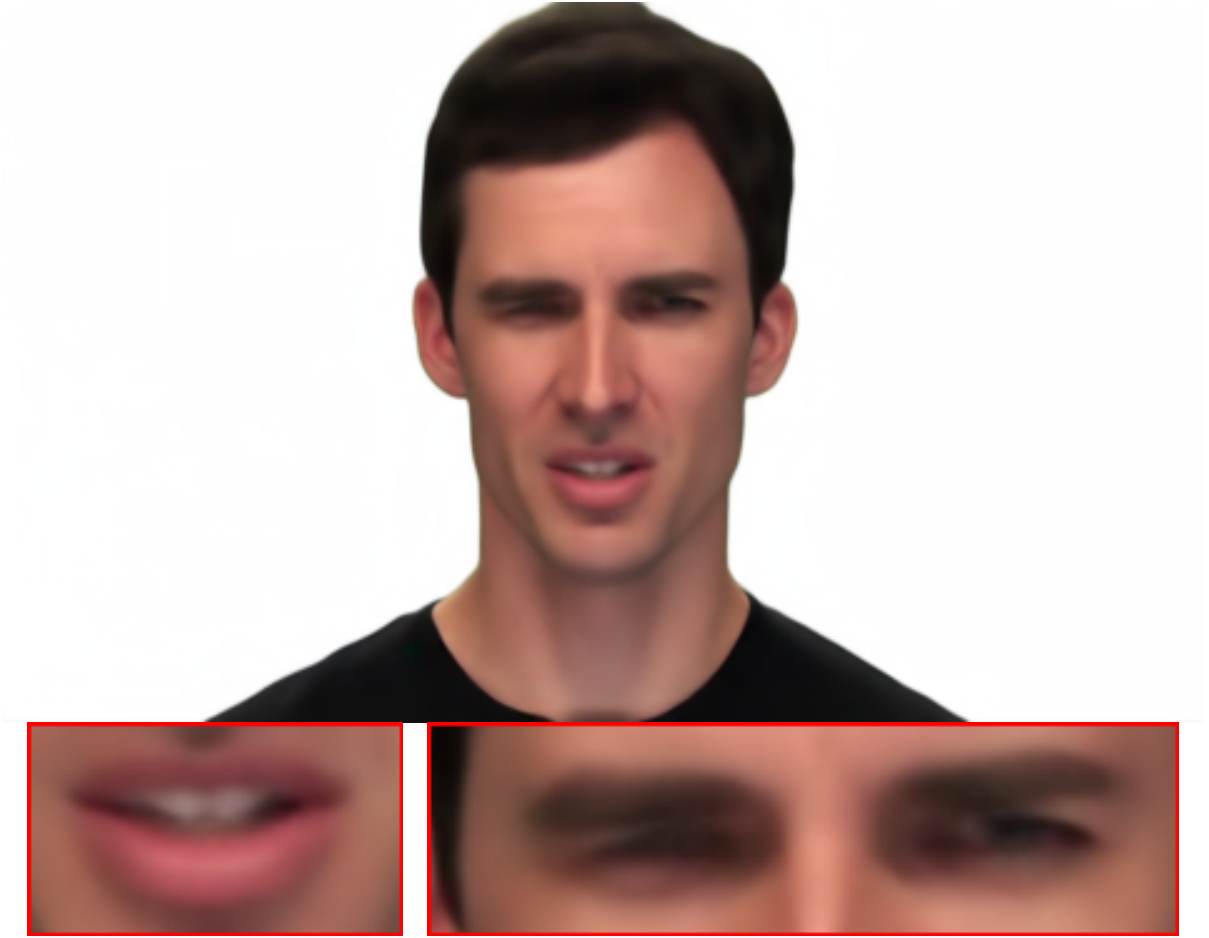}\vspace{1pt}
      \includegraphics[width=1\linewidth]{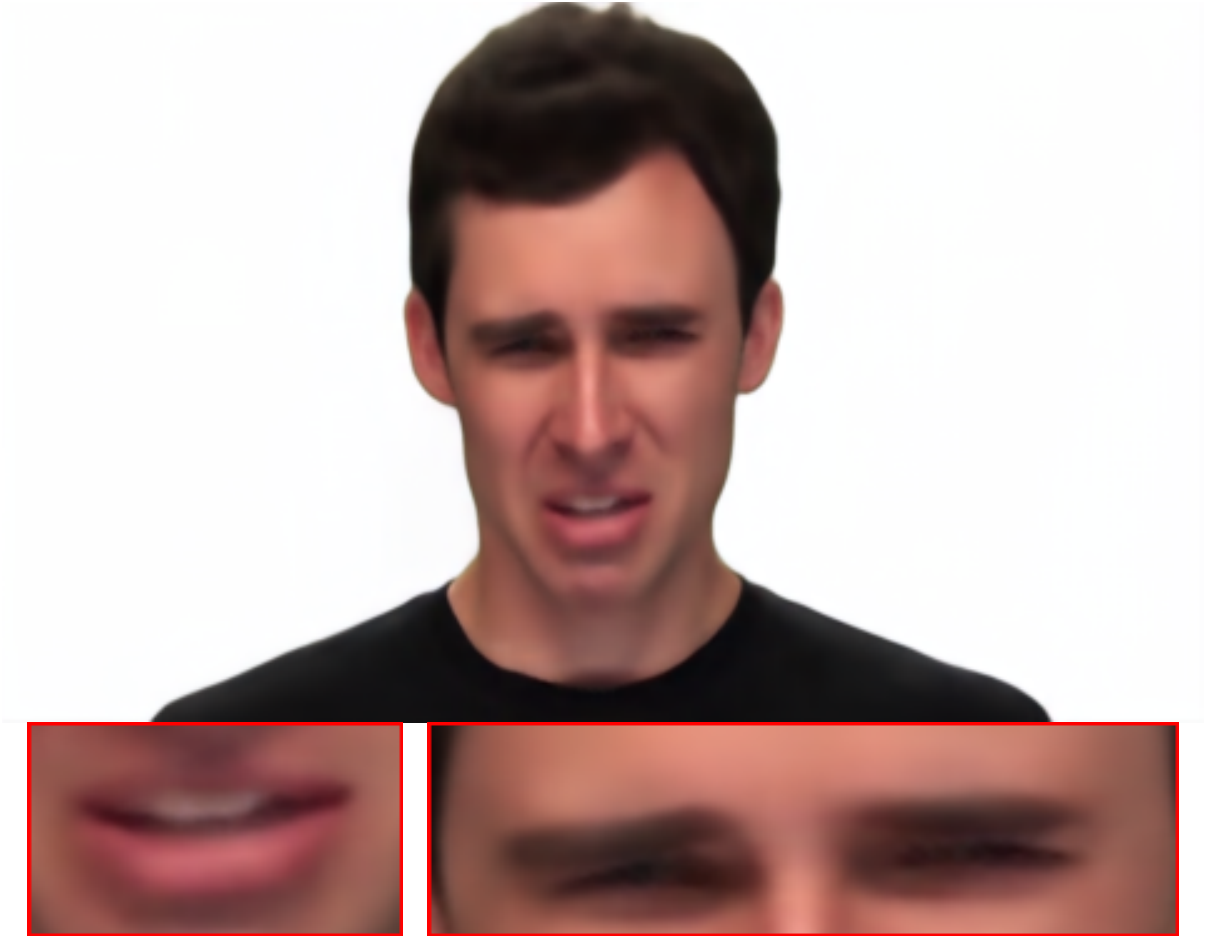}
    \end{minipage}
    \label{MMVR_L1_single}
    }
    \subfigure[MMSD (cGAN)]{
      \begin{minipage}[t]{0.144\linewidth}
        \includegraphics[width=1\linewidth]{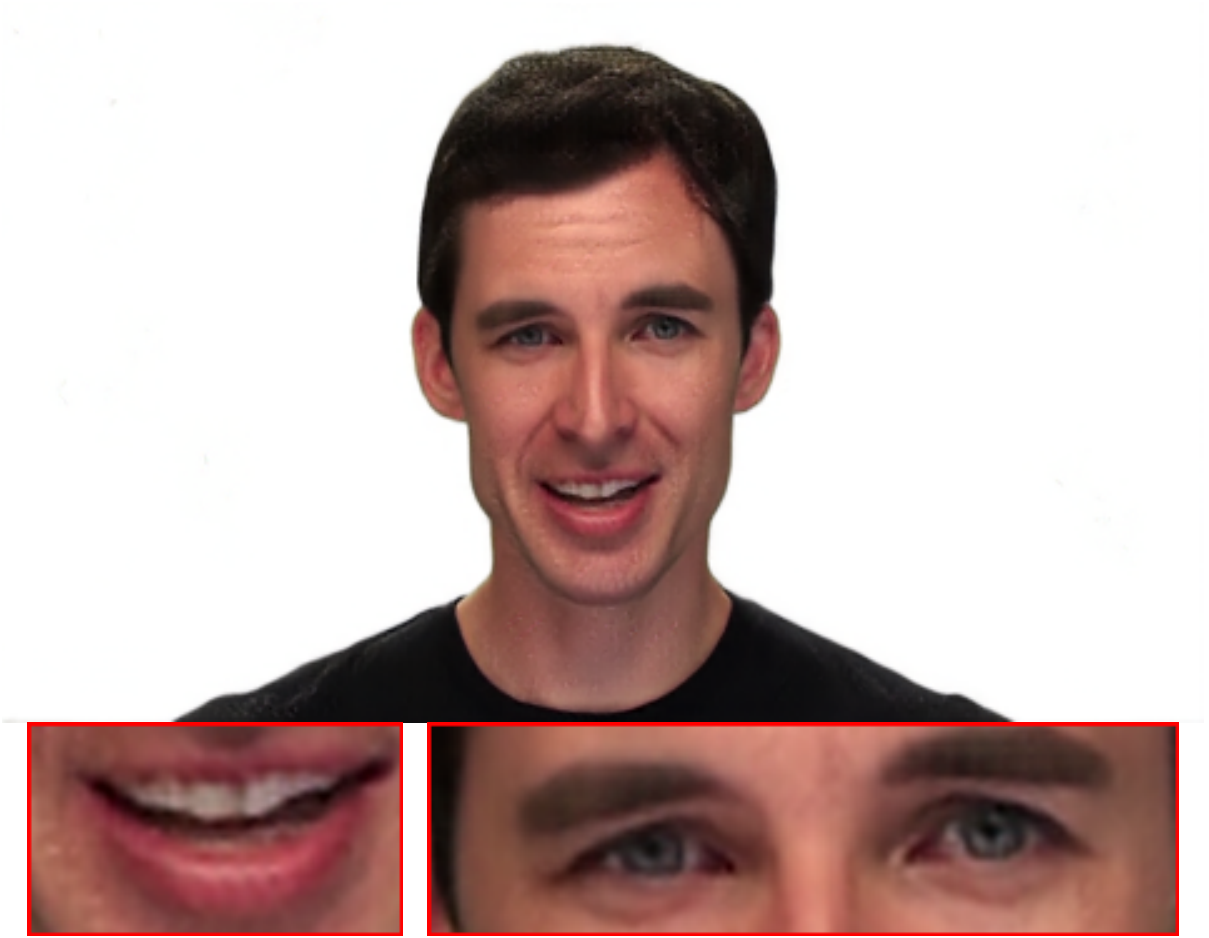}\vspace{1pt}
        \includegraphics[width=1\linewidth]{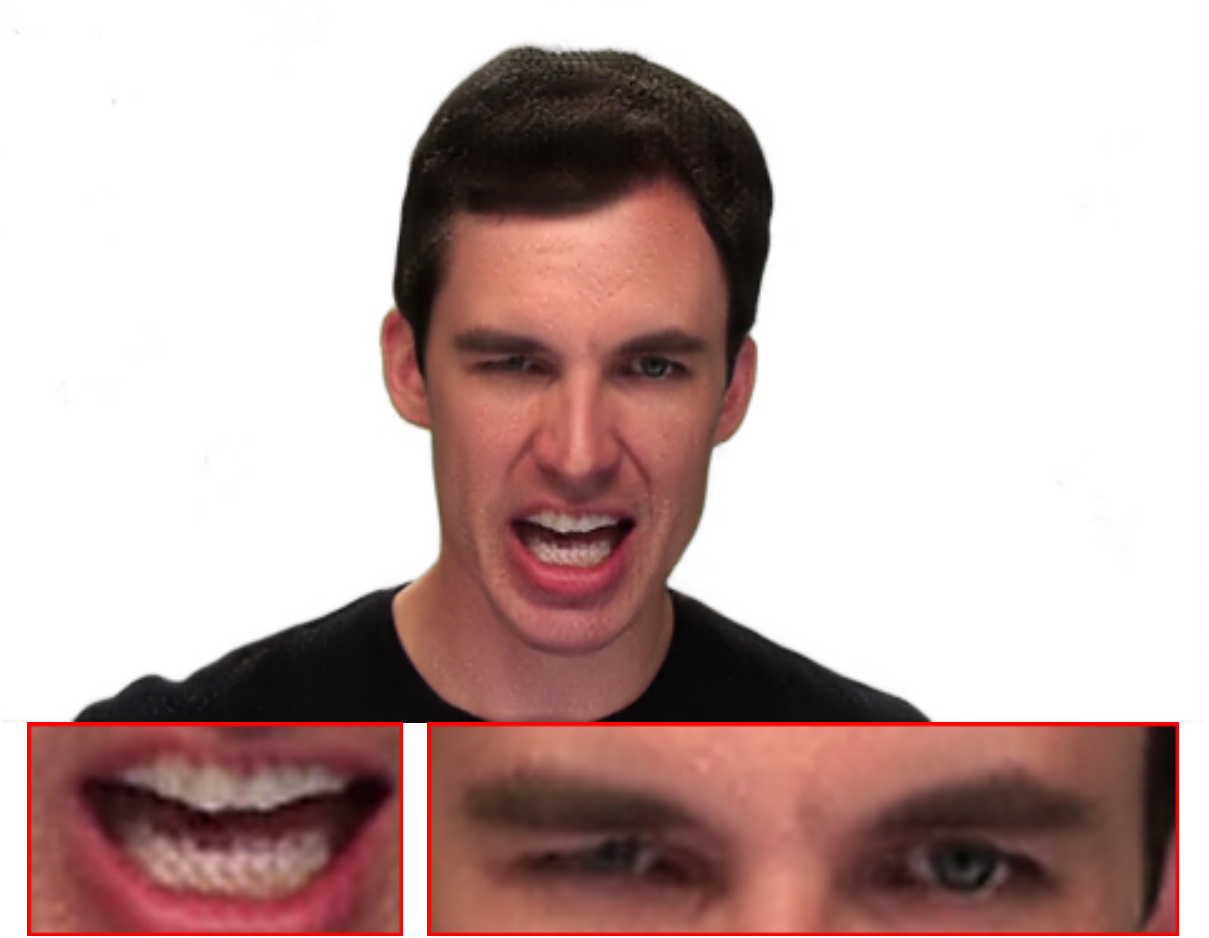}\vspace{1pt}
        \includegraphics[width=1\linewidth]{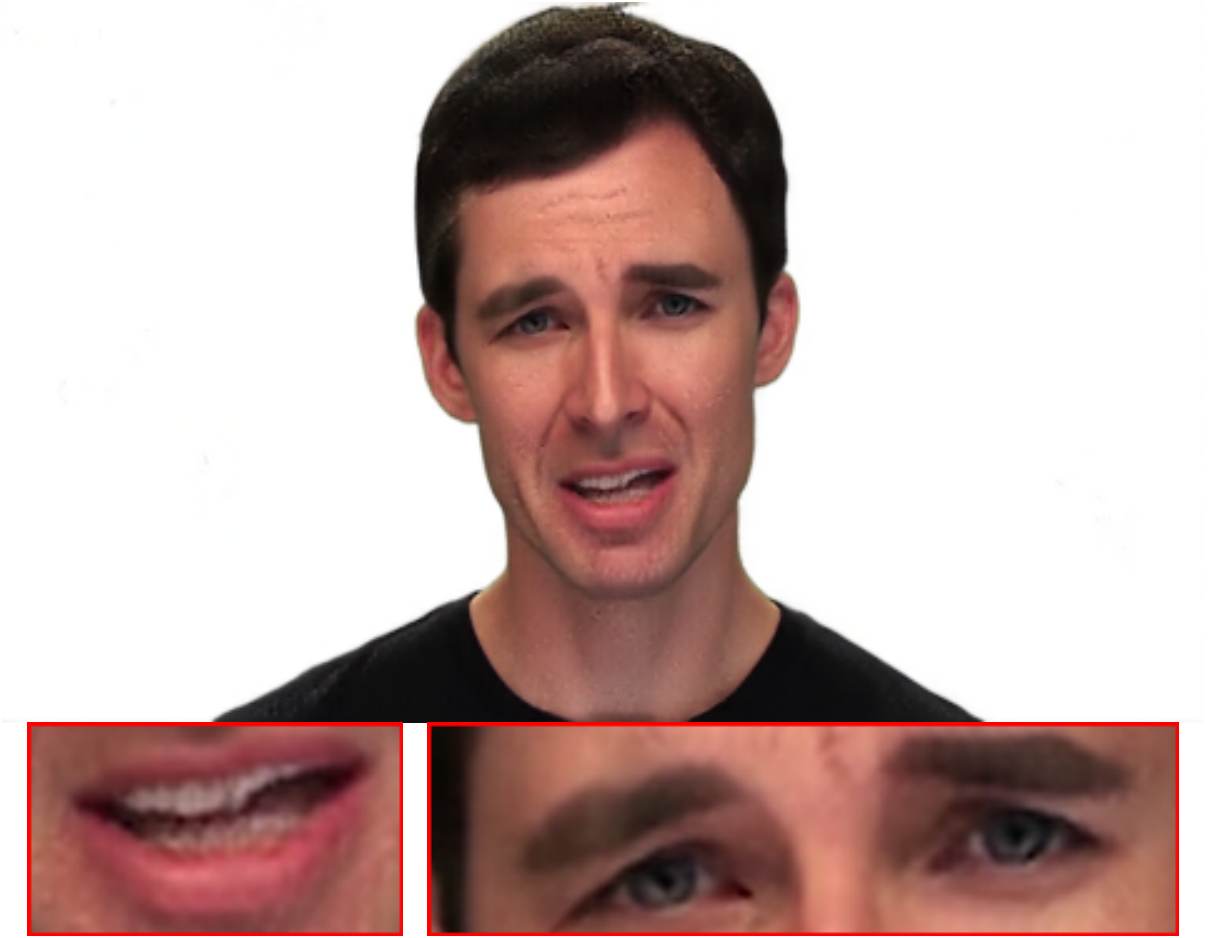}\vspace{1pt}
        \includegraphics[width=1\linewidth]{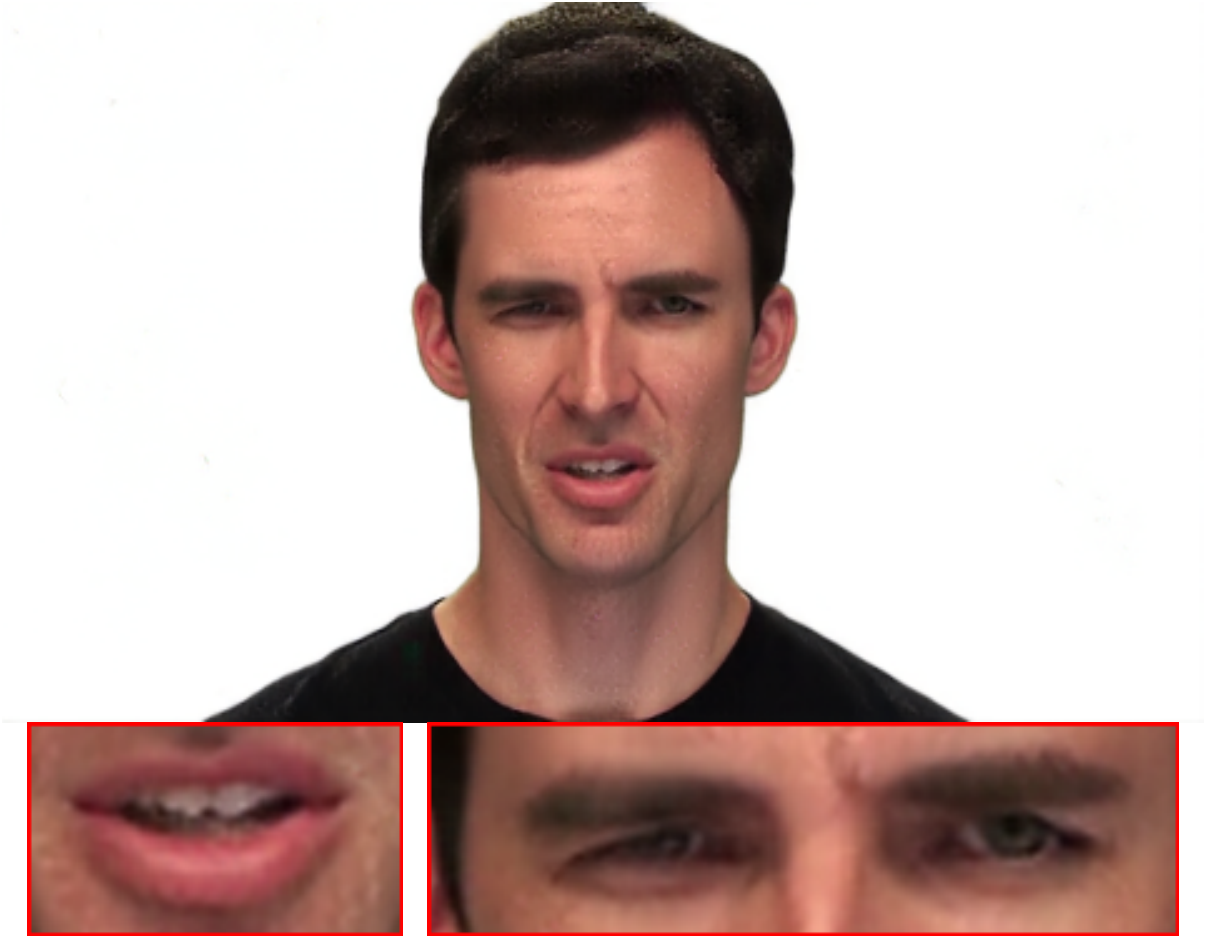}\vspace{1pt}
        \includegraphics[width=1\linewidth]{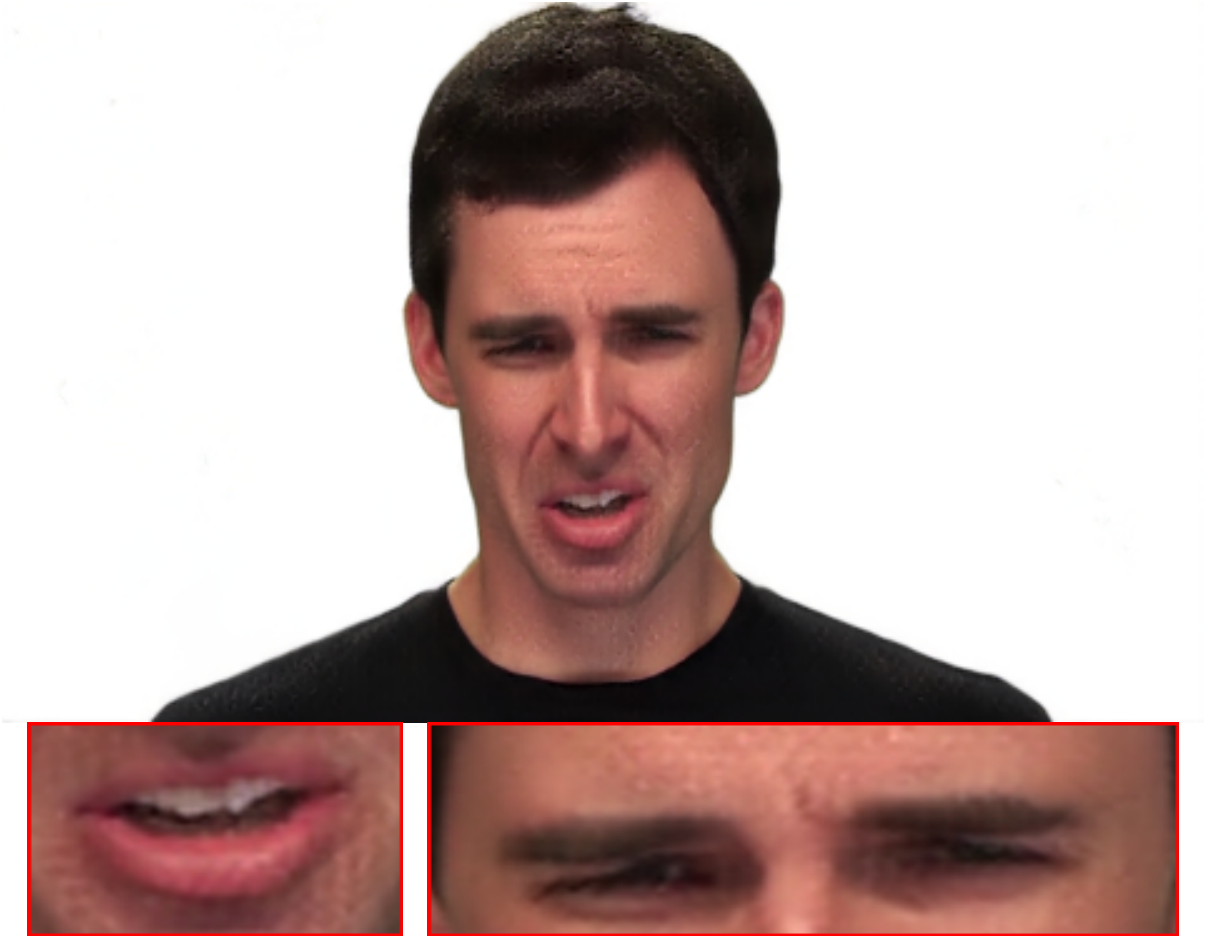}
      \end{minipage}
      }
    \subfigure[Ground Truth]{
      \begin{minipage}[t]{0.144\linewidth}
        \includegraphics[width=1\linewidth]{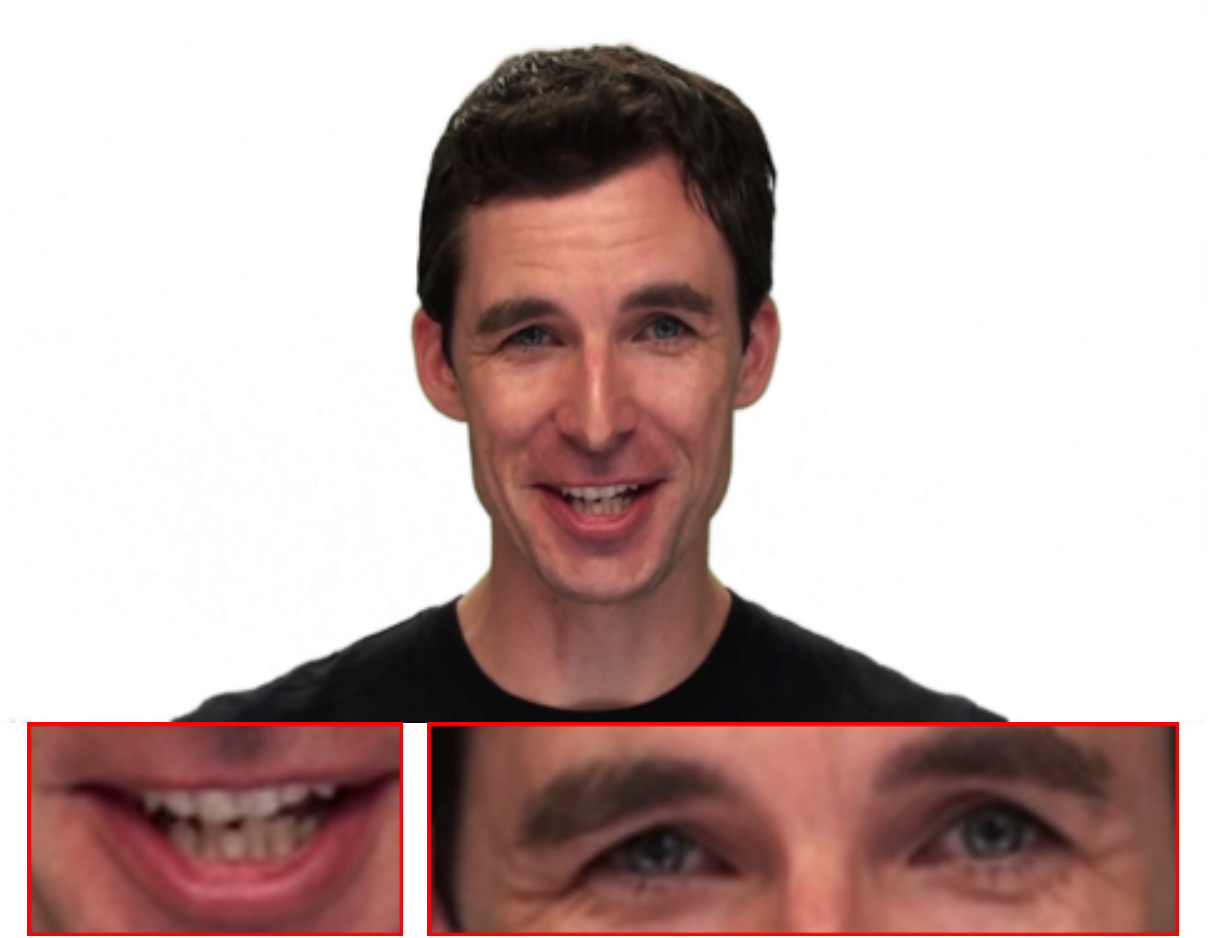}\vspace{1pt}
        \includegraphics[width=1\linewidth]{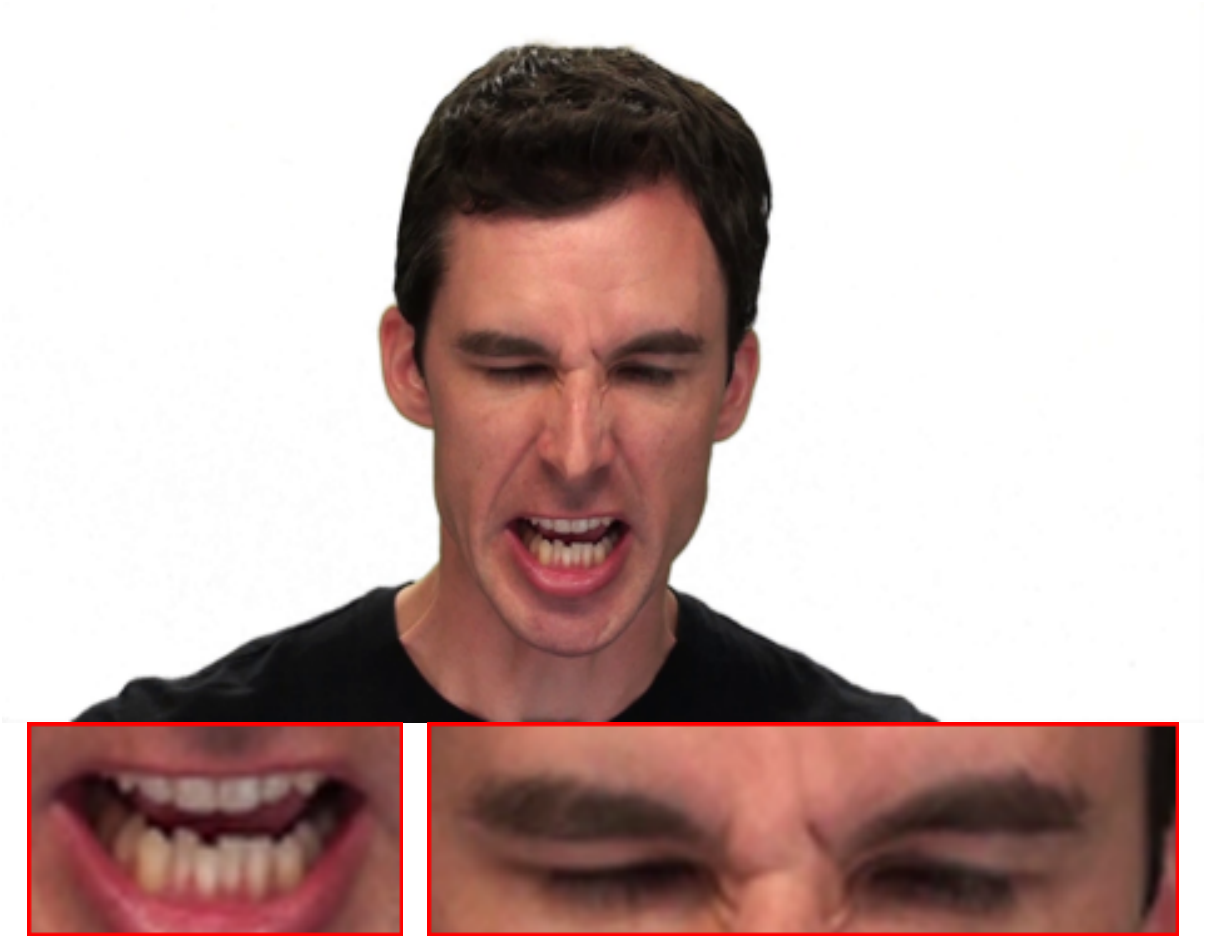}\vspace{1pt}
        \includegraphics[width=1\linewidth]{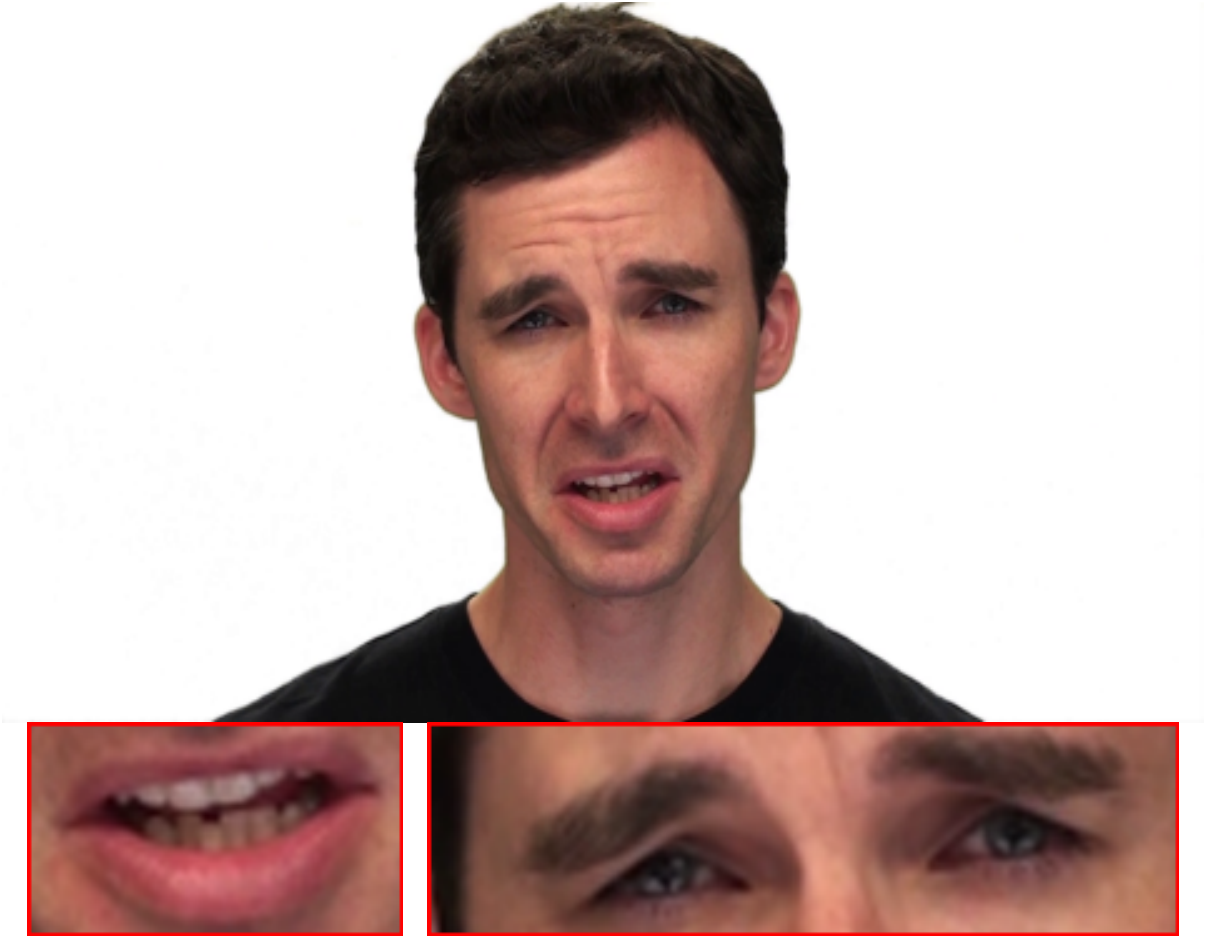}\vspace{1pt}
        \includegraphics[width=1\linewidth]{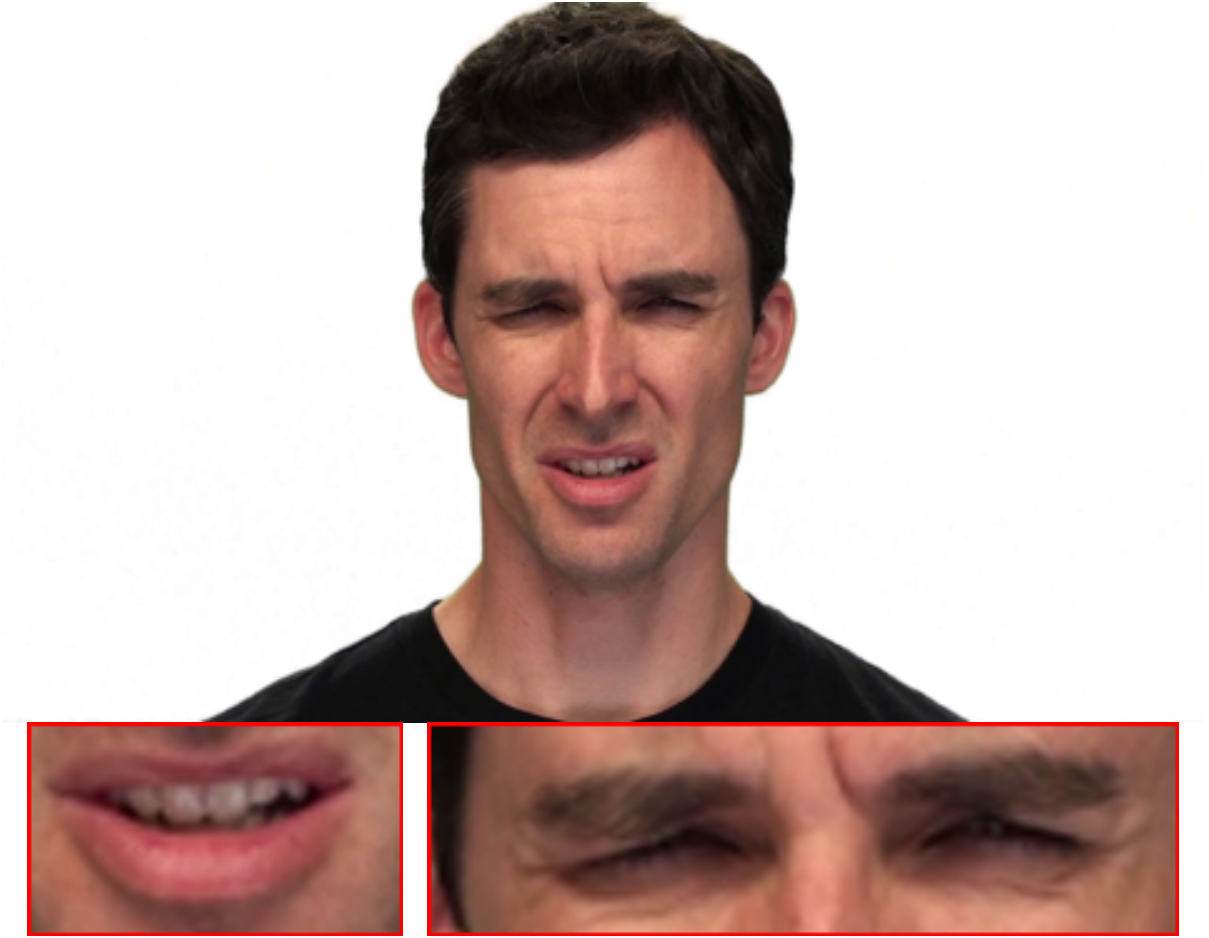}\vspace{1pt}
        \includegraphics[width=1\linewidth]{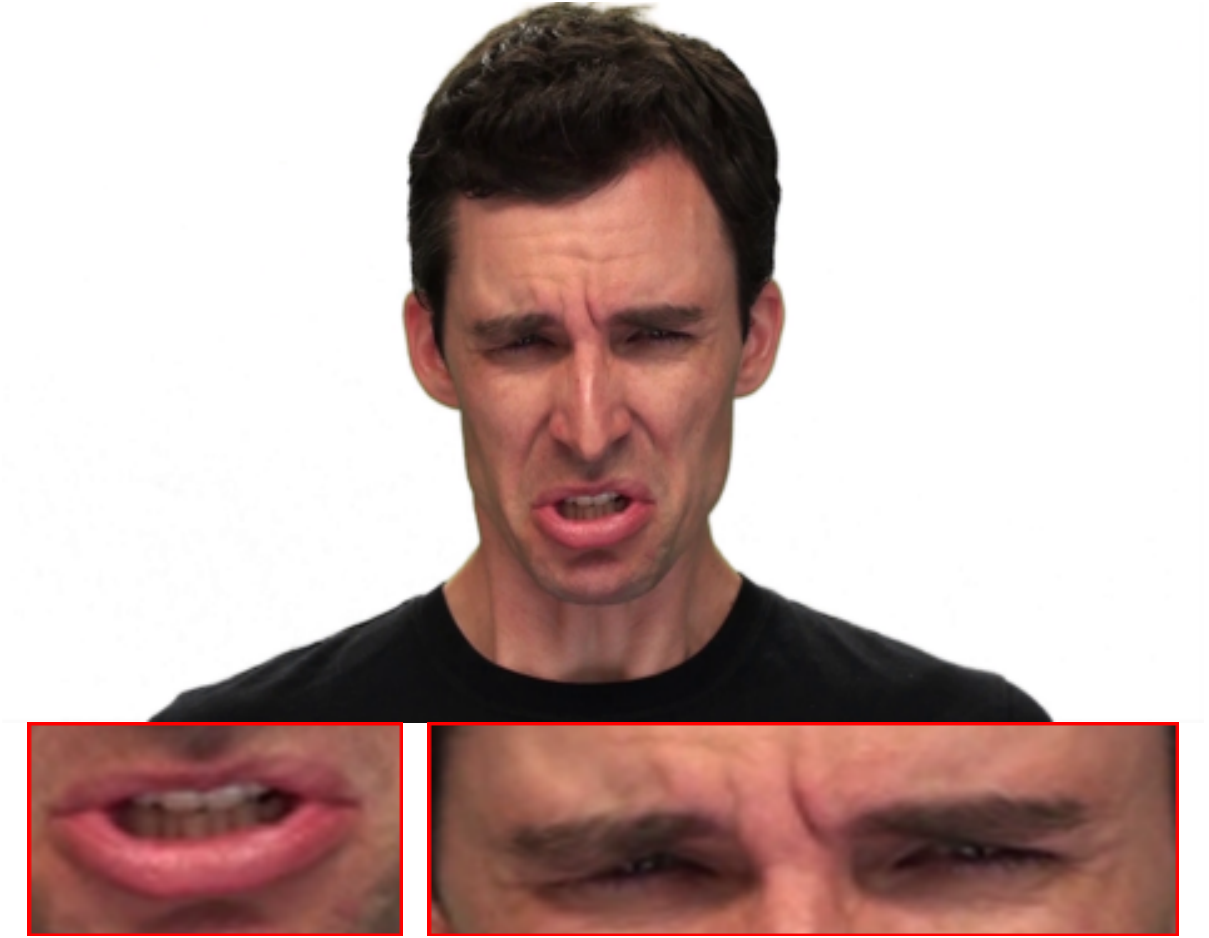}
      \end{minipage}
      }
  \caption{Restored results of different methods on a highly compressed video (50Kb/s, CRF=32, downsampling factor=4) of one speaker in different emotions.}
  \label{Single_Results_Comparison}
\end{figure*}

\subsection{Fusion of three modalities} \label{sec_egf}

In our design, the emotion is classified into eight types and two levels of intensity (strong and normal) as shown in Fig.\ref{figure_overall}.
All together there are 15 emotion states (the "neutral" emotion has no intensity level).
As the encoder has the high quality video of the talking head, it can accurately perform emotion recognition and transmit to the decoder the labels for emotion states.
The 15 emotion states are encoded by a 15-dimensional one-hot vector.

Having the emotion state label and video feature maps $\bm{f}_{V}$, the next task is to predict the face AU values that are emotion features.
This is done by a subnet called face AU generator, which is illustrated in Fig.\ref{figure_eaf}.
The AU generator takes the emotion vector $s$ and video features $\bm{f}_{V}$ as input.
It predicts face AU values by supervised learning in which the ground truth AU values $\bm{f}_E$ are obtained by facial expression analysis tool OpenFace \cite{Face_AUs,RAVDESS}.
The AU prediction subnet is trained to minimize the $\ell_2$ loss 
\begin{equation}
\mathnormal{L}_E = \left\|\bm{\hat{f}}_E - \bm{f}_E\right\|^2_2.
\end{equation}

Considering that video and audio signals vary their patterns in different emotion states, we pass the predicted emotion features $\bm{\hat{f}}_E$ through three full connected layers to produce a 64-dimensional channel attention vector
$\omega_{VA}$.  After being weighted by $\omega_{VA}$, the joint video-audio  features $\bm{f}_{V,A}$ are fused with the emotion state $s$, as shown in Fig.~\ref{figure_eaf}. This final fusion step combines the features of all three modalities can formulated as following:
\begin{equation}
\label{Eq_egf}
  \bm{f}_{V,A,E} = \mathit{Conv}\left(\small \left[\omega_{VA} \otimes \bm{f}_{V,A}, \; s  \small \right]\right)
\end{equation}
where $\otimes$ stands for the channel-wise multiplication, and $\mathit{Conv}$ is the fusion convolution using $1\times 1$ kernels.
The final combined features $\bm{f}_{V,A,E}$ are fed to the reconstruction module that restores the high quality frame $\hat{I}$.

\begin{figure*}[ht]
  \centering
\subfigure[Input]{
  \begin{minipage}[t]{0.05\linewidth}
      \includegraphics[width=1\linewidth]{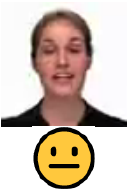}\vspace{30pt}
      \includegraphics[width=1\linewidth]{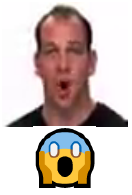}\vspace{35pt}
      \includegraphics[width=1\linewidth]{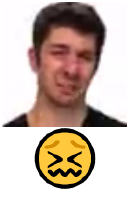}\vspace{35pt}
      \includegraphics[width=1\linewidth]{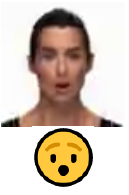}\vspace{35pt}
      \includegraphics[width=1\linewidth]{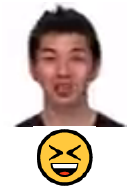}\vspace{34.5pt}
      \includegraphics[width=1\linewidth]{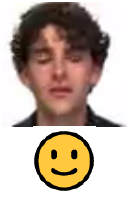}
  \end{minipage}
  }
  \subfigure[Bicubic]{
    \begin{minipage}[t]{0.14\linewidth}
      \includegraphics[width=1\linewidth]{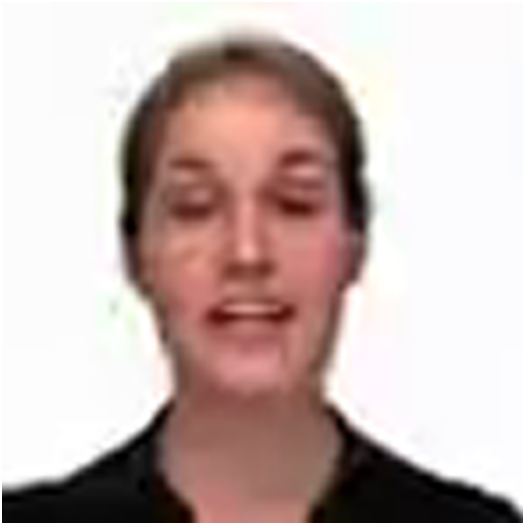}\vspace{1pt}
      \includegraphics[width=1\linewidth]{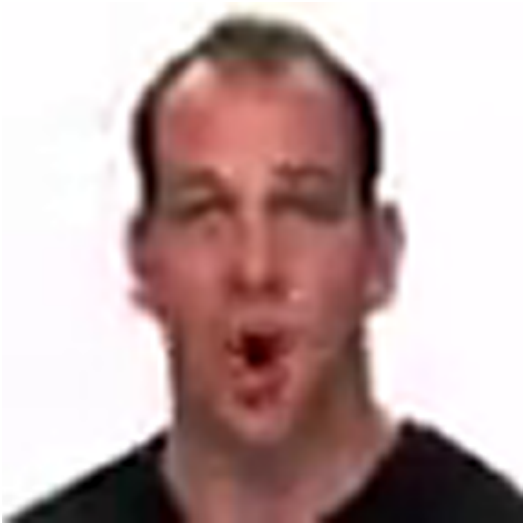}\vspace{1pt}
      \includegraphics[width=1\linewidth]{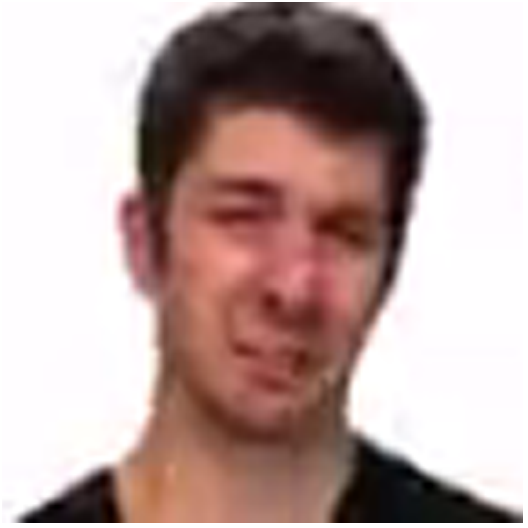}\vspace{1pt}
      \includegraphics[width=1\linewidth]{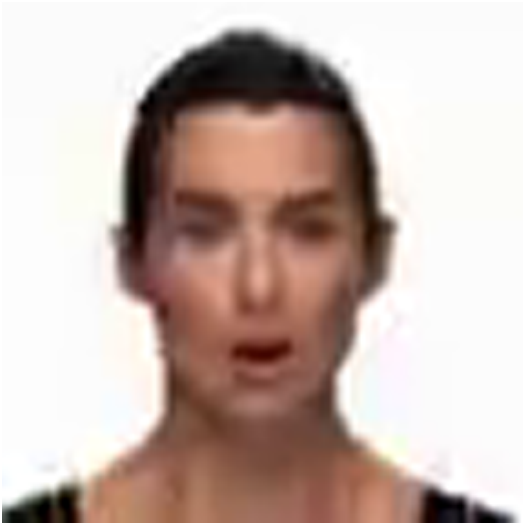}\vspace{1pt}
      \includegraphics[width=1\linewidth]{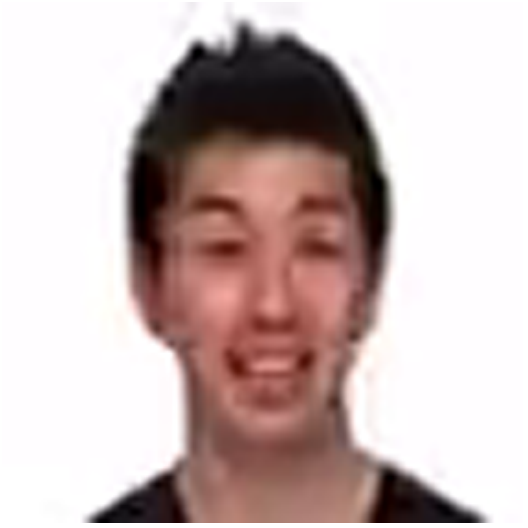}\vspace{1pt}
      \includegraphics[width=1\linewidth]{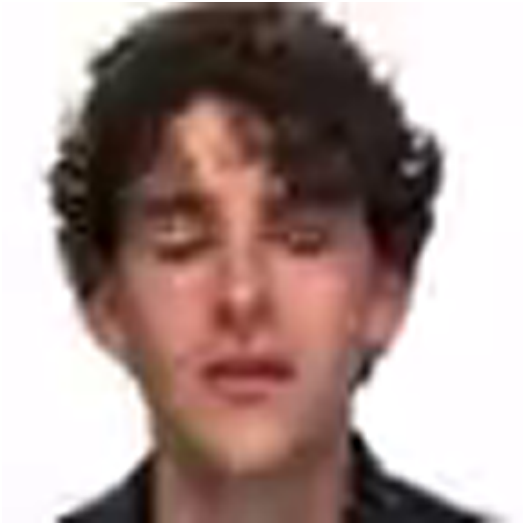}
    \end{minipage}
  }
  \subfigure[DBPN]{
    \begin{minipage}[t]{0.14\linewidth}
      \includegraphics[width=1\linewidth]{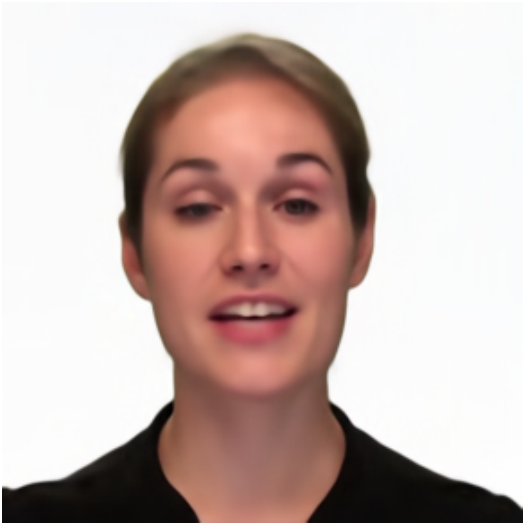}\vspace{1pt}
      \includegraphics[width=1\linewidth]{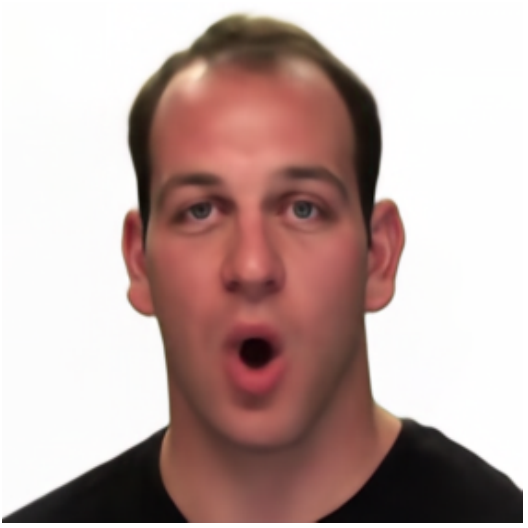}\vspace{1pt}
      \includegraphics[width=1\linewidth]{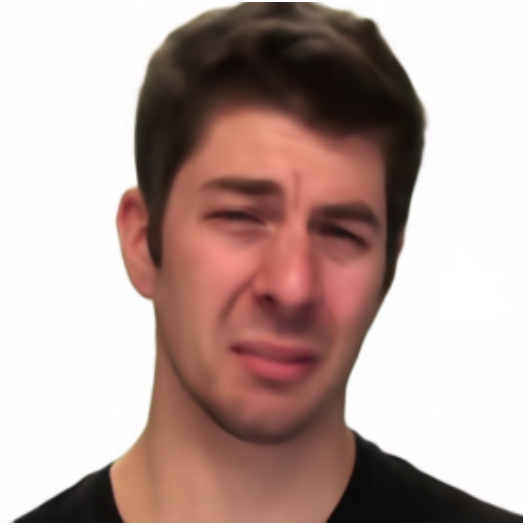}\vspace{1pt}
      \includegraphics[width=1\linewidth]{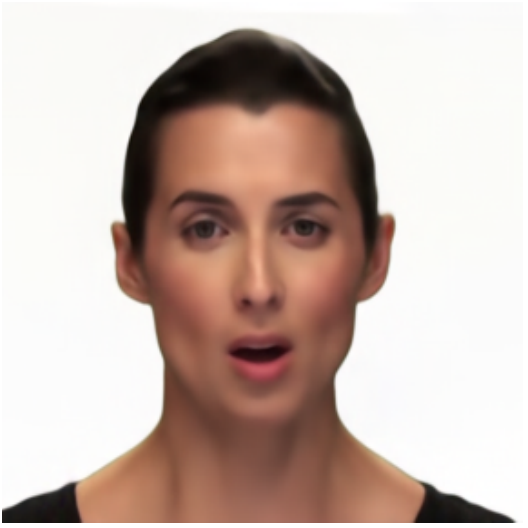}\vspace{1pt}
      \includegraphics[width=1\linewidth]{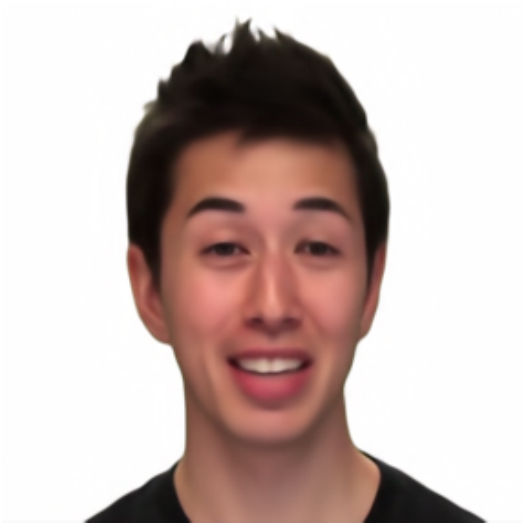}\vspace{1pt}
      \includegraphics[width=1\linewidth]{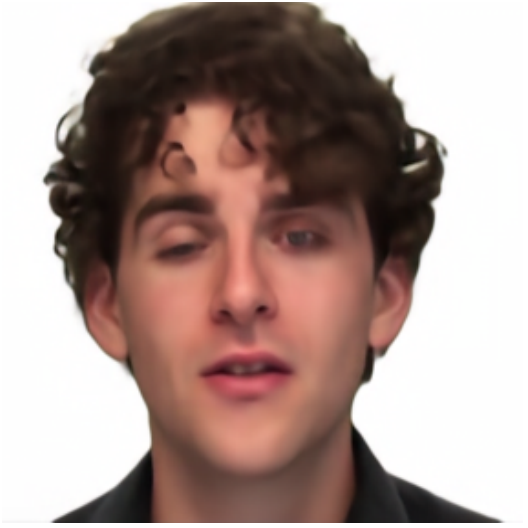}
    \end{minipage}
    }
  \subfigure[EDVR]{
    \begin{minipage}[t]{0.14\linewidth}
      \includegraphics[width=1\linewidth]{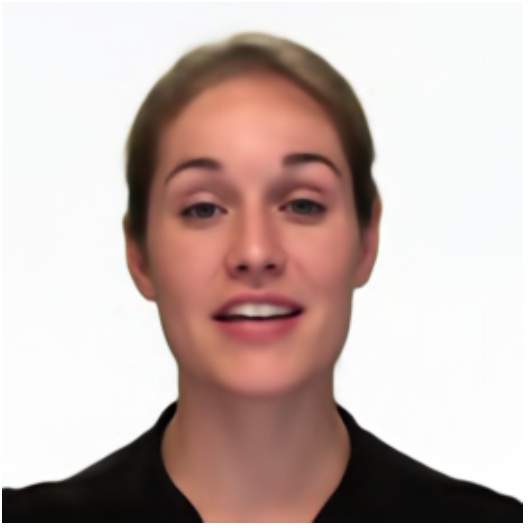}\vspace{1pt}
      \includegraphics[width=1\linewidth]{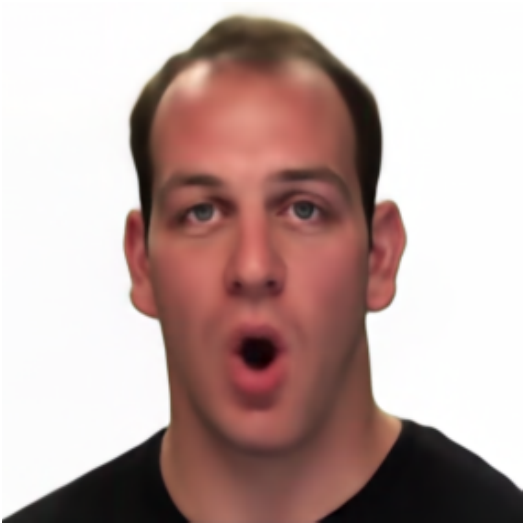}\vspace{1pt}
      \includegraphics[width=1\linewidth]{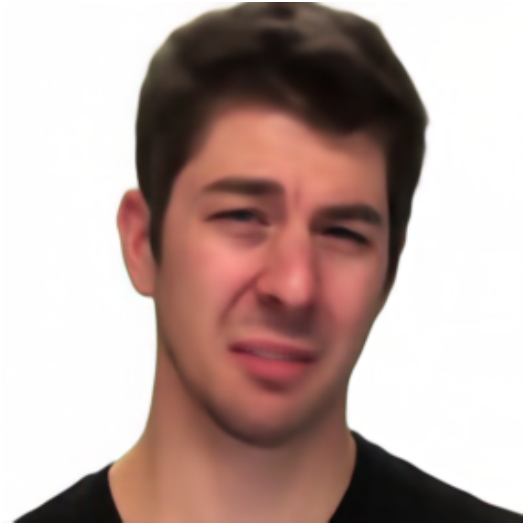}\vspace{1pt}
      \includegraphics[width=1\linewidth]{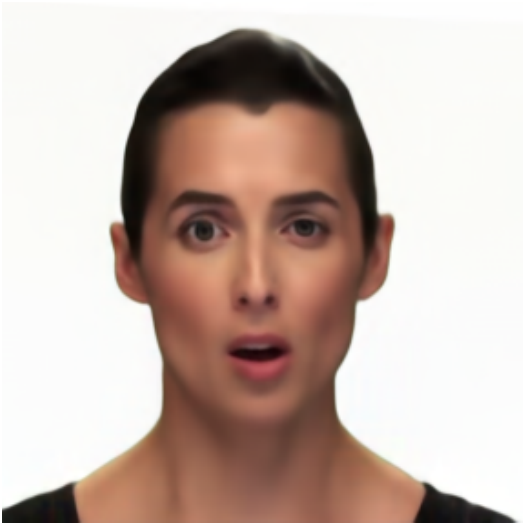}\vspace{1pt}
      \includegraphics[width=1\linewidth]{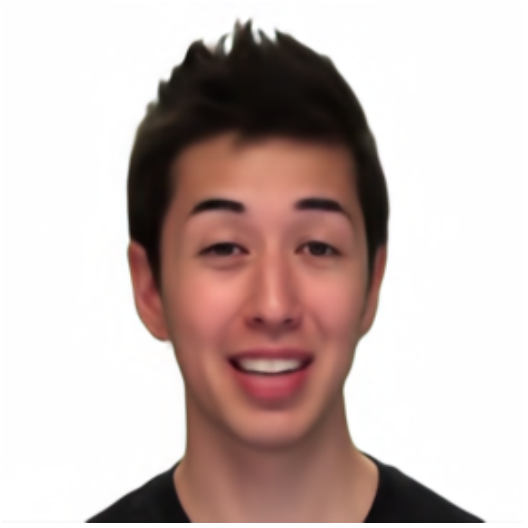}\vspace{1pt}
      \includegraphics[width=1\linewidth]{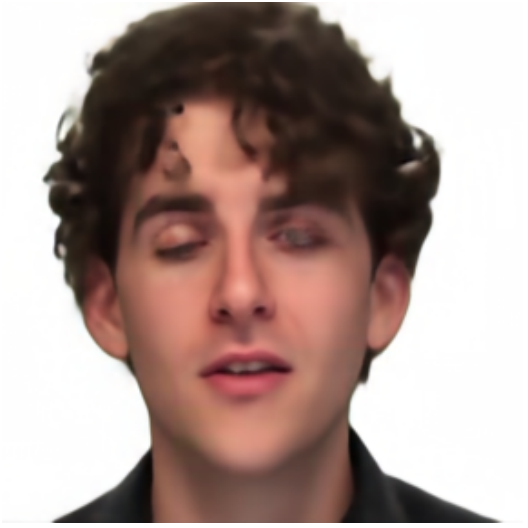}
    \end{minipage}
    }
  \subfigure[MMSD]{
    \begin{minipage}[t]{0.14\linewidth}
      \includegraphics[width=1\linewidth]{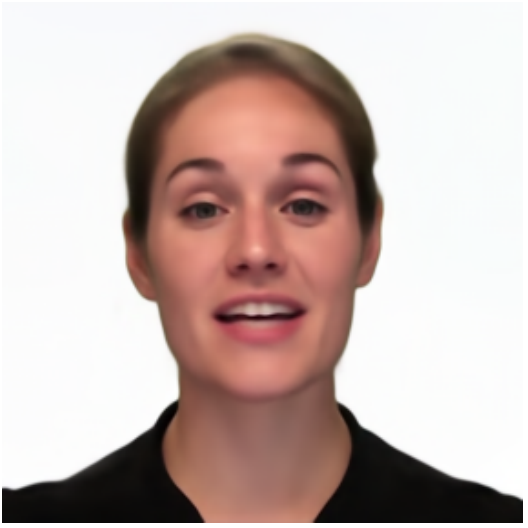}\vspace{1pt}
      \includegraphics[width=1\linewidth]{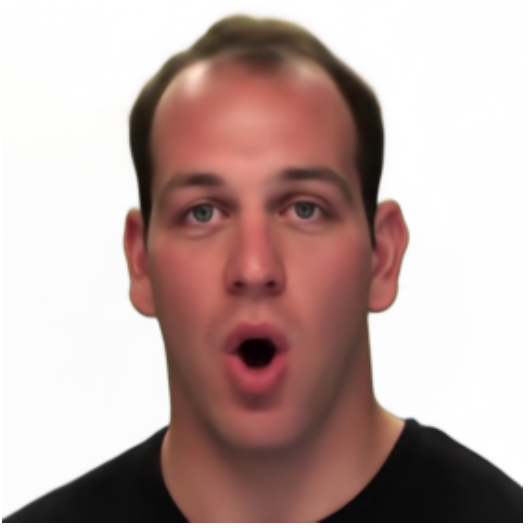}\vspace{1pt}
      \includegraphics[width=1\linewidth]{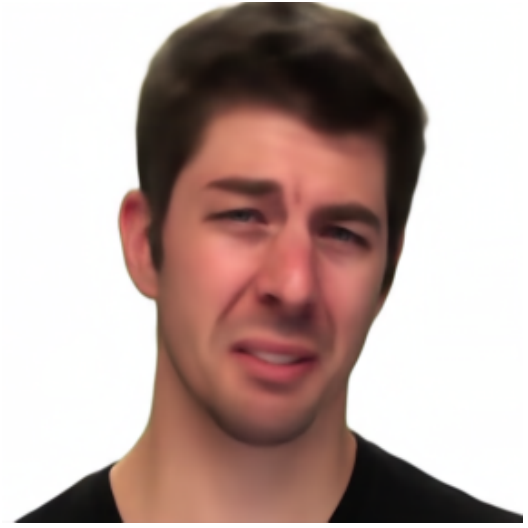}\vspace{1pt}
      \includegraphics[width=1\linewidth]{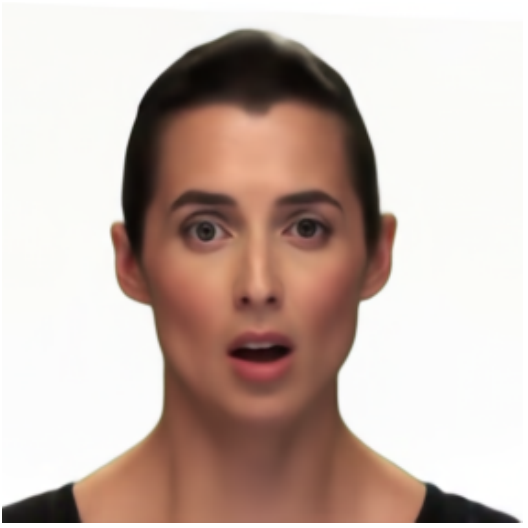}\vspace{1pt}
      \includegraphics[width=1\linewidth]{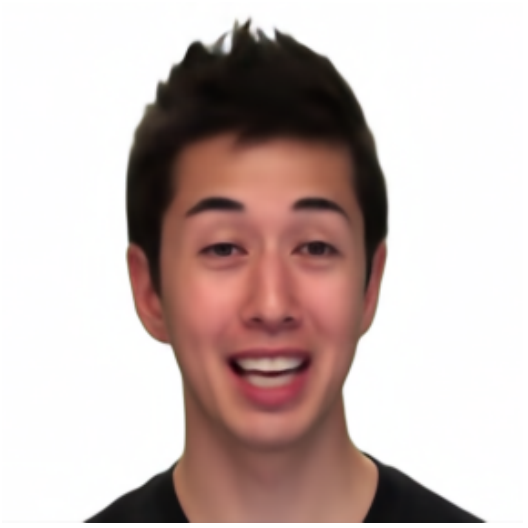}\vspace{1pt}
      \includegraphics[width=1\linewidth]{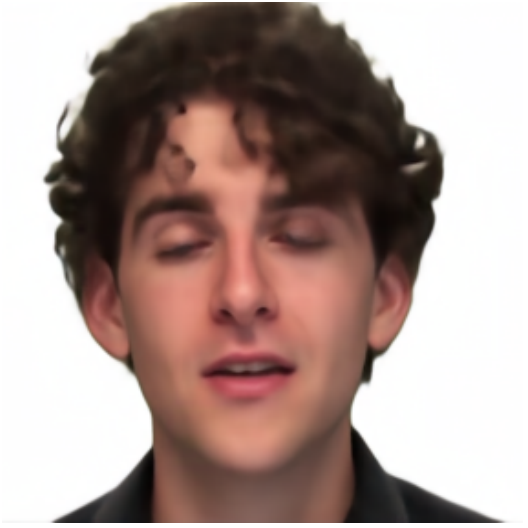}
    \end{minipage}
    \label{MMVR_L1_multi}
    }
    \subfigure[MMSD (cGAN)]{
      \begin{minipage}[t]{0.14\linewidth}
        \includegraphics[width=1\linewidth]{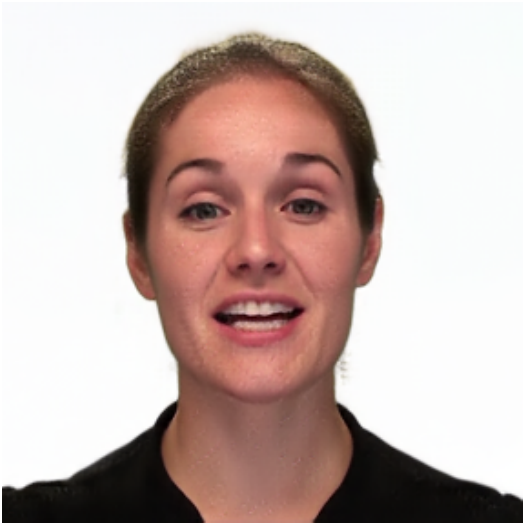}\vspace{1pt}
        \includegraphics[width=1\linewidth]{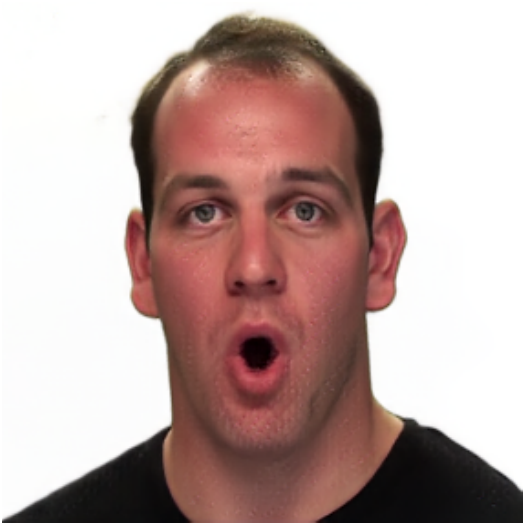}\vspace{1pt}
        \includegraphics[width=1\linewidth]{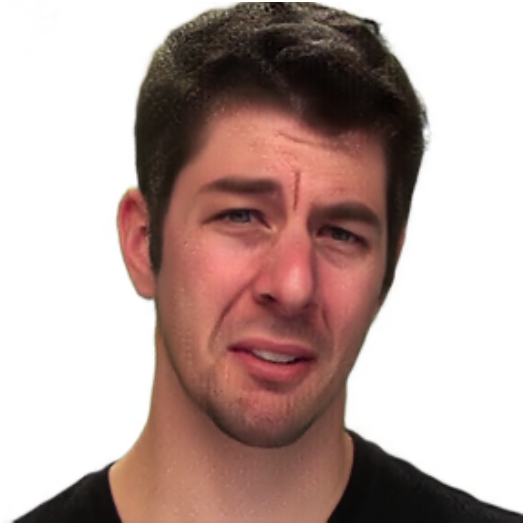}\vspace{1pt}
        \includegraphics[width=1\linewidth]{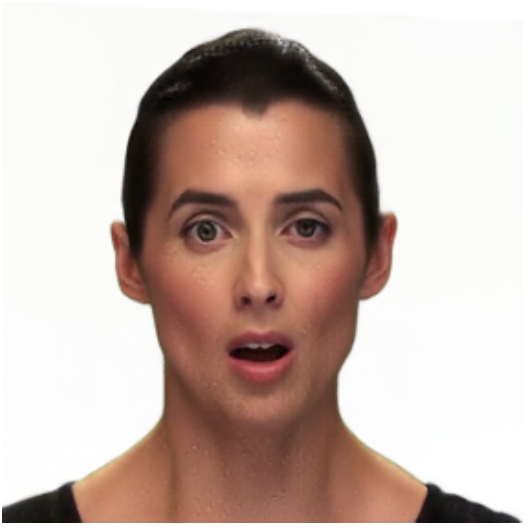}\vspace{1pt}
        \includegraphics[width=1\linewidth]{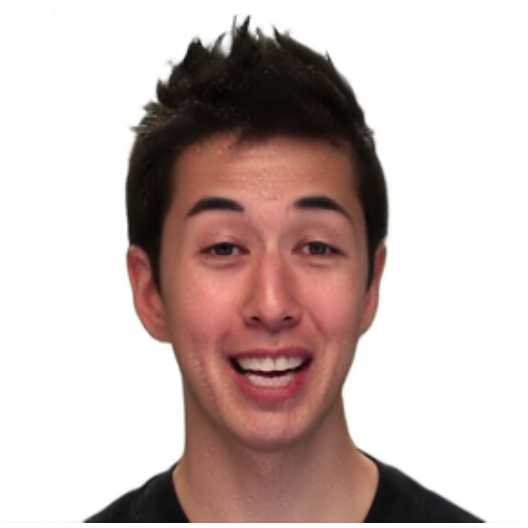}\vspace{1pt}
        \includegraphics[width=1\linewidth]{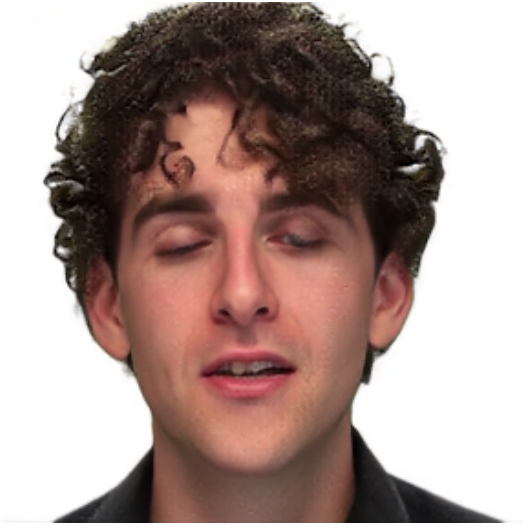}
      \end{minipage}
      }
    \subfigure[Ground Truth]{
      \begin{minipage}[t]{0.14\linewidth}
        \includegraphics[width=1\linewidth]{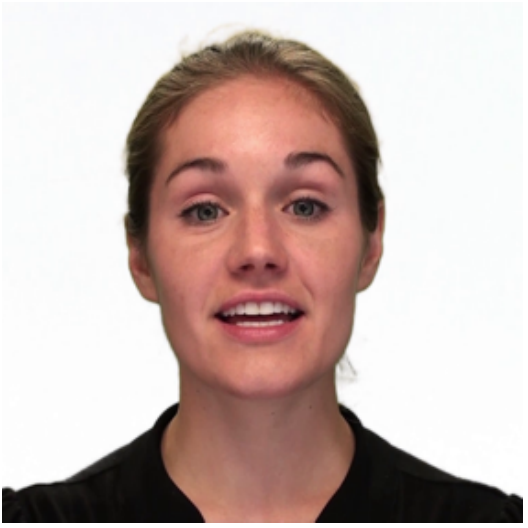}\vspace{1pt}
        \includegraphics[width=1\linewidth]{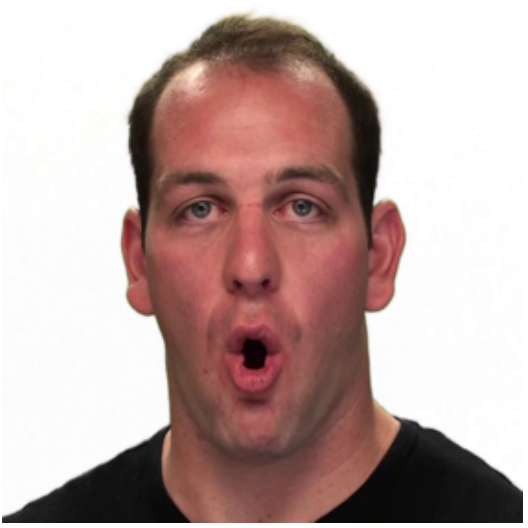}\vspace{1pt}
        \includegraphics[width=1\linewidth]{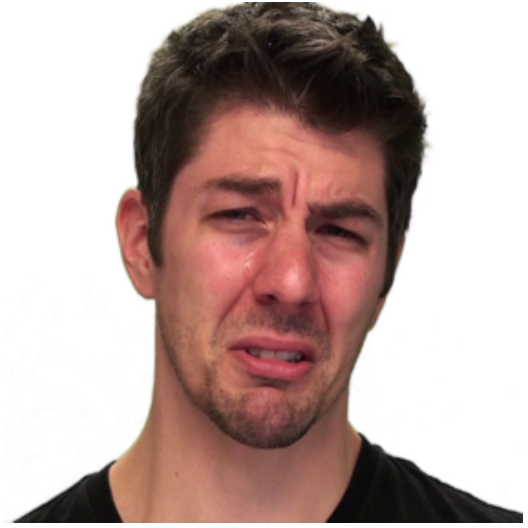}\vspace{1pt}
        \includegraphics[width=1\linewidth]{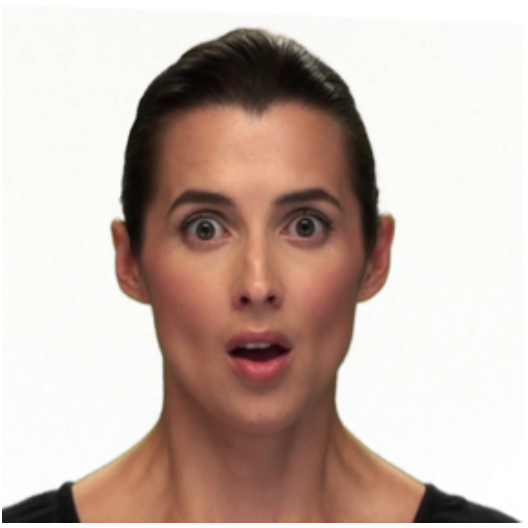}\vspace{1pt}
        \includegraphics[width=1\linewidth]{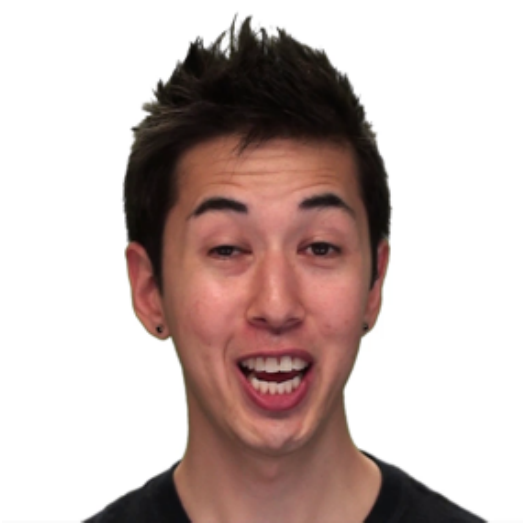}\vspace{1pt}
        \includegraphics[width=1\linewidth]{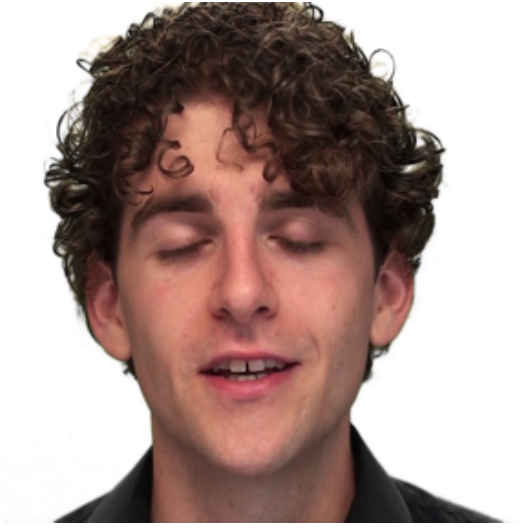}
      \end{minipage}
      }
  \caption{More results of different methods on highly compressed talking head videos (50Kb/s, CRF=32, downsampling factor=4) of different persons in varied emotions.}
  \label{Multi_Results_Comparison}
\end{figure*}

\subsection{Emotion-guided conditional GAN}
One more way of benefiting from the emotion side information in low bit rate video restoration is to add a discrimination loss of adversary neural network (GAN) \cite{he2018tip}. There are two justifications for using GAN here.  First, emotions shape facial expressions and thus the face image;
second, perceptual quality of talking head video largely depends on accurate depiction of facial expressions.
We adopt the conditional GAN (cGAN) technique \cite{perarnau2016invertible,lu2018attribute} to incorporate the prior knowledge of emotion state $s$ into statistical inference.

Specifically in our implementation, the added cGAN subnet forces the restored image $\hat{I}$ of MMSD-Net to pass the test of being the original face image $I_t$ in the given emotion state $s$  without compression or down sampling.
The test is done by examining whether $\hat{I}$ and $I$ obey the same conditional probability distribution of face images in the given emotion state $s$.

The discriminator $D$, with the same architecture as \cite{perarnau2016invertible},
takes the restored image $\hat{I}$ and given emotion state $s$ as input and outputs a single scalar $D(\hat{I},s)$ representing the probability that $\hat{I}$ came from real images with given emotion state $s$.
The MMSD-Net is trained to maximize $D(\hat{I},s)$ and discriminator is trained to minimize $D(\hat{I},s)$, simultaneously.
The adversarial loss function is defined as:
\begin{equation}
  \mathnormal{L}_{adv} = -\mathbb{E}_x\left[log\left(\mathnormal{D}(\mathnormal{G}(\check{I},s),s)\right)\right]
\end{equation}
where $\mathnormal{G}$ represents the proposed MMSD neural network and $\mathnormal{D}$ is the discriminator.

Combining all the loss terms introduced above, the overall objective function for optimizing the MMSD-Net is
\begin{equation}
L = \| \hat{I} - I\|_1 + \lambda_1 L_{adv} + \lambda_2 L_E
\end{equation}
where $\lambda_1$ and $\lambda_2$ are hyper-parameters.  Recall from the discussions around Eq(\ref{Eq_egf}) that the restored frame $\hat{I}$ is an inference based on the three-modality features $\bm{f}_{V,A,E}$.



\section{Experiments}\label{Experiments}

We have conducted extensive experiments to evaluate the performance of the proposed MMSD-Net in comparison with existing methods.  The setup and results of the experiments are reported in this section.

\subsection{Data preparation and training details}


The video data for our experiments are from the RAVDESS dataset \cite{RAVDESS}.
It includes 3628 talking head videos, each of which is about 3 seconds long. In the video,
the person speaks or sings with a specific emotion.  This dataset is split into two parts: 2540 videos
for training the MMSD-Net and 1088 videos for validation.  To generate paired training video data,
we first down sample the original video by a factor of 4 to $120\times 72$ resolution and compress the down-sampled videos using
FFmpeg with x264 video codec of which the coding parameters are: CRF=15, CRF=32, and CRF=40.
The emotion states of the talking head videos are provided by the RAVDESS dataset.

In our implementation,
all modules of the MMSD-Net are trained end-to-end.  In the training process we set the mini-batch size to 8, $\lambda_1 = 0.01$, $\lambda_2=0.001$.
To make training more stable and converge faster, we optimize the MMSD-Net model without adversarial loss $L_{adv}$ in the first two epochs, and then include the term $L_{adv}$ in the total loss for subsequent epochs.

We choose Adam optimizer \cite{kingma2014adam} to train the MMSD-Net by setting $\beta_1=0.5$ and $\beta_2=0.9$ with initializing learning rate $10^{-4}$.

\subsection{Results and comparison study}
We compare the results of the MMSD-Net with two state-of-the-art video restoration methods: DBPN \cite{DBPN} and EDVR \cite{EDVR}.  EDVR is claimed to be a unified framework to suit various video restoration tasks.
Their pre-trained models are based on a video super-resolution dataset, our model trained with our dataset performs much better than the pre-trained DBPN and EDVR models. For fair comparison, we use our dataset to retrain
the DBPN model and the EDVR model with their default training parameters.

We implement and test the proposed MMSD-Net with and without adversarial loss. Table \ref{comparasion} reports the PSNR and SSIM results of competing video restoration methods for different compression quality
factor CRF, in which the MMSD-Net is optimized without the adversarial loss of cGAN. The numbers in the table are averages over 1088 videos of talking heads in the test set.  In the computations of PSNR and SSIM we
include only the face regions of the restored videos to factor out the background influences.

As shown in Table \ref{comparasion}, the MMSD-Net restoration method outperforms others by about 0.5dB on average.  This is quite remarkable as the MMSD-Net does not require any extra bit budget.  The superior
performance of the proposed method is because hidden correlations between the three modalities video, audio and emotion are fully exploited.  They provide new information and enable the CNN soft decoding process
to solve the underlying inverse problem better.

For most applications in the Internet and multimedia, the users are more interested in the perceptual quality of reconstructed videos.  The results of different competing methods can be visually compared in
Figs.\ref{Single_Results_Comparison} and \ref{Multi_Results_Comparison}. 
Fig.\ref{Single_Results_Comparison} presents restored face video frames of the same person in different emotion states, whereas
Fig.\ref{Multi_Results_Comparison} shows more results of different persons in various emotion states.
Note that in these figures the MMSD-Net optimized with the cGAN adversary loss term achieves the best perceptual quality in the comparison group.

\subsection{Results with person identification prior}

\begin{figure}[ht]
  \centering
  \includegraphics[width=0.45\textwidth]{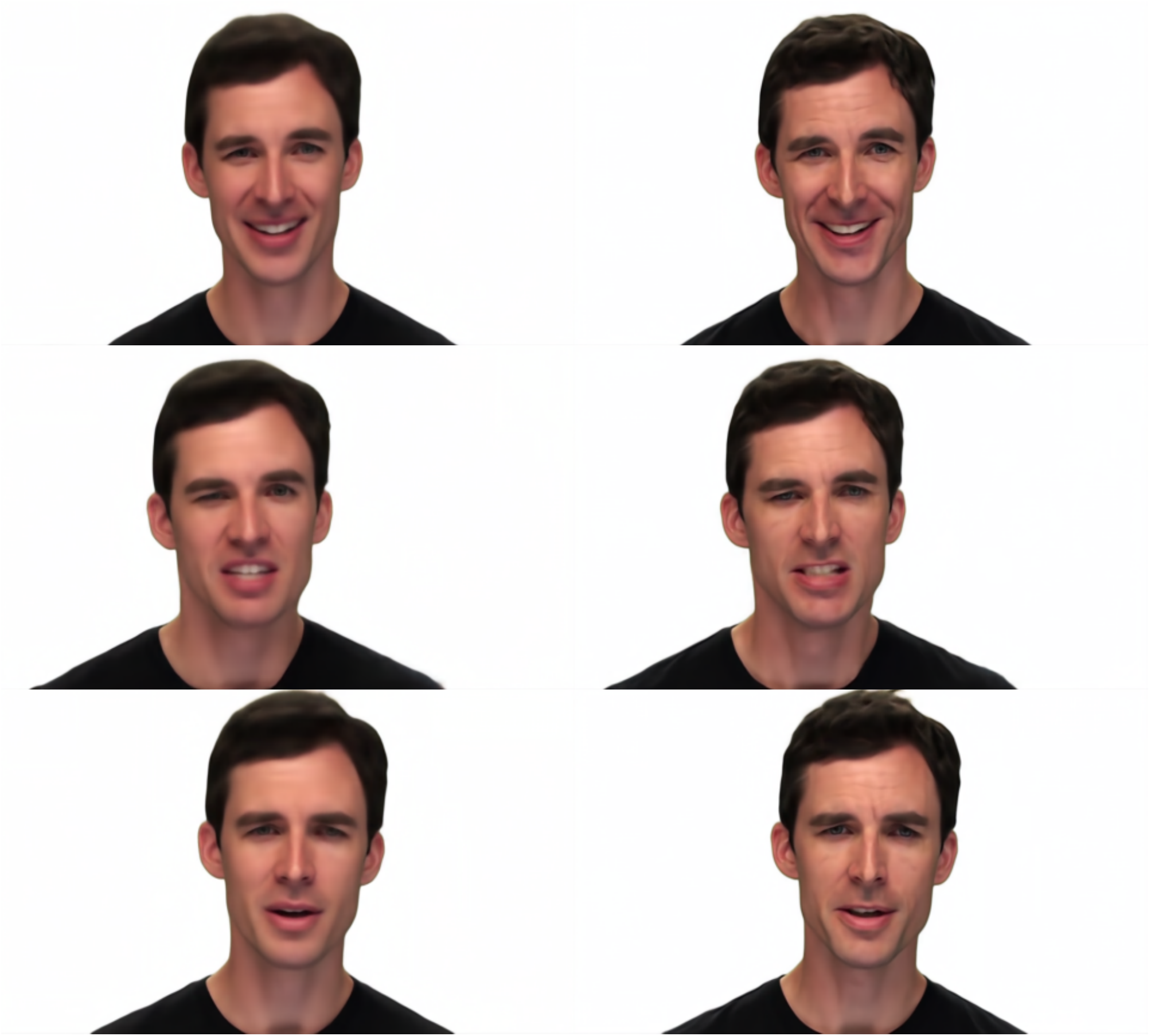}
  \caption{Results of with/without person ID prior. Left: results of the MMSD-Net optimized for a large training set. Right: results of the MMSD-Net optimized for a given person. Input compressed video is of 50Kb/s, CRF=32, downsampling factor=4.}
  \label{conclusion_img}
\end{figure}

The priors on videos of talking heads can be further strengthened if the specific speaker in the video is also known.  This is the case in many practical scenarios of internet multimedia communications.
For example, in face to face conversions among friends and relatives using smart phones the person in the video is known at each receiver side.  The MMSD-Net can be trained for a given person or a small
group of persons so that the CNN model for video restoration can be tailored to unique features of faces and voices of these individuals, and it can therefore perform even better than if the model is
trained for general public.  We have done some preliminary experiments along this line and results are presented in Fig.~ \ref{conclusion_img}.
\begin{table}[]
  \setlength{\tabcolsep}{1.8mm}{
  \begin{tabular}{@{}l rr rr rr}
    \hline
    \multirow{2}{*}{\bfseries Methods} & \multicolumn{2}{c}{\bfseries CRF=15 (4x)} &\multicolumn{2}{c}{\bfseries CRF=32 (4x) } & \multicolumn{2}{c}{\bfseries CRF=40 (4x)}\\
    & PSNR & SSIM & PSNR & SSIM & PSNR &SSIM \\
    \hline
    Bicubic    & 27.73  &	0.834 &	23.30	& 0.689 &	21.55 &	0.622 \\
    DBPN\cite{DBPN}	     & 29.62	& 0.876	& 27.02	& 0.835	& 24.73	& 0.784 \\
    EDVR\cite{EDVR}	     & 29.45	& 0.872	& 27.25 &	0.837 &	24.81 &	0.783 \\
    MMSD(Ours) & \textbf{30.12}	& \textbf{0.888}	& \textbf{27.64}	& \textbf{0.841}	& \textbf{25.38}	& \textbf{0.803} \\
    \hline
  \end{tabular}
  }
  \caption{Quantitative results (PSNR(dB)/SSIM) for compression quality parameter CRF=15, CRF=32 and CRF=40. And the downsampling factor is 4.}
  \label{comparasion}
\end{table}

\section{Conclusion}\label{Conclusion}
Videos of talking heads, an important class of video contents, can be compressed to very low bit rates and still recoverable at
satisfactory perceptual quality, if the correlations between video, audio and facial expression of the speaker can be fully exploited.
As such cross-modality correlations are highly non-linear, complex and difficult to model analytically and explicitly, the modern CNN deep
learning approach seems to be the most suitable for the video restoration task.

\bibliographystyle{ACM-Reference-Format}
\balance 
\bibliography{reference}   


\begin{thebibliography}{37}


\ifx \showCODEN    \undefined \def \showCODEN     #1{\unskip}     \fi
\ifx \showDOI      \undefined \def \showDOI       #1{#1}\fi
\ifx \showISBNx    \undefined \def \showISBNx     #1{\unskip}     \fi
\ifx \showISBNxiii \undefined \def \showISBNxiii  #1{\unskip}     \fi
\ifx \showISSN     \undefined \def \showISSN      #1{\unskip}     \fi
\ifx \showLCCN     \undefined \def \showLCCN      #1{\unskip}     \fi
\ifx \shownote     \undefined \def \shownote      #1{#1}          \fi
\ifx \showarticletitle \undefined \def \showarticletitle #1{#1}   \fi
\ifx \showURL      \undefined \def \showURL       {\relax}        \fi
\providecommand\bibfield[2]{#2}
\providecommand\bibinfo[2]{#2}
\providecommand\natexlab[1]{#1}
\providecommand\showeprint[2][]{arXiv:#2}

\bibitem[\protect\citeauthoryear{Andrew, Arora, Bilmes, and Livescu}{Andrew
  et~al\mbox{.}}{2013}]%
        {andrew2013pmlr}
\bibfield{author}{\bibinfo{person}{Galen Andrew}, \bibinfo{person}{Raman
  Arora}, \bibinfo{person}{Jeff Bilmes}, {and} \bibinfo{person}{Karen
  Livescu}.} \bibinfo{year}{2013}\natexlab{}.
\newblock \showarticletitle{Deep Canonical Correlation Analysis}. In
  \bibinfo{booktitle}{\emph{Proceedings of the 30th International Conference on
  Machine Learning}} \emph{(\bibinfo{series}{Proceedings of Machine Learning
  Research})}, \bibfield{editor}{\bibinfo{person}{Sanjoy Dasgupta} {and}
  \bibinfo{person}{David McAllester}} (Eds.), Vol.~\bibinfo{volume}{28}.
  \bibinfo{publisher}{PMLR}, \bibinfo{address}{Atlanta, Georgia, USA},
  \bibinfo{pages}{1247--1255}.
\newblock
\urldef\tempurl%
\url{http://proceedings.mlr.press/v28/andrew13.html}
\showURL{%
\tempurl}


\bibitem[\protect\citeauthoryear{Baltrusaitis, Zadeh, Lim, and
  Morency}{Baltrusaitis et~al\mbox{.}}{2018}]%
        {Face_AUs}
\bibfield{author}{\bibinfo{person}{Tadas Baltrusaitis}, \bibinfo{person}{Amir
  Zadeh}, \bibinfo{person}{Yao~Chong Lim}, {and}
  \bibinfo{person}{Louis-Philippe Morency}.} \bibinfo{year}{2018}\natexlab{}.
\newblock \showarticletitle{Openface 2.0: Facial behavior analysis toolkit}. In
  \bibinfo{booktitle}{\emph{2018 13th IEEE International Conference on
  Automatic Face \& Gesture Recognition (FG 2018)}}. IEEE,
  \bibinfo{pages}{59--66}.
\newblock


\bibitem[\protect\citeauthoryear{Dai, Qi, Xiong, Li, Zhang, Hu, and Wei}{Dai
  et~al\mbox{.}}{2017}]%
        {dai2017deformable}
\bibfield{author}{\bibinfo{person}{Jifeng Dai}, \bibinfo{person}{Haozhi Qi},
  \bibinfo{person}{Yuwen Xiong}, \bibinfo{person}{Yi Li},
  \bibinfo{person}{Guodong Zhang}, \bibinfo{person}{Han Hu}, {and}
  \bibinfo{person}{Yichen Wei}.} \bibinfo{year}{2017}\natexlab{}.
\newblock \showarticletitle{Deformable convolutional networks}. In
  \bibinfo{booktitle}{\emph{Proceedings of the IEEE international conference on
  computer vision}}. \bibinfo{pages}{764--773}.
\newblock


\bibitem[\protect\citeauthoryear{{Haris}, {Shakhnarovich}, and {Ukita}}{{Haris}
  et~al\mbox{.}}{2018}]%
        {DBPN}
\bibfield{author}{\bibinfo{person}{M. {Haris}}, \bibinfo{person}{G.
  {Shakhnarovich}}, {and} \bibinfo{person}{N. {Ukita}}.}
  \bibinfo{year}{2018}\natexlab{}.
\newblock \showarticletitle{Deep Back-Projection Networks for
  Super-Resolution}. In \bibinfo{booktitle}{\emph{2018 IEEE/CVF Conference on
  Computer Vision and Pattern Recognition}}. \bibinfo{pages}{1664--1673}.
\newblock


\bibitem[\protect\citeauthoryear{He, Hu, Zhang, Chongyang, Lin, and Han}{He
  et~al\mbox{.}}{2018}]%
        {heHevc2018}
\bibfield{author}{\bibinfo{person}{Xiaoyi He}, \bibinfo{person}{Qiang Hu},
  \bibinfo{person}{Xiaoyun Zhang}, \bibinfo{person}{Zhang Chongyang},
  \bibinfo{person}{Weiyao Lin}, {and} \bibinfo{person}{Xintong Han}.}
  \bibinfo{year}{2018}\natexlab{}.
\newblock \showarticletitle{Enhancing HEVC Compressed Videos with a
  Partition-Masked Convolutional Neural Network}. \bibinfo{pages}{216--220}.
\newblock
\urldef\tempurl%
\url{https://doi.org/10.1109/ICIP.2018.8451086}
\showDOI{\tempurl}


\bibitem[\protect\citeauthoryear{{He}, {Zuo}, {Kan}, {Shan}, and {Chen}}{{He}
  et~al\mbox{.}}{2019}]%
        {he2018tip}
\bibfield{author}{\bibinfo{person}{Z. {He}}, \bibinfo{person}{W. {Zuo}},
  \bibinfo{person}{M. {Kan}}, \bibinfo{person}{S. {Shan}}, {and}
  \bibinfo{person}{X. {Chen}}.} \bibinfo{year}{2019}\natexlab{}.
\newblock \showarticletitle{AttGAN: Facial Attribute Editing by Only Changing
  What You Want}.
\newblock \bibinfo{journal}{\emph{IEEE Transactions on Image Processing}}
  \bibinfo{volume}{28}, \bibinfo{number}{11} (\bibinfo{year}{2019}),
  \bibinfo{pages}{5464--5478}.
\newblock


\bibitem[\protect\citeauthoryear{Jamaludin, Chung, and Zisserman}{Jamaludin
  et~al\mbox{.}}{2019}]%
        {Jamaludin_2019}
\bibfield{author}{\bibinfo{person}{Amir Jamaludin}, \bibinfo{person}{Joon~Son
  Chung}, {and} \bibinfo{person}{Andrew Zisserman}.}
  \bibinfo{year}{2019}\natexlab{}.
\newblock \showarticletitle{You Said That?: Synthesising Talking Faces from
  Audio}.
\newblock \bibinfo{journal}{\emph{International Journal of Computer Vision}}
  \bibinfo{volume}{127}, \bibinfo{number}{11-12} (\bibinfo{date}{feb}
  \bibinfo{year}{2019}), \bibinfo{pages}{1767--1779}.
\newblock
\urldef\tempurl%
\url{https://doi.org/10.1007/s11263-019-01150-y}
\showDOI{\tempurl}


\bibitem[\protect\citeauthoryear{Jo, Oh, Kang, and Kim}{Jo
  et~al\mbox{.}}{2018}]%
        {Jo_2018_CVPR}
\bibfield{author}{\bibinfo{person}{Younghyun Jo}, \bibinfo{person}{Seoung~Wug
  Oh}, \bibinfo{person}{Jaeyeon Kang}, {and} \bibinfo{person}{Seon~Joo Kim}.}
  \bibinfo{year}{2018}\natexlab{}.
\newblock \showarticletitle{Deep Video Super-Resolution Network Using Dynamic
  Upsampling Filters Without Explicit Motion Compensation}. In
  \bibinfo{booktitle}{\emph{The IEEE Conference on Computer Vision and Pattern
  Recognition (CVPR)}}.
\newblock


\bibitem[\protect\citeauthoryear{Joze, Shaban, Iuzzolino, and Koishida}{Joze
  et~al\mbox{.}}{2019}]%
        {joze2019mmtm}
\bibfield{author}{\bibinfo{person}{Hamid Reza~Vaezi Joze},
  \bibinfo{person}{Amirreza Shaban}, \bibinfo{person}{Michael~L Iuzzolino},
  {and} \bibinfo{person}{Kazuhito Koishida}.} \bibinfo{year}{2019}\natexlab{}.
\newblock \showarticletitle{MMTM: Multimodal Transfer Module for CNN Fusion}.
\newblock \bibinfo{journal}{\emph{arXiv preprint arXiv:1911.08670}}
  (\bibinfo{year}{2019}).
\newblock


\bibitem[\protect\citeauthoryear{Kim, Lee, Roh, and Lee}{Kim
  et~al\mbox{.}}{2015}]%
        {kimICMI2015}
\bibfield{author}{\bibinfo{person}{Bo-Kyeong Kim}, \bibinfo{person}{Hwaran
  Lee}, \bibinfo{person}{Jihyeon Roh}, {and} \bibinfo{person}{Soo-Young Lee}.}
  \bibinfo{year}{2015}\natexlab{}.
\newblock \showarticletitle{Hierarchical Committee of Deep CNNs with
  Exponentially-Weighted Decision Fusion for Static Facial Expression
  Recognition}. In \bibinfo{booktitle}{\emph{Proceedings of the 2015 ACM on
  International Conference on Multimodal Interaction}} (Seattle, Washington,
  USA) \emph{(\bibinfo{series}{ICMI ’15})}. \bibinfo{publisher}{Association
  for Computing Machinery}, \bibinfo{address}{New York, NY, USA},
  \bibinfo{pages}{427–434}.
\newblock
\showISBNx{9781450339124}
\urldef\tempurl%
\url{https://doi.org/10.1145/2818346.2830590}
\showDOI{\tempurl}


\bibitem[\protect\citeauthoryear{Kingma and Ba}{Kingma and Ba}{2014}]%
        {kingma2014adam}
\bibfield{author}{\bibinfo{person}{Diederik~P Kingma} {and}
  \bibinfo{person}{Jimmy Ba}.} \bibinfo{year}{2014}\natexlab{}.
\newblock \showarticletitle{Adam: A method for stochastic optimization}.
\newblock \bibinfo{journal}{\emph{arXiv preprint arXiv:1412.6980}}.
\newblock


\bibitem[\protect\citeauthoryear{{Liao}, {Tao}, {Li}, {Ma}, and {Jia}}{{Liao}
  et~al\mbox{.}}{2015a}]%
        {liao2015videosp}
\bibfield{author}{\bibinfo{person}{R. {Liao}}, \bibinfo{person}{X. {Tao}},
  \bibinfo{person}{R. {Li}}, \bibinfo{person}{Z. {Ma}}, {and}
  \bibinfo{person}{J. {Jia}}.} \bibinfo{year}{2015}\natexlab{a}.
\newblock \showarticletitle{Video Super-Resolution via Deep Draft-Ensemble
  Learning}. In \bibinfo{booktitle}{\emph{2015 IEEE International Conference on
  Computer Vision (ICCV)}}. \bibinfo{pages}{531--539}.
\newblock


\bibitem[\protect\citeauthoryear{{Liao}, {Tao}, {Li}, {Ma}, and {Jia}}{{Liao}
  et~al\mbox{.}}{2015b}]%
        {liaovideosp2015}
\bibfield{author}{\bibinfo{person}{R. {Liao}}, \bibinfo{person}{X. {Tao}},
  \bibinfo{person}{R. {Li}}, \bibinfo{person}{Z. {Ma}}, {and}
  \bibinfo{person}{J. {Jia}}.} \bibinfo{year}{2015}\natexlab{b}.
\newblock \showarticletitle{Video Super-Resolution via Deep Draft-Ensemble
  Learning}. In \bibinfo{booktitle}{\emph{2015 IEEE International Conference on
  Computer Vision (ICCV)}}. \bibinfo{pages}{531--539}.
\newblock


\bibitem[\protect\citeauthoryear{Livingstone and Russo}{Livingstone and
  Russo}{2018}]%
        {RAVDESS}
\bibfield{author}{\bibinfo{person}{Steven~R. Livingstone} {and}
  \bibinfo{person}{Frank~A. Russo}.} \bibinfo{year}{2018}\natexlab{}.
\newblock \showarticletitle{The Ryerson Audio-Visual Database of Emotional
  Speech and Song (RAVDESS): A dynamic, multimodal set of facial and vocal
  expressions in North American English}.
\newblock \bibinfo{journal}{\emph{PLOS ONE}} \bibinfo{volume}{13},
  \bibinfo{number}{5} (\bibinfo{date}{05} \bibinfo{year}{2018}),
  \bibinfo{pages}{1--35}.
\newblock
\urldef\tempurl%
\url{https://doi.org/10.1371/journal.pone.0196391}
\showDOI{\tempurl}


\bibitem[\protect\citeauthoryear{Lu, Ouyang, Xu, Zhang, Gao, and Sun}{Lu
  et~al\mbox{.}}{2018a}]%
        {lu2018deep}
\bibfield{author}{\bibinfo{person}{Guo Lu}, \bibinfo{person}{Wanli Ouyang},
  \bibinfo{person}{Dong Xu}, \bibinfo{person}{Xiaoyun Zhang},
  \bibinfo{person}{Zhiyong Gao}, {and} \bibinfo{person}{Ming-Ting Sun}.}
  \bibinfo{year}{2018}\natexlab{a}.
\newblock \showarticletitle{Deep kalman filtering network for video compression
  artifact reduction}. In \bibinfo{booktitle}{\emph{Proceedings of the European
  Conference on Computer Vision (ECCV)}}. \bibinfo{pages}{568--584}.
\newblock


\bibitem[\protect\citeauthoryear{Lu, Tai, and Tang}{Lu et~al\mbox{.}}{2018b}]%
        {lu2018attribute}
\bibfield{author}{\bibinfo{person}{Yongyi Lu}, \bibinfo{person}{Yu-Wing Tai},
  {and} \bibinfo{person}{Chi-Keung Tang}.} \bibinfo{year}{2018}\natexlab{b}.
\newblock \showarticletitle{Attribute-guided face generation using conditional
  cyclegan}. In \bibinfo{booktitle}{\emph{Proceedings of the European
  Conference on Computer Vision (ECCV)}}. \bibinfo{pages}{282--297}.
\newblock


\bibitem[\protect\citeauthoryear{Ngiam, Khosla, Kim, Nam, Lee, and Ng}{Ngiam
  et~al\mbox{.}}{2011}]%
        {ngiam2011multimodal}
\bibfield{author}{\bibinfo{person}{Jiquan Ngiam}, \bibinfo{person}{Aditya
  Khosla}, \bibinfo{person}{Mingyu Kim}, \bibinfo{person}{Juhan Nam},
  \bibinfo{person}{Honglak Lee}, {and} \bibinfo{person}{Andrew~Y Ng}.}
  \bibinfo{year}{2011}\natexlab{}.
\newblock \showarticletitle{Multimodal deep learning}.
\newblock  (\bibinfo{year}{2011}).
\newblock


\bibitem[\protect\citeauthoryear{P~Ekman and Hager}{P~Ekman and Hager}{1978}]%
        {ekman1978}
\bibfield{author}{\bibinfo{person}{WV~Friesen P~Ekman} {and} \bibinfo{person}{J
  Hager}.} \bibinfo{year}{1978}\natexlab{}.
\newblock \showarticletitle{Facial Action Coding System: A Technique for the
  Measurement of Facial Movement}.
\newblock


\bibitem[\protect\citeauthoryear{Perarnau, Van De~Weijer, Raducanu, and
  {\'A}lvarez}{Perarnau et~al\mbox{.}}{2016}]%
        {perarnau2016invertible}
\bibfield{author}{\bibinfo{person}{Guim Perarnau}, \bibinfo{person}{Joost Van
  De~Weijer}, \bibinfo{person}{Bogdan Raducanu}, {and} \bibinfo{person}{Jose~M
  {\'A}lvarez}.} \bibinfo{year}{2016}\natexlab{}.
\newblock \showarticletitle{Invertible conditional gans for image editing}.
\newblock \bibinfo{journal}{\emph{arXiv preprint arXiv:1611.06355}}
  (\bibinfo{year}{2016}).
\newblock


\bibitem[\protect\citeauthoryear{P{\'e}rez-R{\'u}a, Vielzeuf, Pateux,
  Baccouche, and Jurie}{P{\'e}rez-R{\'u}a et~al\mbox{.}}{2019}]%
        {perez2019mfas}
\bibfield{author}{\bibinfo{person}{Juan-Manuel P{\'e}rez-R{\'u}a},
  \bibinfo{person}{Valentin Vielzeuf}, \bibinfo{person}{St{\'e}phane Pateux},
  \bibinfo{person}{Moez Baccouche}, {and} \bibinfo{person}{Fr{\'e}d{\'e}ric
  Jurie}.} \bibinfo{year}{2019}\natexlab{}.
\newblock \showarticletitle{Mfas: Multimodal fusion architecture search}. In
  \bibinfo{booktitle}{\emph{Proceedings of the IEEE Conference on Computer
  Vision and Pattern Recognition}}. \bibinfo{pages}{6966--6975}.
\newblock


\bibitem[\protect\citeauthoryear{Sahidullah and Saha}{Sahidullah and
  Saha}{2012a}]%
        {MFCC_2}
\bibfield{author}{\bibinfo{person}{Md. Sahidullah} {and}
  \bibinfo{person}{Goutam Saha}.} \bibinfo{year}{2012}\natexlab{a}.
\newblock \showarticletitle{Design, analysis and experimental evaluation of
  block based transformation in MFCC computation for speaker recognition}.
\newblock \bibinfo{journal}{\emph{Speech Communication}} \bibinfo{volume}{54},
  \bibinfo{number}{4} (\bibinfo{year}{2012}), \bibinfo{pages}{543 -- 565}.
\newblock
\showISSN{0167-6393}
\urldef\tempurl%
\url{https://doi.org/10.1016/j.specom.2011.11.004}
\showDOI{\tempurl}


\bibitem[\protect\citeauthoryear{Sahidullah and Saha}{Sahidullah and
  Saha}{2012b}]%
        {MFCC_1}
\bibfield{author}{\bibinfo{person}{Md Sahidullah} {and} \bibinfo{person}{Goutam
  Saha}.} \bibinfo{year}{2012}\natexlab{b}.
\newblock \showarticletitle{A novel windowing technique for efficient
  computation of MFCC for speaker recognition}.
\newblock \bibinfo{journal}{\emph{IEEE signal processing letters}}
  \bibinfo{volume}{20}, \bibinfo{number}{2} (\bibinfo{year}{2012}),
  \bibinfo{pages}{149--152}.
\newblock


\bibitem[\protect\citeauthoryear{{Sikora}}{{Sikora}}{1997}]%
        {MPEG4}
\bibfield{author}{\bibinfo{person}{T. {Sikora}}.}
  \bibinfo{year}{1997}\natexlab{}.
\newblock \showarticletitle{The MPEG-4 video standard verification model}.
\newblock \bibinfo{journal}{\emph{IEEE Transactions on Circuits and Systems for
  Video Technology}} \bibinfo{volume}{7}, \bibinfo{number}{1}
  (\bibinfo{year}{1997}), \bibinfo{pages}{19--31}.
\newblock


\bibitem[\protect\citeauthoryear{Sullivan, Ohm, Han, and Wiegand}{Sullivan
  et~al\mbox{.}}{2012}]%
        {HEVC}
\bibfield{author}{\bibinfo{person}{Gary~J Sullivan},
  \bibinfo{person}{Jens-Rainer Ohm}, \bibinfo{person}{Woo-Jin Han}, {and}
  \bibinfo{person}{Thomas Wiegand}.} \bibinfo{year}{2012}\natexlab{}.
\newblock \showarticletitle{Overview of the high efficiency video coding (HEVC)
  standard}.
\newblock \bibinfo{journal}{\emph{IEEE Transactions on circuits and systems for
  video technology}} \bibinfo{volume}{22}, \bibinfo{number}{12}
  (\bibinfo{year}{2012}), \bibinfo{pages}{1649--1668}.
\newblock


\bibitem[\protect\citeauthoryear{Suwajanakorn, Seitz, and
  Kemelmacher-Shlizerman}{Suwajanakorn et~al\mbox{.}}{2017}]%
        {suwajanakorn2017synthesizing}
\bibfield{author}{\bibinfo{person}{Supasorn Suwajanakorn},
  \bibinfo{person}{Steven~M Seitz}, {and} \bibinfo{person}{Ira
  Kemelmacher-Shlizerman}.} \bibinfo{year}{2017}\natexlab{}.
\newblock \showarticletitle{Synthesizing obama: learning lip sync from audio}.
\newblock \bibinfo{journal}{\emph{ACM Transactions on Graphics (TOG)}}
  \bibinfo{volume}{36}, \bibinfo{number}{4} (\bibinfo{year}{2017}),
  \bibinfo{pages}{1--13}.
\newblock


\bibitem[\protect\citeauthoryear{Tao, Gao, Liao, Wang, and Jia}{Tao
  et~al\mbox{.}}{2017}]%
        {tao2017detail}
\bibfield{author}{\bibinfo{person}{Xin Tao}, \bibinfo{person}{Hongyun Gao},
  \bibinfo{person}{Renjie Liao}, \bibinfo{person}{Jue Wang}, {and}
  \bibinfo{person}{Jiaya Jia}.} \bibinfo{year}{2017}\natexlab{}.
\newblock \showarticletitle{Detail-revealing deep video super-resolution}. In
  \bibinfo{booktitle}{\emph{Proceedings of the IEEE International Conference on
  Computer Vision}}. \bibinfo{pages}{4472--4480}.
\newblock


\bibitem[\protect\citeauthoryear{Tian, Zhang, Fu, and Xu}{Tian
  et~al\mbox{.}}{2018}]%
        {TDAN}
\bibfield{author}{\bibinfo{person}{Yapeng Tian}, \bibinfo{person}{Yulun Zhang},
  \bibinfo{person}{Yun Fu}, {and} \bibinfo{person}{Chenliang Xu}.}
  \bibinfo{year}{2018}\natexlab{}.
\newblock \showarticletitle{TDAN: Temporally Deformable Alignment Network for
  Video Super-Resolution}.
\newblock


\bibitem[\protect\citeauthoryear{Vaswani, Shazeer, Parmar, Uszkoreit, Jones,
  Gomez, Kaiser, and Polosukhin}{Vaswani et~al\mbox{.}}{2017}]%
        {Attention_NIPS}
\bibfield{author}{\bibinfo{person}{Ashish Vaswani}, \bibinfo{person}{Noam
  Shazeer}, \bibinfo{person}{Niki Parmar}, \bibinfo{person}{Jakob Uszkoreit},
  \bibinfo{person}{Llion Jones}, \bibinfo{person}{Aidan~N. Gomez},
  \bibinfo{person}{undefinedukasz Kaiser}, {and} \bibinfo{person}{Illia
  Polosukhin}.} \bibinfo{year}{2017}\natexlab{}.
\newblock \showarticletitle{Attention is All You Need}. In
  \bibinfo{booktitle}{\emph{Proceedings of the 31st International Conference on
  Neural Information Processing Systems}} (Long Beach, California, USA)
  \emph{(\bibinfo{series}{NIPS’17})}. \bibinfo{publisher}{Curran Associates
  Inc.}, \bibinfo{address}{Red Hook, NY, USA}, \bibinfo{pages}{6000–6010}.
\newblock
\showISBNx{9781510860964}


\bibitem[\protect\citeauthoryear{Wang, Chan, Yu, Dong, and Loy}{Wang
  et~al\mbox{.}}{2019}]%
        {EDVR}
\bibfield{author}{\bibinfo{person}{Xintao Wang}, \bibinfo{person}{Kelvin C.~K.
  Chan}, \bibinfo{person}{Ke Yu}, \bibinfo{person}{Chao Dong}, {and}
  \bibinfo{person}{Chen~Change Loy}.} \bibinfo{year}{2019}\natexlab{}.
\newblock \showarticletitle{{EDVR:} Video Restoration with Enhanced Deformable
  Convolutional Networks}.
\newblock \bibinfo{journal}{\emph{CoRR}}  \bibinfo{volume}{abs/1905.02716}.
\newblock
\showeprint[arxiv]{1905.02716}
\urldef\tempurl%
\url{http://arxiv.org/abs/1905.02716}
\showURL{%
\tempurl}


\bibitem[\protect\citeauthoryear{Wang, Girshick, Gupta, and He}{Wang
  et~al\mbox{.}}{2018}]%
        {wang2018non}
\bibfield{author}{\bibinfo{person}{Xiaolong Wang}, \bibinfo{person}{Ross
  Girshick}, \bibinfo{person}{Abhinav Gupta}, {and} \bibinfo{person}{Kaiming
  He}.} \bibinfo{year}{2018}\natexlab{}.
\newblock \showarticletitle{Non-local neural networks}. In
  \bibinfo{booktitle}{\emph{Proceedings of the IEEE conference on computer
  vision and pattern recognition}}. \bibinfo{pages}{7794--7803}.
\newblock


\bibitem[\protect\citeauthoryear{{Wiegand}, {Sullivan}, {Bjontegaard}, and
  {Luthra}}{{Wiegand} et~al\mbox{.}}{2003}]%
        {H264}
\bibfield{author}{\bibinfo{person}{T. {Wiegand}}, \bibinfo{person}{G.~J.
  {Sullivan}}, \bibinfo{person}{G. {Bjontegaard}}, {and} \bibinfo{person}{A.
  {Luthra}}.} \bibinfo{year}{2003}\natexlab{}.
\newblock \showarticletitle{Overview of the H.264/AVC video coding standard}.
\newblock \bibinfo{journal}{\emph{IEEE Transactions on Circuits and Systems for
  Video Technology}} \bibinfo{volume}{13}, \bibinfo{number}{7}
  (\bibinfo{year}{2003}), \bibinfo{pages}{560--576}.
\newblock


\bibitem[\protect\citeauthoryear{Wiles, Sophia~Koepke, and Zisserman}{Wiles
  et~al\mbox{.}}{2018}]%
        {Wiles_2018_ECCV}
\bibfield{author}{\bibinfo{person}{Olivia Wiles}, \bibinfo{person}{A.
  Sophia~Koepke}, {and} \bibinfo{person}{Andrew Zisserman}.}
  \bibinfo{year}{2018}\natexlab{}.
\newblock \showarticletitle{X2Face: A network for controlling face generation
  using images, audio, and pose codes}. In \bibinfo{booktitle}{\emph{The
  European Conference on Computer Vision (ECCV)}}.
\newblock


\bibitem[\protect\citeauthoryear{Xu, Gao, Tian, Zhou, and Sun}{Xu
  et~al\mbox{.}}{2019}]%
        {xu2019non}
\bibfield{author}{\bibinfo{person}{Yi Xu}, \bibinfo{person}{Longwen Gao},
  \bibinfo{person}{Kai Tian}, \bibinfo{person}{Shuigeng Zhou}, {and}
  \bibinfo{person}{Huyang Sun}.} \bibinfo{year}{2019}\natexlab{}.
\newblock \showarticletitle{Non-Local ConvLSTM for Video Compression Artifact
  Reduction}. In \bibinfo{booktitle}{\emph{Proceedings of the IEEE
  International Conference on Computer Vision}}. \bibinfo{pages}{7043--7052}.
\newblock


\bibitem[\protect\citeauthoryear{Xue, Chen, Wu, Wei, and Freeman}{Xue
  et~al\mbox{.}}{2019}]%
        {xue2019video}
\bibfield{author}{\bibinfo{person}{Tianfan Xue}, \bibinfo{person}{Baian Chen},
  \bibinfo{person}{Jiajun Wu}, \bibinfo{person}{Donglai Wei}, {and}
  \bibinfo{person}{William~T Freeman}.} \bibinfo{year}{2019}\natexlab{}.
\newblock \showarticletitle{Video enhancement with task-oriented flow}.
\newblock \bibinfo{journal}{\emph{International Journal of Computer Vision}}
  \bibinfo{volume}{127}, \bibinfo{number}{8} (\bibinfo{year}{2019}),
  \bibinfo{pages}{1106--1125}.
\newblock


\bibitem[\protect\citeauthoryear{Yang, Xu, Wang, and Li}{Yang
  et~al\mbox{.}}{2018a}]%
        {yang2018multi}
\bibfield{author}{\bibinfo{person}{Ren Yang}, \bibinfo{person}{Mai Xu},
  \bibinfo{person}{Zulin Wang}, {and} \bibinfo{person}{Tianyi Li}.}
  \bibinfo{year}{2018}\natexlab{a}.
\newblock \showarticletitle{Multi-frame quality enhancement for compressed
  video}. In \bibinfo{booktitle}{\emph{Proceedings of the IEEE Conference on
  Computer Vision and Pattern Recognition}}. \bibinfo{pages}{6664--6673}.
\newblock


\bibitem[\protect\citeauthoryear{Yang, Xu, Wang, and Li}{Yang
  et~al\mbox{.}}{2018b}]%
        {yang2018mfqe}
\bibfield{author}{\bibinfo{person}{Ren Yang}, \bibinfo{person}{Mai Xu},
  \bibinfo{person}{Zulin Wang}, {and} \bibinfo{person}{Tianyi Li}.}
  \bibinfo{year}{2018}\natexlab{b}.
\newblock \showarticletitle{Multi-Frame Quality Enhancement for Compressed
  Video}.
\newblock  (\bibinfo{date}{03} \bibinfo{year}{2018}).
\newblock


\bibitem[\protect\citeauthoryear{Zhang, Zhang, Huang, and Gao}{Zhang
  et~al\mbox{.}}{2016}]%
        {zhangmdcnn2016}
\bibfield{author}{\bibinfo{person}{Shiqing Zhang}, \bibinfo{person}{Shiliang
  Zhang}, \bibinfo{person}{Tiejun Huang}, {and} \bibinfo{person}{Wen Gao}.}
  \bibinfo{year}{2016}\natexlab{}.
\newblock \showarticletitle{Multimodal Deep Convolutional Neural Network for
  Audio-Visual Emotion Recognition}. \bibinfo{pages}{281--284}.
\newblock
\urldef\tempurl%
\url{https://doi.org/10.1145/2911996.2912051}
\showDOI{\tempurl}


\end{thebibliography}
\end{document}